\RequirePackage{fix-cm}
\documentclass{article}
\usepackage{iclr2027_conference,times}

\usepackage[T5]{fontenc}
\DeclareUnicodeCharacter{266B}{}
\DeclareUnicodeCharacter{1ED3}{\`{\^o}}

\usepackage{amsmath}
\usepackage{amssymb}
\usepackage{mathtools}
\usepackage{amsthm}
\usepackage{bbm}

\usepackage{amsmath,amsfonts,bm}

\def\eqref#1{equation~\ref{#1}}

\def\1{\bm{1}}

\DeclareMathAlphabet{\mathsfit}{\encodingdefault}{\sfdefault}{m}{sl}
\SetMathAlphabet{\mathsfit}{bold}{\encodingdefault}{\sfdefault}{bx}{n}

\usepackage{xcolor}
\usepackage{graphicx}
\usepackage{wrapfig}
\usepackage{tikz}
\usepackage{pgfplots}
\usepackage{pifont}
\usepackage{fontawesome5}

\pgfplotsset{compat=1.18}
\usepgfplotslibrary{groupplots}
\usetikzlibrary{
  arrows.meta,
  fit,
  backgrounds,
  calc,
  positioning,
  shapes.geometric
}

\usepackage{array}
\usepackage{booktabs}
\usepackage{multirow}
\usepackage{makecell}
\usepackage{tabularx}
\usepackage{tabularray}
\UseTblrLibrary{booktabs}
\usepackage{colortbl}
\usepackage{arydshln}

\newcolumntype{Y}{>{\centering\arraybackslash}X}

\usepackage{algorithm}
\usepackage{algpseudocode}
\usepackage{enumitem}
\usepackage[normalem]{ulem}
\usepackage{comment}
\usepackage[most]{tcolorbox}

\usepackage{url}
\usepackage{hyperref}

\theoremstyle{plain}

\theoremstyle{definition}
\newtheorem{definition}{Definition}

\definecolor{evidencebg}{RGB}{255,242,224}
\definecolor{dependencybg}{RGB}{230,240,250}

\definecolor{examplebg}{HTML}{F0F9FF}
\definecolor{exampleframe}{HTML}{D0D6DA}

\newcommand{\method}{\textsc{E-Closure}}

\newcommand{\tightmath}{%
  \footnotesize
  \setlength{\abovedisplayskip}{3pt}%
  \setlength{\belowdisplayskip}{3pt}%
  \setlength{\abovedisplayshortskip}{2pt}%
  \setlength{\belowdisplayshortskip}{2pt}%
}

\newtcolorbox{motivationbox}{
  enhanced,
  colback=examplebg,
  colframe=exampleframe,
  boxrule=0.45pt,
  arc=1mm,
  boxsep=0pt,
  left=1.7mm,
  right=1.7mm,
  top=1.5mm,
  bottom=1.5mm,
  before skip=6pt,
  after skip=8pt,
  fontupper=\footnotesize
}

\title{
  {\fontsize{16.2pt}{7pt}\selectfont
    Locally Sound, Globally Insufficient:
    The Local-Global Gap in Multi-Hop Reasoning}
}

\author{\textbf{Bohao Chu}\textsuperscript{1,*},\;
\textbf{Hendrik Damm}\textsuperscript{2,5},\;
\textbf{Qianli Wang}\textsuperscript{3},\;
\textbf{Hui Wang}\textsuperscript{1},\;
\textbf{Shuning Zhang}\textsuperscript{4},\; \\
\textbf{Christoph M. Friedrich}\textsuperscript{2,5},\;
\textbf{Norbert Fuhr}\textsuperscript{1} \\
\textsuperscript{1}University of Duisburg-Essen,
\textsuperscript{2}University of Applied Sciences and Arts Dortmund, \\
\textsuperscript{3}Technische Universit\"at Berlin,
\textsuperscript{4}Tsinghua University, \\
\textsuperscript{5}Institute for Medical Informatics, Biometry and Epidemiology (IMIBE) \\
\texttt{\textsuperscript{*} \faEnvelope \; bohao.chu@qq.com}
}

\iclrfinalcopy

\begin{document}

\maketitle
\fancyhead{}
\fancyhead[L]{Preprint}

\vspace{-15pt}
\begin{abstract}
Reliable multi-hop reasoning requires more than locally supported steps: a trace can be sound at every reasoning step yet still fail to answer the question as a whole. We call this failure regime the \emph{local--global gap} (LGG), in which the trace is \textbf{locally sound yet globally insufficient}. Local soundness requires each step to be supported by the available evidence and preceding steps, whereas global sufficiency requires the reasoning trace to align with the question and establish the submitted answer. In a human-adjudicated diagnostic of \(2{,}598\) responses across three multi-hop QA benchmarks and three models, we find that the LGG occurs in every benchmark--model combination and accounts for nearly half of globally insufficient responses overall. However, conventional faithfulness verifiers that check claims against the evidence largely miss these failures: at thresholds retaining at least \(95\%\) of reliable traces, recall for LGG cases is substantially lower than that for locally unsound traces. To address these failures, we formalize three dependencies for reliable reasoning: evidence-to-step support, question-to-trace alignment, and trace-to-answer closure. Instead of post-hoc diagnosis, we introduce \textbf{\method{}} to supervise these dependencies during training, combining generation supervision on supported original and counterfactual responses with bidirectional switching constraints. 
Averaged over three benchmarks and three backbones, existing fine-tuning baselines improve accuracy and local soundness over the base models, but at the cost of global sufficiency. 
\textbf{\method{}} improves both: among all fine-tuned methods, it achieves the highest average accuracy (92.8\%) and trace reliability (89.0\%) while yielding the lowest LGG rate (6.2\%). \looseness=-1


\end{abstract}

\section{Introduction}
\label{sec:intro}

Multi-hop reasoning often combines external evidence~\citep{NEURIPS2020_6b493230} with chain-of-thought (CoT) traces~\citep{wei2022chain,kojima2022large}. Yet a trace can fail even when each step is supported by the available evidence and preceding steps. It may pursue the wrong target, or establish the correct answer while the model submits a different one. Neither is a standard \emph{hallucination} within the trace, as every substantive step remains supported. The latter case, illustrated below, cannot be resolved by additional evidence or reasoning: the failure lies entirely in the trace-to-answer link. \looseness=-1

\begin{motivationbox}
\textbf{Question:} Which film has the director born first, \emph{`Please Give'} or \emph{`Moonstruck'}? \textsc{(2WikiMultihopQA)}
\vspace{3pt}
{\color{exampleframe}\hrule}
\vspace{3pt}
\noindent
\begin{minipage}[t]{0.53\linewidth}
\textsc{Contextual evidence}
\vspace{1pt}

\textbf{[0]} Norman Jewison was born in 1926.
\vspace{2pt}

\textbf{[4]} \emph{Moonstruck} was directed by Norman Jewison.
\vspace{2pt}

\textbf{[7]} Nicole Holofcener was born in 1960.
\vspace{2pt}

\textbf{[9]} \emph{Please Give} was directed by Nicole Holofcener.
\vspace{2pt}

{\centering \textbf{\ldots}\par}
\end{minipage}
\hfill
\begin{minipage}[t]{0.46\linewidth}
\textsc{Model trace (Qwen3-8B)}
\vspace{1pt}

1. \emph{`Moonstruck'} was directed by Jewison [4].
\vspace{1pt}

2. Jewison was born in 1926 [0].
\vspace{1pt}

3. \emph{`Please Give'} was directed by Holofcener [9].
\vspace{1pt}

4. Holofcener was born in 1960 [7].
\vspace{1pt}

5. Jewison was born before Holofcener.
\vspace{1pt}

\textbf{Answer:} \textcolor{red!70!black}{\emph{`Please Give'}}.
\end{minipage}
\vspace{3pt}
{\color{exampleframe}\hrule}
\vspace{3pt}
{\centering
\textbf{Local soundness} $\checkmark$
\qquad
\textbf{Alignment} $\checkmark$
\qquad
\textbf{Closure} $\times$
\par}
\vspace{1pt}
{\centering
\emph{The trace already establishes \textcolor{green!45!black}{`Moonstruck'}, but the submitted answer names the wrong film \textcolor{red!70!black}{`Please Give'}.}
\par}
\end{motivationbox}

We formalize this mismatch by distinguishing \textbf{local soundness}, which asks whether each substantive step is supported by the evidence and preceding steps, from \textbf{global sufficiency}, which requires the trace to align with the question and establish its stated answer. We say that a response exhibits the \emph{local--global gap} (LGG) when it is \textbf{locally sound yet globally insufficient}. In a human-adjudicated diagnostic of \(2{,}598\) responses from three multi-hop question answering (QA) benchmarks and three models, we identify \(145\) LGG cases across every benchmark and model combination, accounting for \(7.0\%\) of locally sound responses and \(46.3\%\) of globally insufficient responses. The LGG primarily manifests as \emph{path deviation} at the question-to-trace dependency (\(35.2\%\)) and \emph{answer decoupling} at the trace-to-answer dependency (\(47.6\%\)), while the remaining \(17.2\%\) reflects \emph{incomplete reasoning}.

However, conventional faithfulness verifiers that check claims against the evidence largely miss these failures. We evaluate RAGAS~\citep{es-etal-2024-ragas}, VeriScore~\citep{song-etal-2024-veriscore}, and AlignScore~\citep{zha-etal-2023-alignscore} as evidence-support verifiers. At thresholds retaining at least \(95\%\) of reliable traces, their highest recall is only \(11.0\%\) for LGG cases, compared with \(40.8\%\) for locally unsound responses. We therefore assess the two global dependencies directly with an \textit{alignment} judge, which checks whether the trace pursues the question's target, and a \textit{closure} judge that checks whether the trace establishes the submitted answer. Neither judge receives the evidence, so each can focus on its respective link. Under the same retention criterion, alignment detects \(66.7\%\) of \emph{path deviations} and closure detects \(56.5\%\) of \emph{answer-decoupling} cases within the LGG. These results highlight the limits of evidence-support verification alone and the need to check both global dependencies.

To address these failures, we formalize three dependencies for reliable reasoning: evidence-to-step support, question-to-trace alignment, and trace-to-answer closure. We introduce \textbf{\method{}} to supervise these dependencies during training through paired interventions: changing supporting evidence must switch its first dependent step, changing the question target must switch the trace, and changing the trace must switch the answer. Each constraint applies in both directions, requiring the preference to reverse with the conditioning input. Across three benchmarks and three model backbones, \method{} achieves the highest average answer accuracy (\(92.8\%\)) and trace reliability (\(89.0\%\)), together with the lowest average LGG rate (\(6.2\%\)) among the evaluated fine-tuned methods.

Our contributions: \textbf{(1)} We formalize the \textbf{local--global gap} in multi-hop reasoning by distinguishing local soundness from global sufficiency, which requires question alignment and answer closure. \textbf{(2)} In a human-adjudicated study of \(2{,}598\) responses across three benchmarks and three models, we identify the LGG as a major source of global insufficiency. However, conventional faithfulness verifiers miss most LGG cases, whereas alignment and closure judges detect them more effectively. \textbf{(3)} We introduce \textbf{\method{}}, a training objective combining supported original and counterfactual generation anchors with bidirectional supervision of the three dependencies to improve both local soundness and global sufficiency, without changing inference.

\section{Related Work}
\label{sec:related}

\textbf{Multi-hop reasoning.}
Multi-hop reasoning combines supporting evidence across dependent steps, often amid plausible distractors~\citep{ho-etal-2020-constructing,trivedi-etal-2022-musique,yang-etal-2018-hotpotqa}.
Prior work supports this process through retrieval-augmented generation (RAG)~\citep{NEURIPS2020_6b493230}, chain-of-thought (CoT) prompting~\citep{wei2022chain,kojima2022large}, question decomposition~\citep{press-etal-2023-measuring,zhou2023leasttomost}, and interleaved retrieval and generation~\citep{trivedi-etal-2023-interleaving,jiang-etal-2023-active}. Yet sufficient evidence alone does not ensure that a trace addresses the question or establishes the submitted answer. We therefore use the \emph{evidence-present} setting, where sufficient evidence is provided, to focus on reasoning failures rather than failures due to insufficient evidence.

\textbf{Reasoning quality and evaluation.}
Reasoning quality concerns the validity of stated reasoning, whereas CoT faithfulness asks whether the generated trace reflects the model's actual answer-generation process \citep{turpin2023language,lanham2023measuringfaithfulnesschainofthoughtreasoning}.
\texttt{ROSCOE}, \texttt{ReCEval}, and \texttt{REVEAL} assess complementary aspects of multi-step reasoning, including correctness, informativeness, relevance, and evidence support \citep{golovneva2023roscoe,prasad-etal-2023-receval,jacovi-etal-2024-chain}.
\texttt{RACE} jointly evaluates reasoning and answer consistency for hallucination detection in large reasoning models \citep{wang2026race}, while \texttt{MCR} rewards visual grounding and reasoning--answer entailment in medical VQA \citep{jiang-etal-2026-act}.
However, these approaches do not explicitly distinguish question alignment and answer closure as separate global requirements alongside step-level evidence support.
We formalize this distinction to identify and systematically characterize the LGG: all substantive steps are supported, yet alignment or closure fails.
\method{} explicitly supervises these dependencies during training through dependency-specific, bidirectional counterfactual constraints.

\textbf{Counterfactual and preference learning.}
Counterfactual augmentation and contrast sets pair small input edits with the output changes they warrant~\citep{kaushik2020learning,gardner-etal-2020-evaluating,wang-etal-2025-truth}, while preference optimization learns which response to favor under a given input~\citep{rafailov2023direct}.
\method{} combines these ideas through dependency-specific preference supervision.
Evidence interventions supervise evidence-to-step and trace-to-answer switching, while question interventions supervise question-to-trace switching.
Generation anchors train the supported responses under original and counterfactual inputs, while bidirectional margins require preferences to reverse across the paired conditions.
Scoring the first affected step, full trace, and answer separately targets each dependency and gives the answer its own supervision signal.

\section{Problem Formulation and Diagnosis}
\label{sec:diagnosis}

\subsection{Problem Setup and Definitions}
\label{sec:formulation}

A reasoning instance is a tuple \(x=(q,\mathcal{D}_q,a^\star)\), where \(q\) is a question, \(\mathcal{D}_q=\{d_1,\ldots,d_m\}\) is its contextual evidence set, and \(a^\star\) is the reference answer. We assume that \(\mathcal{D}_q\) uniquely warrants \(a^\star\). A model response is \(y=(r,a)\), where \(r=(s_1,\ldots,s_T)\) is the reasoning trace and \(a\) is the submitted answer. We set \(\mathrm{Acc}(y)=1\) if \(a\) is judged semantically equivalent to \(a^\star\), and \(0\) otherwise. Reliability requires three dependencies among \(q\), \(\mathcal{D}_q\), \(r\), and \(a\) to hold simultaneously. We group them into two criteria: whether the evidence warrants the trace (\emph{local soundness}) and whether the trace addresses the question and establishes the submitted answer (\emph{global sufficiency}).

\begin{definition}[Local soundness]
\label{def:local}
A substantive step \(s_t\) is sound if it is supported by the evidence \(\mathcal{D}_q\) and preceding steps \(s_{<t}=(s_1,\ldots,s_{t-1})\). Let \(\ell(s_t\mid\mathcal{D}_q,s_{<t})\) equal \(1\) if \(s_t\) is sound and \(0\) otherwise. With \(\mathcal{S}\subseteq\{1,\ldots,T\}\) indexing the substantive steps, define
{\tightmath
\[
L(y):=\prod_{t\in\mathcal{S}}\ell\!\left(s_t\mid\mathcal{D}_q,s_{<t}\right).
\]
}A trace is locally sound if \(L(y)=1\), meaning that every substantive step is sound. 
\end{definition}

\begin{definition}[Global sufficiency]
\label{def:global}
A response is globally sufficient if its trace addresses the question and establishes the submitted answer.
The trace is aligned, \(A(y)=1\), if it pursues the target requested by \(q\).
The response is closed, \(C(y)=1\), if the claims stated in \(r\), taken as premises, establish the submitted answer \(a\) as a conclusion for the target pursued by the trace.
Global sufficiency is achieved when both alignment and closure hold:
{\tightmath
\[
q \;\xrightarrow{\;A\;}\; r \;\xrightarrow{\;C\;}\; a,
\qquad G(y):=A(y)\,C(y).
\]
}
\end{definition}

\begin{definition}[Local--global gap]
\label{def:gap}
A response exhibits the local--global gap (LGG) when it is locally sound but globally insufficient:
{\tightmath
\[
\mathrm{LGG}(y):=\mathbbm{1}[L(y)=1,\;G(y)=0].
\]
}We classify LGG cases by the first broken global link.
\textbf{Path deviation} occurs when \(A(y)=0\): the trace pursues a target other than the one requested.
When \(A(y)=1\) but \(C(y)=0\), the trace pursues the requested target but does not establish the submitted answer.
Within this case, we further distinguish \textbf{incomplete reasoning}, where the trace does not establish the reference answer, from \textbf{answer decoupling}, where it establishes the reference answer, but the response submits a different one. \looseness=-1

Local soundness is assessed separately: a trace may contain an unsupported step yet remain aligned and establish its answer from its stated premises.
\end{definition}

\subsection{Diagnostic Study}
\label{sec:diagnostic_protocol}

\textbf{Data and models.}
To balance sample sizes across benchmarks after screening, we sample \(1{,}300\) questions from the held-out evaluation splits in the distractor setting: \(300\) from 2WikiMultiHopQA~\citep{ho-etal-2020-constructing}, \(600\) from MuSiQue~\citep{trivedi-etal-2022-musique}, and \(400\) from HotpotQA~\citep{yang-etal-2018-hotpotqa}.
We screen for evidence--gold eligibility to avoid mistaking evidence or reference-answer defects for reasoning failures, retaining \(866\) questions.
Using a shared structured CoT prompt and deterministic decoding, we generate one response per question from Qwen3-8B, Qwen3-32B~\citep{yang2025qwen3technicalreport}, and Ministral-3-8B-Instruct-2512~\citep{liu2026ministral3}, yielding \(2{,}598\) diagnostic responses.
Sampling, generation, and screen details appear in Appendices~\ref{app:diagnostic_sample}, \ref{app:generation}, and \ref{app:annotation}.

\textbf{Screening and annotation.}
Using only the question, provided evidence, and reference answer, two screeners independently verify that the evidence uniquely warrants the reference answer.
Six other annotators then label the retained responses using a balanced incomplete assignment: each response receives two independent annotations, and each annotator sees at most one model response per question.
Annotators first identify substantive steps and assess their soundness, then judge alignment using the same inputs.
They next assess closure after the submitted answer is revealed, and finally judge answer correctness and whether the trace establishes the reference answer after \(a^\star\) is revealed.
A separate adjudicator resolves disagreements, and the adjudicated labels determine \(L\), \(G\), and the failure classes.
Full screening and annotation guidelines appear in Appendix~\ref{app:annotation}.

\textbf{Agreement.}
We report raw agreement and nominal Krippendorff's \(\alpha\)~\citep{krippendorff2004reliability} before adjudication for eligibility screening and response annotation.
Eligibility screening yields \(91.0\%\) raw agreement and \(\alpha=0.783\).
For response annotation, raw agreement ranges from \(95.0\%\) to \(99.9\%\) across the directly annotated binary variables.
The corresponding \(\alpha\) values are \(0.989\) for substantive status, \(0.726\) for step-level soundness, \(0.739\) for alignment, \(0.797\) for closure, \(0.759\) for completeness, and \(0.990\) for answer correctness.
Any step- or response-level disagreement or consistency-check violation triggers adjudication, affecting \(823\) responses (\(31.7\%\)).
All diagnostic analyses use the final adjudicated labels.
Full agreement statistics appear in Appendix~\ref{app:annotation}.

\begin{figure}[t!]
    \centering
    \includegraphics[width=\textwidth]{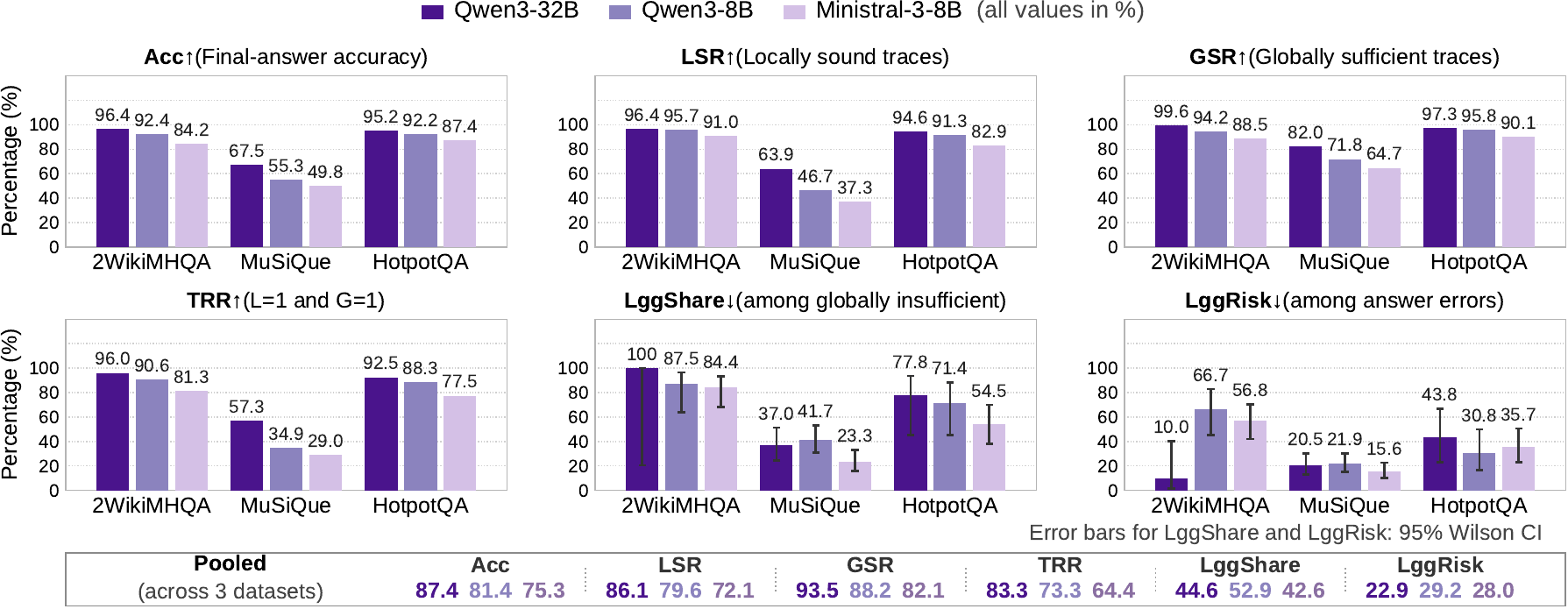}
    \vspace{-17pt}
    \caption{\textbf{Local and global reasoning diagnostics.}
    Bars show percentages for each dataset--model pair.
    The bottom row pools all three datasets, with \(866\) responses per model.
    Pooled LggShare and LggRisk are computed using their respective conditional denominators.
    The main text reports results pooled over all \(2{,}598\) responses, with the full joint distribution provided in Appendix~\ref{app:failure_composition}.}
    \vspace{-13pt}
    \label{fig:diagnostic_results}
\end{figure}

\textbf{Measures.}
Let \(\mathcal{R}\) denote the \(2{,}598\) adjudicated responses and \(\mathcal{R}_{\mathrm{LGG}}\) the subset exhibiting the LGG (Definition~\ref{def:gap}).
Answer accuracy (Acc), local soundness rate (LSR), and global sufficiency rate (GSR) are the proportions of correct, locally sound, and globally sufficient responses, respectively.
Trace reliability rate (TRR) is the proportion that is both locally sound and globally sufficient (\(L=G=1\)).
We further report \textit{LggShare}, the proportion of globally insufficient responses that remain locally sound, and \textit{LggRisk}, the proportion of answer errors that exhibit the LGG:
{\tightmath
\[
\mathrm{LggShare}
=
\frac{|\mathcal{R}_{\mathrm{LGG}}|}
{|\{y\in\mathcal{R}:G(y)=0\}|},
\qquad
\mathrm{LggRisk}
=
\frac{|\{y\in\mathcal{R}_{\mathrm{LGG}}:\mathrm{Acc}(y)=0\}|}
{|\{y\in\mathcal{R}:\mathrm{Acc}(y)=0\}|}.
\]
}TRR is the joint rate of local soundness and global sufficiency, not necessarily \(\mathrm{LSR}\times\mathrm{GSR}\).
Answer accuracy and global sufficiency assess different properties and are reported separately.

\subsection{Diagnostic Results}
\label{sec:diagnostic_findings}

\textbf{Nearly half of globally insufficient responses are locally sound.}
Figure~\ref{fig:diagnostic_results} summarizes the results across datasets and models.
Across \(2{,}598\) adjudicated responses, answer accuracy is \(81.4\%\), LSR is \(79.3\%\), GSR is \(88.0\%\), and TRR is \(73.7\%\).
Of the \(313\) globally insufficient responses, \(145\) are locally sound, yielding an LggShare of \(46.3\%\) (\(95\%\) Wilson CI: \([40.9\%,\,51.9\%]\)).
The reverse mismatch also occurs: \(371\) responses are locally unsound yet globally sufficient.
The LGG also appears in \(132\) of the \(484\) incorrect responses, yielding an LggRisk of \(27.3\%\) (\(95\%\) Wilson CI: \([23.5\%,\,31.4\%]\)).

\textbf{Path deviation and answer decoupling dominate the LGG.}
Among globally insufficient responses, \(190\) (\(60.7\%\)) exhibit path deviation and \(123\) (\(39.3\%\)) exhibit closure failure.
Answer decoupling accounts for \(87\) (\(70.7\%\)) of these closure failures.
Within the LGG, path deviation (\(51\) cases, \(35.2\%\)) and answer decoupling (\(69\) cases, \(47.6\%\)) together account for \(82.8\%\) of cases.
The remaining \(25\) cases (\(17.2\%\)) involve incomplete reasoning.
Thus, most LGG cases involve locally sound traces that pursue the wrong target or establish an answer different from the one submitted.

\textbf{Global sufficiency is distinct from answer correctness.}
Global sufficiency requires a question-aligned trace to establish the submitted answer, whereas correctness compares that answer with the reference.
In our human-adjudicated study, we find \(28\) correct but globally insufficient responses, whose traces fail alignment or closure, and \(199\) incorrect but globally sufficient responses.
The latter can arise when a trace addresses the question and establishes the submitted answer but relies on unsupported premises.
These failures of local soundness are assessed by \(L\) rather than \(G\). \looseness=-1

\textbf{Detector sensitivity follows the failed dependency.}
Table~\ref{tab:detector_blindness} compares detector sensitivity across failure types.
At thresholds retaining at least \(95\%\) of reliable traces, the conventional
faithfulness verifiers detect at most \(11.0\%\) of LGG cases, while the alignment and closure judges detect \(42.1\%\) and \(42.8\%\), respectively.
Their strengths are complementary: within the LGG, alignment detects \(66.7\%\) of path deviations, while closure detects \(56.5\%\) of answer-decoupling cases.
Each evidence-support verifier has higher recall on locally unsound responses than on LGG cases.
Together, these results support checking question alignment and answer closure alongside evidence support.

\begin{table}[t]
\centering
\small
\setlength{\tabcolsep}{2.8pt}
\renewcommand{\arraystretch}{1.12}
\caption{\textbf{Detector sensitivity across failure types.}
Each failure group is compared with the same \(1{,}914\) reliable traces.
AUC denotes AUROC (\(0.5\): random ranking; \(1.0\): perfect separation).
R@95 is failure recall (\%) at the highest empirical threshold that flags at most \(5\%\) of reliable traces.
Higher values are better for both metrics.
For evidence-support detectors, wavy and straight underlines mark recall on LGG and locally unsound responses, respectively.
Bold highlights alignment's sensitivity to path deviation and closure's sensitivity to answer decoupling.
Details appear in Appendix~\ref{app:detectors}.}
\label{tab:detector_blindness}
\vspace{4pt}

\begin{tabular}{@{}l*{10}{c}@{}}
\toprule
& \multicolumn{4}{c}{\(\mathcal D_q \to (r,a)\)}
& \multicolumn{2}{c}{\(\mathcal D_q \to r\)}
& \multicolumn{2}{c}{\(q \to r\)}
& \multicolumn{2}{c}{\(r \to a\)} \\
\cmidrule(lr){2-5}
\cmidrule(lr){6-7}
\cmidrule(lr){8-9}
\cmidrule(lr){10-11}
& \multicolumn{2}{c}{\textbf{RAGAS}}
& \multicolumn{2}{c}{\textbf{VeriScore}}
& \multicolumn{2}{c}{\textbf{AlignScore}}
& \multicolumn{2}{c}{\textbf{Alignment}}
& \multicolumn{2}{c}{\textbf{Closure}} \\
\cmidrule(lr){2-3}
\cmidrule(lr){4-5}
\cmidrule(lr){6-7}
\cmidrule(lr){8-9}
\cmidrule(lr){10-11}
\textbf{Trace group}
& AUC & R@95
& AUC & R@95
& AUC & R@95
& AUC & R@95
& AUC & R@95 \\
\midrule

\rowcolor[gray]{0.95}
LGG (\(n=145\))
& 0.679 & \uwave{11.0}
& 0.635 & \uwave{9.7}
& 0.469 & \uwave{6.2}
& 0.787 & 42.1
& 0.816 & 42.8 \\

\quad Path deviation (\(n=51\))
& 0.715 & 9.8
& 0.644 & 11.8
& 0.458 & 3.9
& 0.921 & \textbf{66.7}
& 0.694 & 13.7 \\

\quad Incomplete reasoning (\(n=25\))
& 0.680 & 20.0
& 0.593 & 4.0
& 0.448 & 0.0
& 0.844 & 60.0
& 0.885 & 64.0 \\

\quad Answer decoupling (\(n=69\))
& 0.651 & 8.7
& 0.643 & 10.1
& 0.486 & 10.1
& 0.667 & 17.4
& 0.881 & \textbf{56.5} \\

\midrule
\rowcolor[gray]{0.95}
Locally unsound (\(n=539\))
& 0.884 & \uline{40.8}
& 0.848 & \uline{39.5}
& 0.642 & \uline{10.9}
& 0.875 & 51.8
& 0.633 & 13.7 \\
\bottomrule
\end{tabular}
\end{table}

\subsection{Implications}
\label{sec:diagnostic_implications}

These findings show that locally sound traces can remain globally insufficient and that conventional faithfulness verifiers largely miss these failures.
Reliable reasoning therefore requires evidence-to-step support, question-to-trace alignment, and trace-to-answer closure.
This motivates \method{}, which combines generation supervision on supported original and counterfactual responses with bidirectional switching constraints to supervise all three dependencies during training (Section~\ref{sec:method}).

\section{\method{}: Edge Closure Learning}
\label{sec:method}

\begin{figure}[t]
\centering
\includegraphics[width=\linewidth]{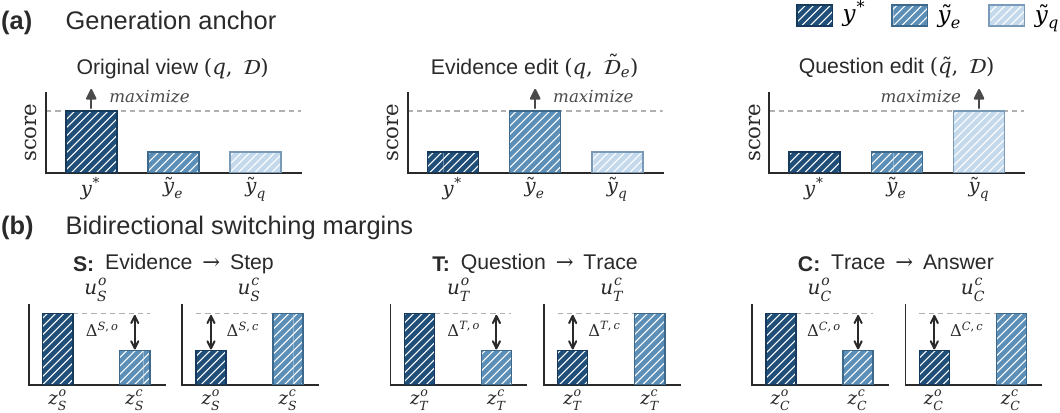}
\vspace{-15pt}
\caption{\textbf{Edge Closure Learning objective.} (a) Three generation anchors train the original and counterfactual responses under their supported inputs. (b) Six directional margins require preferences over the affected step, trace, or answer to reverse as the conditioning input switches between paired views. The \(S\), \(T\), and \(C\) panels instantiate \(\mathcal{D}_q\to s_{t(e)}\), \(q\to r\), and \(r\to a\), respectively.}
\label{fig:closure_objective}
\vspace{-10pt}
\end{figure}

\method{} supervises the three dependencies in Section~\ref{sec:diagnosis} through two complementary objectives.
Generation supervision uses one original and two counterfactual responses as anchors under their supported inputs.
Bidirectional margins require preferences over the affected step, trace, or answer to reverse when the conditioning input changes.
Figure~\ref{fig:closure_objective} summarizes the design.

\subsection{Intervention Views}
\label{sec:method_setup}
Each training instance contains a question \(q\), evidence \(\mathcal{D}\), and a target response \(\smash{y^\star=(r^\star,a^\star)}\) satisfying local soundness, alignment, closure, and answer correctness. We construct two additional supported responses. An evidence intervention replaces a supporting item \(e\) on which \(\smash{r^\star}\) depends with a matched counterfactual, yielding \(\smash{(\widetilde{\mathcal{D}}_e,\widetilde y_e)}\). The edited trace preserves the original prefix \(\smash{r^\star_{<t(e)}}\) and first diverges at \(t(e)\), the earliest step whose warrant depends on \(e\). A question intervention changes only the requested target while holding the evidence fixed, yielding \(\smash{(\widetilde q,\widetilde y_q)}\). These interventions define three paired views:
\[
S:\;(\mathcal{D},s^\star_{t(e)})\leftrightarrow(\widetilde{\mathcal{D}}_e,\widetilde s_{t(e),e})
\qquad
T:\;(q,r^\star)\leftrightarrow(\widetilde q,\widetilde r_q)
\qquad
C:\;(r^\star,a^\star)\leftrightarrow(\widetilde r_e,\widetilde a_e).
\]
For \(S\), \(q\) and the shared prefix remain fixed; for \(T\), \(\mathcal{D}\) remains fixed; and for \(C\), \(q\) and \(\mathcal{D}\) remain fixed while the trace changes. Thus, the evidence intervention supplies both \(S\) and \(C\), while the question intervention supplies \(T\). Full construction details appear in Appendix~\ref{app:training_construction}.

\subsection{Bidirectional Closure Objective}
\label{sec:closure_objective}

For each edge \(x\in\{S,T,C\}\), let \((u_x^{\mathrm{o}},z_x^{\mathrm{o}})\) and \((u_x^{\mathrm{c}},z_x^{\mathrm{c}})\) denote the conditioning input and supported dependent output in the original and counterfactual views, with shared context omitted. We score each output by its mean token log-likelihood, \(\ell_\theta(z\mid u)=\frac{1}{|z|}\log\pi_\theta(z\mid u)\). The two directional margins are
\[
\Delta_\theta^{x,\mathrm{o}}
=
\ell_\theta(z_x^{\mathrm{o}}\mid u_x^{\mathrm{o}})
-
\ell_\theta(z_x^{\mathrm{c}}\mid u_x^{\mathrm{o}}),
\qquad
\Delta_\theta^{x,\mathrm{c}}
=
\ell_\theta(z_x^{\mathrm{c}}\mid u_x^{\mathrm{c}})
-
\ell_\theta(z_x^{\mathrm{o}}\mid u_x^{\mathrm{c}}).
\]
The first measures preference for the original output under the original input, while the second reverses this preference under the counterfactual input. Requiring both margins to exceed the edge-specific target \(\gamma_x\) prevents a fixed preference from satisfying the objective.

Let \(\mathcal{L}_{\mathrm{gen}}\) be the mean generation loss over the three supported anchors, and let
\(h_\beta(t)=\beta\log(1+\exp(t/\beta))\)
denote the smooth hinge.
Define \(\mathcal{X}=\{S,T,C\}\) and
\(\mathcal{V}=\{\mathrm{o},\mathrm{c}\}\).
The full objective is
\[
\mathcal{L}_{\method{}}
=
\mathcal{L}_{\mathrm{gen}}
+
\lambda\,
\mathbb{E}_{(x,v)\sim
\operatorname{Unif}(\mathcal{X}\times\mathcal{V})}
\left[
h_{\beta_x}\!\left(
\gamma_x-\Delta_\theta^{x,v}
\right)
\right].
\]
The expectation averages uniformly over the six dependency--direction pairs, so \(\lambda\) controls their total weight relative to generation.
We use \(\lambda=1\) by default, with a shared target margin \(\gamma_x=0.5\) and hinge smoothing parameter \(\beta_x=0.1\) across all dependencies.
Separating \(T\) from \(C\) gives the answer a dedicated supervision signal, while \(S\) targets the first affected step.
Appendix~\ref{app:closure_objective_details} provides the conditioning views, expanded generation loss, and response factorization.

\subsection{Training and Inference}

Each training pack contains one accepted evidence intervention paired with the source's unique question intervention.
All generation losses and margin scores are computed with teacher forcing under the corresponding inputs.
The three anchors \(y^\star\), \(\widetilde y_e\), and \(\widetilde y_q\) serve as generation targets only under their supported inputs.
Swapped dependent outputs are used only in the corresponding margins and never as generation targets.
At inference, \method{} preserves the standard mapping \((q,\mathcal{D})\mapsto(r,a)\), with no verifier, additional model calls, or decoding changes.

\section{Experiments and Results}
\label{sec:experiments}

Our experiments address three questions:
\textbf{(1)} How do different training methods compare in answer accuracy and local and global reasoning quality? (Section~\ref{sec:main_results})
\textbf{(2)} How do individual dependency objectives and bidirectional supervision contribute to \method{}? (Section~\ref{sec:objective_ablations})
\textbf{(3)} Does \method{} switch its preferences as intended under dependency interventions? (Section~\ref{sec:dependency_interventions})

\subsection{Experimental Setup}
\label{sec:experimental_setup}

\textbf{Data and models.}
We use the distractor settings of 2WikiMultiHopQA, MuSiQue, and HotpotQA~\citep{ho-etal-2020-constructing,trivedi-etal-2022-musique,yang-etal-2018-hotpotqa}.
Evaluation uses the same questions from the official validation/test splits as the diagnostic study (Section~\ref{sec:diagnosis}).
For training, we sample sources from the official training splits and construct counterfactuals for each.
After validation, we split the sources \(9{:}1\) into training and development sets, containing \(2{,}566\) and \(286\) sources, respectively.
We use the same three backbones: Qwen3-8B, Qwen3-32B~\citep{yang2025qwen3technicalreport}, and Ministral-3-8B-Instruct-2512~\citep{liu2026ministral3}.
Each adaptation uses training seeds \(42\), \(43\), and \(44\).
Data construction and training details appear in Appendices~\ref{app:training_construction} and \ref{app:training_implementation}, respectively.

\textbf{Comparisons.} We compare \method{} with the untuned backbone, supervised fine-tuning (SFT) on supported original responses, counterfactual SFT (CF-SFT) on supported original and counterfactual responses, and response-margin training (RM). CF-SFT, RM, and \method{} share the same training data and generation anchors. RM adds bidirectional margins over complete responses (trace and final answer) under evidence and question interventions. Comparisons with CF-SFT and RM assess the value of margin supervision and dependency-specific supervision, respectively. For each backbone, all adapted methods use the same initialization, epoch budget, and checkpoint-selection rule. Objectives and hyperparameters appear in Appendices~\ref{app:objectives} and~\ref{app:training_implementation}.

\textbf{Evaluation.}
For semantic evaluation at scale, we use GPT-5.6 Sol~\citep{openai2026gpt56} with \texttt{high} reasoning effort to assess answer correctness, local soundness \(L\), alignment \(A\), closure \(C\), and reference completeness \(K\) under the frozen diagnostic guidelines.
We first apply the evaluator to the backbone's fixed diagnostic responses and measure agreement with human-adjudicated labels (Appendix~\ref{app:evaluator}).
Each adapted checkpoint then generates one response per diagnostic question using the shared structured CoT prompt and decoding protocol.
These responses are assessed with the same evaluator and guidelines.
A human audit covers \(90\) shared questions across the four fine-tuned methods using seed-\(42\) checkpoints (Appendix~\ref{app:experimental_human_audit}).
Deterministic matching accuracy appears in Appendix~\ref{app:matching_acc}.
We average adapted results over three seeds within each dataset--backbone setting, then average all methods equally across the nine settings.
We compute \(95\%\) paired hierarchical bootstrap confidence intervals by resampling backbones, questions within fixed datasets, and training seeds.
Draws are shared across methods, with seed draws also shared across datasets.

\subsection{Main Results}
\label{sec:main_results}

\begin{figure*}[t]
\centering
\includegraphics[width=\textwidth]{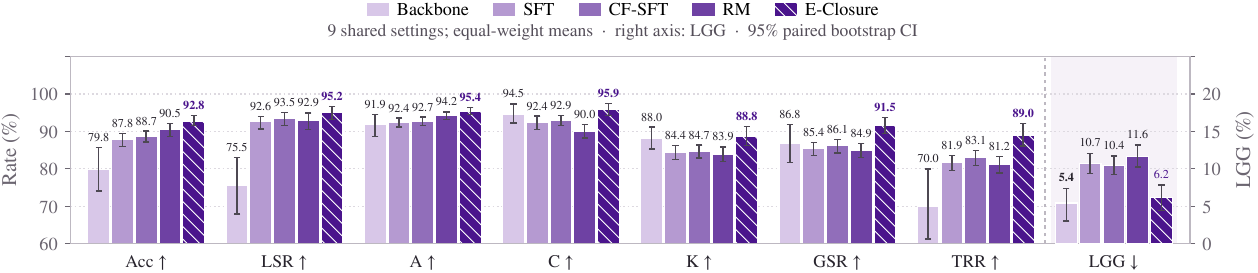}
\vspace{-20pt}
\caption{\textbf{Aggregate performance.}
Bars show equal-weight means across nine dataset--backbone settings, with adapted methods first averaged over three seeds within each setting.
Backbone uses fixed diagnostic generations.
Error bars show \(95\%\) hierarchical bootstrap confidence intervals with paired resampling across methods.
All metrics are percentages.}
\label{fig:overall_results}
\end{figure*}

\textbf{Local gains can accompany global degradation.}
Figure~\ref{fig:overall_results} shows that, on average, SFT and CF-SFT improve answer accuracy and LSR over the backbone while reducing closure and GSR.
Their LGG rates rise from the backbone's \(5.4\%\) to \(10.4\%\)--\(10.7\%\), showing that gains in answer accuracy and local soundness do not necessarily yield globally sufficient reasoning.

\textbf{\method{} improves local and global quality.}
\method{} achieves the highest average accuracy (\(92.8\%\)), LSR (\(95.2\%\)), GSR (\(91.5\%\)), and TRR (\(89.0\%\)) among the evaluated methods.
Its LGG rate is \(6.2\%\), lower than those of the evaluated fine-tuning baselines but remains above the backbone's (5.4\%).
Table~\ref{tab:per_setting_results} reports accuracy and TRR gains over CF-SFT in all nine settings and over RM in seven.
The largest TRR gains over RM occur on MuSiQue, ranging from \(15.9\) to \(26.1\) percentage points.
Here, GSR exceeds the backbone's for all three models, but the LGG rate remains higher on MuSiQue, indicating that improved global sufficiency does not fully eliminate the gap.

\textbf{Dependency-specific supervision improves on response-level margins.}
We compare \method{} with RM, which applies preference margins to complete responses.
\method{} increases average accuracy from \(90.5\%\) to \(92.8\%\) and TRR from \(81.2\%\) to \(89.0\%\), while reducing the LGG rate from \(11.6\%\) to \(6.2\%\).
These gains in both answer correctness and trace reliability support supervising individual dependencies rather than relying solely on response-level preferences.

\begin{table*}[t]
\centering
\fontsize{8.5}{9.5}\selectfont
\setlength{\tabcolsep}{2pt}
\renewcommand{\arraystretch}{1.12}
\vspace{-10pt}
\caption{\textbf{Performance by dataset and backbone.}
Adapted methods report means over three seeds.
All metrics are percentages: Acc.\(\uparrow\), LSR\(\uparrow\), GSR\(\uparrow\), TRR\(\uparrow\), and LGG\(\downarrow\).
Bold and underlining mark the best and second-best displayed values, respectively, for each metric within each dataset--backbone setting, including ties.
RM denotes Response-margin.
Per-seed results appear in Appendix~\ref{app:per_seed_results}.}
\label{tab:per_setting_results}
\vspace{4pt}

\begin{tabular*}{\textwidth}{@{\extracolsep{\fill}}l*{15}{c}@{}}
\toprule
\multirow{2}{*}{\textbf{Method}}
& \multicolumn{5}{c}{\textbf{2WikiMultiHopQA}}
& \multicolumn{5}{c}{\textbf{MuSiQue}}
& \multicolumn{5}{c}{\textbf{HotpotQA}} \\
\cmidrule(lr){2-6}
\cmidrule(lr){7-11}
\cmidrule(lr){12-16}
& \textbf{Acc.} & \textbf{LSR} & \textbf{GSR}
& \textbf{TRR} & \textbf{LGG}
& \textbf{Acc.} & \textbf{LSR} & \textbf{GSR}
& \textbf{TRR} & \textbf{LGG}
& \textbf{Acc.} & \textbf{LSR} & \textbf{GSR}
& \textbf{TRR} & \textbf{LGG} \\
\midrule

\multicolumn{16}{c}{\textit{Qwen3-8B}} \\
\addlinespace[2pt]

Backbone
& 92.4 & 95.7 & 94.2 & 90.6 & 5.0
& 55.3 & 43.1 & \underline{71.8} & 32.2 & \textbf{11.0}
& 91.9 & 88.9 & \underline{95.2} & 85.0 & \underline{3.9} \\

SFT
& 98.4 & 98.1 & 99.2 & 98.0 & \underline{0.1}
& 70.1 & 81.3 & 62.5 & 55.3 & 26.0
& 89.3 & \underline{94.3} & 90.4 & 86.6 & 7.7 \\

CF-SFT
& \underline{99.0} & 98.9 & \underline{99.4} & \underline{98.8} & \underline{0.1}
& 73.6 & \underline{82.4} & 62.9 & \underline{58.2} & 24.2
& 89.7 & 94.1 & 92.2 & 87.8 & 6.3 \\

RM
& \underline{99.0} & \underline{99.0} & \textbf{99.8} & \underline{98.8} & 0.2
& \underline{74.0} & 79.2 & 57.1 & 49.5 & 29.7
& \underline{93.6} & 93.0 & 93.1 & \underline{88.3} & 4.7 \\

\method{}
& \textbf{99.8} & \textbf{99.5} & \textbf{99.8} & \textbf{99.5} & \textbf{0.0}
& \textbf{81.2} & \textbf{84.2} & \textbf{72.9} & \textbf{65.9} & \underline{18.3}
& \textbf{95.3} & \textbf{96.3} & \textbf{96.7} & \textbf{93.6} & \textbf{2.7} \\

\addlinespace[3pt]
\cdashline{1-16}[3pt/2pt]
\addlinespace[3pt]
\multicolumn{16}{c}{\textit{Qwen3-32B}} \\
\addlinespace[2pt]

Backbone
& 96.4 & 96.4 & 99.6 & 96.0 & \underline{0.4}
& 65.9 & 58.4 & \underline{79.6} & 52.5 & \textbf{5.9}
& 94.9 & 94.6 & \underline{96.7} & 91.9 & \underline{2.7} \\

SFT
& 98.8 & 98.6 & \underline{99.9} & 98.6 & \textbf{0.0}
& 72.3 & 82.2 & 63.8 & 56.7 & 25.5
& 94.6 & \underline{98.4} & 93.8 & 92.7 & 5.7 \\

CF-SFT
& 98.8 & 98.7 & \textbf{100.0} & 98.7 & \textbf{0.0}
& 74.8 & \underline{85.8} & 67.2 & \underline{60.0} & 25.8
& 94.5 & 97.9 & 93.8 & 92.1 & 5.8 \\

RM
& \textbf{99.6} & \textbf{99.5} & \textbf{100.0} & \textbf{99.5} & \textbf{0.0}
& \underline{78.4} & 83.0 & 61.4 & 53.1 & 29.9
& \underline{96.8} & 97.5 & 96.3 & \underline{94.5} & 3.0 \\

\method{}
& \underline{99.0} & \underline{99.0} & \textbf{100.0} & \underline{99.0} & \textbf{0.0}
& \textbf{84.4} & \textbf{91.4} & \textbf{82.2} & \textbf{79.2} & \underline{12.2}
& \textbf{98.3} & \textbf{99.4} & \textbf{98.7} & \textbf{98.3} & \textbf{1.1} \\

\addlinespace[3pt]
\cdashline{1-16}[3pt/2pt]
\addlinespace[3pt]
\multicolumn{16}{c}{\textit{Ministral-3-8B}} \\
\addlinespace[2pt]

Backbone
& 84.2 & 91.0 & 87.8 & 80.9 & 10.1
& 49.4 & 30.2 & 65.5 & 24.7 & \textbf{5.5}
& 87.4 & 80.8 & 90.7 & 76.3 & 4.5 \\

SFT
& 99.5 & 99.5 & \textbf{100.0} & 99.5 & \textbf{0.0}
& 75.2 & 86.1 & 65.6 & 60.1 & 26.0
& 91.7 & 94.6 & 93.9 & 89.2 & 5.4 \\

CF-SFT
& 99.6 & 99.4 & \underline{99.9} & 99.4 & \textbf{0.0}
& 75.8 & 88.4 & \underline{66.5} & \underline{62.6} & 25.8
& 92.1 & 96.0 & 93.4 & 90.1 & 5.9 \\

RM
& \textbf{99.9} & \textbf{100.0} & \underline{99.9} & \textbf{99.9} & \underline{0.1}
& \underline{77.9} & \underline{88.5} & 61.7 & 55.7 & 32.8
& \underline{95.3} & \underline{96.1} & \underline{94.8} & \underline{91.9} & \underline{4.2} \\

\method{}
& \underline{99.8} & \underline{99.9} & \underline{99.9} & \underline{99.8} & \underline{0.1}
& \textbf{82.0} & \textbf{90.2} & \textbf{76.6} & \textbf{71.6} & \underline{18.6}
& \textbf{95.4} & \textbf{96.6} & \textbf{96.8} & \textbf{93.8} & \textbf{2.8} \\

\bottomrule
\end{tabular*}
\end{table*}

\subsection{Objective Ablations}
\label{sec:objective_ablations}

\setlength{\intextsep}{6pt}
\setlength{\columnsep}{12pt}

\begin{wrapfigure}{r}{0.5\textwidth}
\centering
\includegraphics[width=\linewidth]{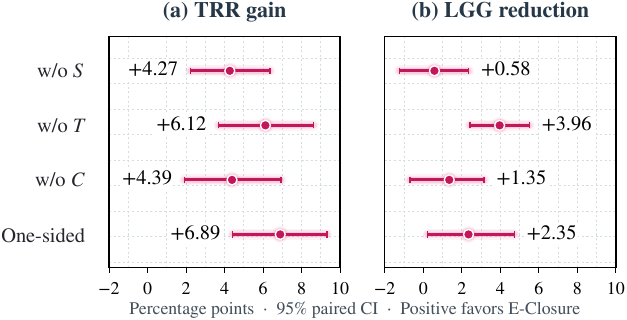}
\vspace{-20pt}
\caption{\textbf{TRR gains and LGG reductions with the full training objective.}
Positive values favor \method{} over each ablated variant.}
\label{fig:objective_ablation_gains}
\vspace{-5pt}
\end{wrapfigure}

On Qwen3-8B, we ablate \method{} by removing both directional margins for \(S\), \(T\), or \(C\) in turn.
A \emph{one-sided} variant retains only the original-condition margin for each dependency, dropping reverse preferences under counterfactual conditions.
All variants share the same training data, generation anchors, optimization budget, selection rule, and seeds.
Retained margin terms are scaled by \(3/2\) for dependency removals and \(2\) for the one-sided variant, preserving the total margin weight \(\lambda\).

\begin{table}[t]
\centering
\footnotesize
\setlength{\tabcolsep}{3.5pt}
\renewcommand{\arraystretch}{1.08}
\caption{\textbf{Objective ablations on Qwen3-8B.}
Percentages are pooled over all \(866\) questions per seed, then averaged across three seeds.
Bold marks the best value in each column.}
\label{tab:main_ablations}
\vspace{4pt}

\begin{tabular*}{\linewidth}{@{\extracolsep{\fill}}lccccccc@{}}
\toprule
\textbf{Variant}
& \textbf{Acc.} \(\uparrow\)
& \textbf{LSR} \(\uparrow\)
& \(\boldsymbol{A}\) \(\uparrow\)
& \(\boldsymbol{C}\) \(\uparrow\)
& \textbf{GSR} \(\uparrow\)
& \textbf{TRR} \(\uparrow\)
& \textbf{LGG} \(\downarrow\) \\
\midrule
\method{} w/o \(S\)
& 88.65 & 90.07 & 93.49 & 94.84 & 88.34 & 83.06 & 7.01 \\
\method{} w/o \(T\)
& 86.64 & 91.61 & 92.15 & 92.46 & 85.07 & 81.22 & 10.39 \\
\method{} w/o \(C\)
& 89.95 & 90.72 & 94.38 & 93.88 & 88.68 & 82.95 & 7.78 \\

\addlinespace[3pt]
\cdashline{1-8}[3pt/2pt]
\addlinespace[3pt]

\method{} one-sided
& 89.49 & 89.22 & 93.92 & 91.96 & 86.34 & 80.45 & 8.78 \\
\midrule
\method{}
& \textbf{92.57} & \textbf{93.76} & \textbf{95.30}
& \textbf{95.03} & \textbf{90.69} & \textbf{87.34}
& \textbf{6.43} \\
\bottomrule
\end{tabular*}
\end{table}

\textbf{Dependency objectives.}
The full objective achieves the best point estimates across all seven metrics (Table~\ref{tab:main_ablations}).
Removing any dependency objective reduces TRR by \(4.27\)--\(6.12\) percentage points, with the largest drop when removing \(T\).
All three paired TRR gains have confidence intervals above zero (Figure~\ref{fig:objective_ablation_gains}).
For LGG reduction, the interval lies above zero for w/o \(T\) but includes zero for w/o \(S\) and w/o \(C\).
These results support jointly supervising all three dependencies.

\textbf{Bidirectional supervision.}
Compared with the one-sided variant, the full objective improves TRR by \(6.89\) percentage points (\(95\%\) CI: \([4.43,\,9.31]\)) and reduces LGG by \(2.35\) points (\(95\%\) CI: \([0.23,\,4.73]\)).
Both intervals lie above zero, supporting the contribution of reverse preferences to improving trace reliability and reducing the LGG.
See Appendix~\ref{app:ablations} for details.

\subsection{Dependency Intervention Results}
\label{sec:dependency_interventions}

\begingroup
\setlength{\intextsep}{6pt}
\setlength{\columnsep}{12pt}
\begin{wraptable}{r}{0.5\textwidth}
\centering
\footnotesize
\setlength{\tabcolsep}{3pt}
\renewcommand{\arraystretch}{1.12}
\vspace{-15pt}
\caption{\textbf{Gains in dependency control.}
Switch-rate differences (\method{} minus each comparator), in percentage points.
Positive values favor \method{}.}
\label{tab:dependency_interventions}
\vspace{4pt}
\begin{tabular*}{\linewidth}{@{\extracolsep{\fill}}lrrrr@{}}
\toprule
\textbf{Dep.} & \textbf{Backbone} & \textbf{SFT} & \textbf{CF-SFT} & \textbf{RM} \\
\midrule
\(S\) & \(+1.67\) & \(+0.23\) & \(-0.03\) & \(0.00\) \\
\(T\) & \(+34.88\) & \(+5.05\) & \(+4.54\) & \(+4.99\) \\
\(C\) & \(+25.36\) & \(+13.37\) & \(+10.06\) & \(+69.59\) \\
\bottomrule
\end{tabular*}
\end{wraptable}

We test whether controlled changes in evidence (\(S\)), question target (\(T\)), and reasoning trace (\(C\)) switch the model's preferred step, trace, and answer, respectively.
We use intervention pairs from \(286\) development questions excluded from gradient updates.
Checkpoints are selected by answer accuracy on the same development split, without using intervention scores.
For each intervention pair, successful switching requires positive preference margins under both original and counterfactual conditions; we compute these directly from token-normalized model log-likelihoods without an LLM evaluator.
We first average rates over three seeds for adapted methods, then equally across nine dataset--backbone settings.

Table~\ref{tab:dependency_interventions} shows gains on both global dependencies over four comparators, with little change in \(S\) relative to CF-SFT and RM.
Gains over RM hold for \(T\) and \(C\) in all nine settings.
On \(C\), \method{} switches answer preferences on every evaluated pair, outperforming RM by \(69.59\) percentage points.
Here, \(C\) fixes the original evidence, creating evidence--trace conflict: \method{} rewards the edited trace's answer, whereas RM penalizes the counterfactual complete response.
Question-guided trace selection remains the main weakness, with switch rates of \(59.90\%\) on 2WikiMultiHopQA, \(94.64\%\) on MuSiQue, and \(98.69\%\) on HotpotQA.
Per-setting results appear in Appendix~\ref{app:interventions}.
\par
\endgroup

\section{Conclusion}
\label{sec:conclusion}

Our human-adjudicated diagnostic study reveals a local--global gap (LGG) in multi-hop reasoning: locally sound traces can fail to align with the question or establish the submitted answer.
However, the evaluated conventional faithfulness verifiers miss most of these failures.
Existing fine-tuning baselines improve accuracy and local soundness over the base models, but at the cost of global sufficiency.
To address this gap, we formalize three dependencies for reliable reasoning: evidence-to-step support, question-to-trace alignment, and trace-to-answer closure.
\textbf{\method{}} supervises these dependencies during training by combining supported generation anchors with bidirectional counterfactual constraints, without changing inference.
On average across three benchmarks and three backbones, it improves accuracy, local soundness, global sufficiency, and trace reliability while reducing the LGG rate relative to all evaluated fine-tuned baselines.
Although average global sufficiency gains over untuned backbones are modest and the average LGG rate remains higher, improvements over fine-tuning baselines support jointly evaluating and supervising all three dependencies.

\clearpage
\section*{Reproducibility Statement}
We document diagnostic sampling, annotation guidelines, and adjudication in Appendix~\ref{app:diagnosis}.
Counterfactual data construction and training objectives are detailed in Appendices~\ref{app:training_construction} and~\ref{app:closure_objective_details}, respectively.
Training hyperparameters, random seeds, checkpoint selection, and hardware are specified in Appendix~\ref{app:training_implementation}.
Reproducibility measures for generation and training include fixed random seeds, CUDA determinism settings, and a generation temperature of \(0\).
Evaluation protocols, human validation, and uncertainty estimation are described in Appendices~\ref{app:evaluator}, \ref{app:experimental_protocol}, and~\ref{app:full_results}.

\section*{AI Use Statement}

We used \texttt{ChatGPT} to polish the writing, with its role limited to language editing of author-written content.
We used \texttt{Cursor} with \texttt{Grok}-family models for coding assistance.
We used \texttt{GPT-5.6 Sol} to construct, screen, and verify counterfactual training data under stage-specific frozen guidelines.
Data construction and the subsequent human audit are detailed in Appendices~\ref{app:training_construction} and~\ref{app:packing_and_audit}, respectively.
For semantic evaluation at scale, \texttt{GPT-5.6 Sol} also served as the automated evaluator under a frozen evaluation guideline.
We assessed its agreement with human-adjudicated diagnostic labels (Appendix~\ref{app:evaluator}) and conducted a separate human audit of the main experimental results (Appendix~\ref{app:experimental_human_audit}).
All fine-tuning experiments used publicly available open-weight backbones (Appendix~\ref{app:generation}).
The authors developed the research ideas, designed and conducted the experiments, and reviewed and verified every sentence of the manuscript.
They take full responsibility for the manuscript, code, data, results, and conclusions.

\bibliography{iclr2027_conference, anthology-1, anthology-2}
\bibliographystyle{iclr2027_conference}

\appendix
\newpage

\section*{Appendix}
\vspace{10pt}

\noindent
\hyperref[app:diagnosis]{\textbf{A\quad Diagnostic Study: Additional Details}\dotfill\textbf{\pageref{app:diagnosis}}}\par
\vspace{6pt}

\hspace{1.8em}\hyperref[app:diagnostic_sample]{A.1\quad Diagnostic Sample\dotfill\pageref{app:diagnostic_sample}}\par
\vspace{3pt}

\hspace{1.8em}\hyperref[app:generation]{A.2\quad Generation Protocol\dotfill\pageref{app:generation}}\par
\vspace{3pt}

\hspace{1.8em}\hyperref[app:annotation]{A.3\quad Annotation Protocol\dotfill\pageref{app:annotation}}\par
\vspace{3pt}

\hspace{1.8em}\hyperref[app:failure_composition]{A.4\quad Failure Composition\dotfill\pageref{app:failure_composition}}\par
\vspace{3pt}

\hspace{1.8em}\hyperref[app:detectors]{A.5\quad Reference-Free Verification\dotfill\pageref{app:detectors}}\par

\vspace{8pt}

\noindent
\hyperref[app:training_construction]{\textbf{B\quad Training Data Construction: Additional Details}\dotfill\textbf{\pageref{app:training_construction}}}\par
\vspace{6pt}

\hspace{1.8em}\hyperref[app:training_sampling]{B.1\quad Sampling\dotfill\pageref{app:training_sampling}}\par
\vspace{3pt}

\hspace{1.8em}\hyperref[app:source_screening]{B.2\quad Source Screening\dotfill\pageref{app:source_screening}}\par
\vspace{3pt}

\hspace{1.8em}\hyperref[app:target_response_construction]{B.3\quad Target Response Construction\dotfill\pageref{app:target_response_construction}}\par
\vspace{3pt}

\hspace{1.8em}\hyperref[app:evidence_intervention]{B.4\quad Evidence Intervention\dotfill\pageref{app:evidence_intervention}}\par
\vspace{3pt}

\hspace{1.8em}\hyperref[app:question_intervention]{B.5\quad Question Intervention\dotfill\pageref{app:question_intervention}}\par
\vspace{3pt}

\hspace{1.8em}\hyperref[app:packing_and_audit]{B.6\quad Packing, Splitting, and Audit\dotfill\pageref{app:packing_and_audit}}\par

\vspace{8pt}

\noindent
\hyperref[app:closure_objective_details]{\textbf{C\quad Edge Closure Learning: Additional Details}\dotfill\textbf{\pageref{app:closure_objective_details}}}\par
\vspace{6pt}

\hspace{1.8em}\hyperref[app:closure_views]{C.1\quad Training Views and Validity\dotfill\pageref{app:closure_views}}\par
\vspace{3pt}

\hspace{1.8em}\hyperref[app:closure_scores]{C.2\quad Edge-Specific Scores\dotfill\pageref{app:closure_scores}}\par
\vspace{3pt}

\hspace{1.8em}\hyperref[app:closure_design]{C.3\quad Complete Objective and Design Rationale\dotfill\pageref{app:closure_design}}\par

\vspace{8pt}

\noindent
\hyperref[app:experiments]{\textbf{D\quad Experiments: Additional Details}\dotfill\textbf{\pageref{app:experiments}}}\par
\vspace{6pt}

\hspace{1.8em}\hyperref[app:evaluator]{D.1\quad Evaluator and Validation\dotfill\pageref{app:evaluator}}\par
\vspace{3pt}

\hspace{1.8em}\hyperref[app:experimental_protocol]{D.2\quad Experimental Evaluation Protocol\dotfill\pageref{app:experimental_protocol}}\par
\vspace{3pt}

\hspace{1.8em}\hyperref[app:objectives]{D.3\quad Comparison Objectives\dotfill\pageref{app:objectives}}\par
\vspace{3pt}

\hspace{1.8em}\hyperref[app:training_implementation]{D.4\quad Training Implementation and Compute\dotfill\pageref{app:training_implementation}}\par
\vspace{3pt}

\hspace{1.8em}\hyperref[app:full_results]{D.5\quad Extended Main Results\dotfill\pageref{app:full_results}}\par
\vspace{3pt}

\hspace{1.8em}\hyperref[app:ablations]{D.6\quad Objective Ablations\dotfill\pageref{app:ablations}}\par
\vspace{3pt}

\hspace{1.8em}\hyperref[app:interventions]{D.7\quad Dependency Interventions\dotfill\pageref{app:interventions}}\par
\vspace{3pt}

\newpage
\section{Diagnostic Study: Additional Details}
\label{app:diagnosis}

This appendix describes the diagnostic workflow in Section~\ref{sec:diagnosis}, from question sampling and response generation to eligibility screening and response annotation. It also provides annotation guidelines, agreement statistics, adjudication procedures, and detailed results for the reference-free detectors. Following the main text's notation, we denote the question by \(q\), its provided evidence by \(\mathcal{D}_q\), and the reference answer by \(a^\star\). A model response is \(y=(r,a)\), where \(r=(s_1,\ldots,s_T)\) is the reasoning trace, \(s_t\) is its \(t\)-th step, and \(a\) is the submitted answer. Other definitions appear in Section~\ref{sec:formulation}.

\begin{center}
\small
\textbf{Sampling, generation, and screening}
\[
\underbrace{\text{Sampled questions}}_{1{,}300}
\;\longrightarrow\;
\left.
\begin{array}{@{}l@{}}
\text{Generation complete: }1{,}291\\[2pt]
\text{Evidence eligible: }866
\end{array}
\right\}
\;\xrightarrow{\text{both pass}}\;
\underbrace{\text{Retained questions}}_{866}
\]
\textbf{Response annotation}

\smallskip
\setlength{\tabcolsep}{8pt}
\renewcommand{\arraystretch}{1.15}
\begin{tabular}{@{}cll@{}}
\toprule
\textbf{Pass} & \textbf{Judgment} & \textbf{Visible information} \\
\midrule
A & Substantive-step identification and soundness
& \(q,\mathcal{D}_q,r\) \\
B & Question alignment
& \(q,\mathcal{D}_q,r\) \\
C & Answer closure
& \(q,\mathcal{D}_q,r,a\) \\
D & Answer correctness and reference-answer support
& \(q,\mathcal{D}_q,r,a,a^\star\) \\
\bottomrule
\end{tabular}
\end{center}

\subsection{Diagnostic Sample}
\label{app:diagnostic_sample}

Using seed \(42\), we first sample \(300\) questions without replacement from the held-out evaluation split of each benchmark in the distractor setting: 2WikiMultiHopQA~\citep{ho-etal-2020-constructing}, MuSiQue~\citep{trivedi-etal-2022-musique}, and HotpotQA~\citep{yang-etal-2018-hotpotqa}. To obtain more balanced sample sizes across the three benchmarks after eligibility screening, we extend the MuSiQue and HotpotQA samples to \(600\) and \(400\) questions, respectively, yielding \(1{,}300\) questions in total. These extensions preserve the original seeded order and are determined without consulting model outputs or diagnostic labels.

\textbf{Sample characteristics.}
Table~\ref{tab:diagnostic_sample_stats} summarizes evidence composition and context length for the full sampled pool.
A passage is counted as supporting if it contains at least one benchmark-annotated supporting fact; all other provided passages are treated as distractors.
Support density is the proportion of provided passages that are supporting.
Context length counts whitespace-delimited words.

\begin{table}[h]
\centering
\small
\setlength{\tabcolsep}{2pt}
\renewcommand{\arraystretch}{1.12}
\vspace{-10pt}
\caption{\textbf{Diagnostic sample characteristics.}
Statistics cover the $1{,}300$ sampled questions.
Values are medians, with the 25th--75th percentiles in brackets.}
\label{tab:diagnostic_sample_stats}
\vspace{4pt}

\begin{tabular*}{\linewidth}{@{\extracolsep{\fill}}lcccc@{}}
\toprule
\textbf{Dataset}
& \shortstack{\textbf{Supporting passages}}
& \shortstack{\textbf{Distractor passages}}
& \shortstack{\textbf{Support density (\%)}}
& \shortstack{\textbf{Context words}} \\
\midrule
2WikiMultiHopQA
& 4 [4--4]
& 6 [6--6]
& 40 [40--40]
& 470 [365--641] \\
MuSiQue
& 3 [3--4]
& 17 [16--17]
& 15 [15--20]
& 1{,}671 [1{,}426--1{,}935] \\
HotpotQA
& 2 [2--2]
& 8 [8--8]
& 20 [20--20]
& 897 [733--1{,}065] \\
\bottomrule
\end{tabular*}
\end{table}

\subsection{Generation Protocol}
\label{app:generation}

We generate one completion for each question--model pair, yielding \(3{,}900\) initial generations.

\textbf{Prompt and input serialization.} 
All three models receive identical system and user content formatted with their official checkpoint chat templates. The shared structured CoT prompt requests a numbered reasoning trace followed by a submitted answer. The exact prompt and generation scripts are provided in the released codebase at \path{diagnostic/generation}. We serialize each normalized evidence passage with its original index, title, and text, preserving the dataset-provided order without retrieval, reranking, shuffling, or truncation. The longest serialized inputs contain \(6{,}041\) tokens under the Qwen3 tokenizer and \(6{,}076\) under the Ministral tokenizer. Including the \(1{,}024\)-token output budget, both remain within the \(8{,}192\)-token context limit.

\textbf{Models and generation configuration.} 
Table~\ref{tab:app-generation} reports checkpoint snapshots and runtime settings for the three models. We disable thinking mode for Qwen3-8B and Qwen3-32B~\citep{yang2025qwen3technicalreport} using \texttt{enable\_thinking=False}; Ministral-3-8B-Instruct-2512~\citep{liu2026ministral3} has no thinking mode. All checkpoints are loaded with \texttt{dtype=auto}: Qwen3 uses bfloat16, while Ministral uses the released FP8 weights with float16 computation. Greedy decoding produces one completion per input under a fixed configuration, avoiding sampling variation in the diagnostic pool.

\begin{table}[t]
    \centering
    \footnotesize
    \caption{\textbf{Generation configuration.} Hashes identify the snapshots referenced by \texttt{refs/main} in the Hugging Face cache at generation time.}
    \label{tab:app-generation}
    \vspace{4pt}

    \begin{tblr}{
        width   = \linewidth,
        colspec = {
            Q[l,m,wd=0.10\linewidth]
            Q[l,m,wd=0.19\linewidth]
            X[l,m]
        },
        colsep    = 5pt,
        rowsep    = 3pt,
        row{1}    = {font=\bfseries},
        column{1} = {leftsep=0pt},
        column{3} = {rightsep=0pt},
    }
        \toprule
        Category & Field & Configuration \\
        \midrule

        \SetCell[r=3]{l,m} Models
        & Qwen3-8B
        & {
            \nolinkurl{Qwen/Qwen3-8B} \\
            \nolinkurl{b968826d9c46dd6066d109eabc6255188de91218} \\
            bfloat16; thinking disabled
        } \\

        & Qwen3-32B
        & {
            \nolinkurl{Qwen/Qwen3-32B} \\
            \nolinkurl{9216db5781bf21249d130ec9da846c4624c16137} \\
            bfloat16; thinking disabled; eager execution
        } \\

        & Ministral-3-8B
        & {
            \nolinkurl{mistralai/Ministral-3-8B-Instruct-2512} \\
            \nolinkurl{5b26027e7b19eeb4b7352e1fed3926375dd2cb4d} \\
            FP8 weights; float16 computation; no thinking mode
        } \\

        \midrule
        \SetCell[r=2]{l,m} Input
        & Prompt
        & Identical system and user content and demonstration order;
          official checkpoint chat templates \\

        & Evidence
        & Normalized passages with original indices, titles, and texts;
          original passage order; no retrieval, reranking, shuffling,
          or truncation \\

        \midrule
        \SetCell[r=2]{l,m} Decoding
        & Generation
        & Greedy decoding; temperature \(0\); top-\(p=1.0\);
          maximum \(1{,}024\) new tokens; one completion per input \\

        & Context and order
        & \texttt{max\_model\_len=8192};
          inputs processed in lexicographic \texttt{uid} order \\

        \midrule
        \SetCell[r=2]{l,m} Runtime
        & Engine
        & vLLM 0.8.5.post1 (V0); tensor parallelism \(4\);
          FlashAttention backend \\

        & Hardware
        & \(4\times\) NVIDIA RTX A6000 \\

        \midrule
        \SetCell[r=2]{l,m} Software
        & Core
        & Python 3.11.15; PyTorch 2.6.0+cu124; CUDA 12.4;
          Transformers 4.51.3 \\

        & Supporting
        & Tokenizers 0.21.4; xFormers 0.0.29.post2; Triton 3.2.0;
          cuDNN 9.1.0.70; NCCL 2.21.5 \\

        \bottomrule
    \end{tblr}
\end{table}

\textbf{Output parsing and incomplete generations.} 
For each generation, we retain the raw completion, parsed trace and answer, model identifier, and decoding metadata. The parser locates the first line matching \texttt{Final answer:}, taking the preceding text as the reasoning trace and the text after the marker on that line as the submitted answer. Of the \(3{,}900\) generations, \(3{,}888\) contain exactly one matching answer line and \(3\) contain two; all \(3{,}891\) yield nonempty parsed answers. The remaining \(9\) reach the output limit without a matching answer line: \(8\) from Qwen3-8B and \(1\) from Qwen3-32B. Requiring complete generations from all three models yields a shared pool of \(1{,}291\) questions and \(3{,}873\) responses, keeping the question sets matched across models for subsequent analyses.

\subsection{Annotation Protocol}
\label{app:annotation}

Eligibility screening and response annotation use separate annotator pools. Two screeners (S1 and S2) independently assess eligibility using only the question, provided evidence, and reference answer. Six annotators (A1--A6) label the \(2{,}598\) responses to the \(866\) retained questions, with two independent annotations per response. Adjudicators S3 and A7 resolve screening and response-level disagreements, respectively. All diagnostic analyses use the final labels after adjudication.

\subsubsection{Question-Level Eligibility Screening}
\label{app:eligibility_screening}

We screen questions before response annotation to avoid attributing defects in the evaluation instance to model reasoning. Eligibility is assessed once per question, without access to model responses.

\textbf{Screeners and criteria.} 
Two screeners (S1 and S2) independently assess all \(1{,}300\) questions under frozen guidelines. They inspect only the question, full evidence context, and reference answer. Model identities, responses, benchmark supporting-fact indicators, and the other screener's decisions are hidden. A question receives \textsc{Yes} only if it is well posed, the evidence supports a unique answer up to semantic equivalence, and the reference matches that answer. Each \textsc{No} decision receives one primary reason from Table~\ref{tab:eligibility_reasons}. If multiple reasons apply, screeners select the first applicable category in the fixed order \(\mathrm{N4}\rightarrow\mathrm{N1}\rightarrow\mathrm{N2}\rightarrow\mathrm{N3}\rightarrow\mathrm{N5}\).

\begin{table}[h]
\centering
\footnotesize
\setlength{\tabcolsep}{4pt}
\renewcommand{\arraystretch}{1}
\renewcommand{\tabularxcolumn}[1]{m{#1}}
\caption{\textbf{Primary reasons for question-level ineligibility.}
Each \textsc{No} decision receives exactly one reason under the stated precedence rule.}
\label{tab:eligibility_reasons}
\vspace{4pt}

\begin{tabularx}{\linewidth}{
@{}
>{\centering\arraybackslash}m{0.07\linewidth}
>{\raggedright\arraybackslash}m{0.37\linewidth}
>{\raggedright\arraybackslash}X
@{}
}
\toprule
\textbf{Code} & \textbf{Reason} & \textbf{Operational criterion} \\
\midrule
N1
& Missing required evidence
& A fact needed to answer the question is absent from the provided evidence. \\
\addlinespace[2pt]

N2
& Missing evidential link
& Relevant facts are present, but a required entity, relation, or inferential connection is not established. \\
\addlinespace[2pt]

N3
& Underdetermined or conflicting evidence
& The evidence supports multiple incompatible answers or contains a conflict that cannot be resolved. \\
\addlinespace[2pt]

N4
& Ambiguous or ill-posed question
& The question's referent, scope, requested relation, or expected answer type cannot be reliably determined. \\
\addlinespace[2pt]

N5
& Incompatible reference answer
& The evidence warrants a unique answer, but the reference is not semantically equivalent to it. \\
\bottomrule
\end{tabularx}
\end{table}

\textbf{Pre-adjudication agreement.}
We assess independent screening records using raw agreement and nominal Krippendorff's \(\alpha\)~\citep{krippendorff2004reliability}. Binary eligibility has \(91.0\%\) raw agreement and \(\alpha=0.783\). Among the \(322\) questions labeled \textsc{No} by both screeners, primary-reason agreement is \(79.2\%\), with \(\alpha=0.702\). We estimate \(95\%\) percentile confidence intervals from \(2{,}000\) paired question-level bootstrap resamples with seed \(42\). For primary-reason agreement, we resample only questions labeled \textsc{No} by both screeners. Tables~\ref{tab:eligibility_agreement} and~\ref{tab:eligibility_agreement_dataset}
report overall agreement and dataset-specific binary eligibility agreement, respectively.

\begin{table}[h]
\centering
\small
\setlength{\tabcolsep}{4pt}
\renewcommand{\arraystretch}{1.12}
\caption{\textbf{Pre-adjudication agreement on eligibility screening.} Brackets show \(95\%\) bootstrap CIs. Primary reasons are compared only when both screeners select \textsc{No}.}
\label{tab:eligibility_agreement}
\vspace{4pt}

\begin{tabular*}{\linewidth}{@{\extracolsep{\fill}}lccc@{}}
\toprule
\textbf{Judgment}
& \(\boldsymbol{n}\)
& \shortstack{\textbf{Raw agreement (\%) [95\% CI]}}
& \shortstack{\(\boldsymbol{\alpha}\) [95\% CI]} \\
\midrule
Binary eligibility
& 1,300
& 91.0 [89.4, 92.5]
& 0.783 [0.745, 0.817] \\

Primary ineligibility reason
& 322
& 79.2 [74.5, 83.5]
& 0.702 [0.640, 0.764] \\
\bottomrule
\end{tabular*}
\end{table}

\begin{table}[h]
\centering
\small
\setlength{\tabcolsep}{4pt}
\renewcommand{\arraystretch}{1.12}
\caption{\textbf{Pre-adjudication eligibility agreement by dataset.} Brackets show \(95\%\) bootstrap CIs for \(\alpha\). The final column counts questions labeled \textsc{No} by both screeners.}
\label{tab:eligibility_agreement_dataset}
\vspace{4pt}

\begin{tabular*}{\linewidth}{@{\extracolsep{\fill}}lcccc@{}}
\toprule
\textbf{Dataset}
& \(\boldsymbol{n}\)
& \shortstack{\textbf{Raw agreement (\%)}}
& \shortstack{\(\boldsymbol{\alpha}\) [95\% CI]}
& \textbf{Both \textsc{No}} \\
\midrule
2WikiMultiHopQA
& 300 & 97.3 & 0.804 [0.653, 0.924] & 18 \\

MuSiQue
& 600 & 86.0 & 0.720 [0.664, 0.773] & 262 \\

HotpotQA
& 400 & 93.8 & 0.735 [0.628, 0.830] & 42 \\
\bottomrule
\end{tabular*}
\end{table}

\textbf{Adjudication and final outcomes.}
A third screener (S3) resolves all eligibility disagreements and primary-reason disagreements on questions labeled \textsc{No} by both screeners. S3 follows the same frozen guidelines, reviewing \(q\), \(\mathcal{D}_q\), \(a^\star\), and both screening records. In total, \(184\) questions undergo adjudication: \(117\) with eligibility disagreements and \(67\) with reason-only disagreements. Screening retains \(866\) questions and excludes \(434\) (Table~\ref{tab:eligibility_adjudication}). Missing evidential links (N2) and ambiguous or ill-posed questions (N4) are the most common exclusion reasons.

\begin{table}[!h]
\centering
\small

\setlength{\tabcolsep}{4pt}
\setlength{\dashlinedash}{3pt}
\setlength{\dashlinegap}{2pt}
\renewcommand{\arraystretch}{1.12}

\caption{\textbf{Final eligibility outcomes.}
Eligibility rates use all 1,300 sampled questions;
exclusion-reason rates use the 434 ineligible questions.
Codes N1--N5 are defined in Table~\ref{tab:eligibility_reasons}.}
\label{tab:eligibility_adjudication}
\vspace{4pt}

\begin{tabularx}{\linewidth}{
@{}l
*{2}{>{\centering\arraybackslash}X}
:
*{5}{>{\centering\arraybackslash}X}
@{}
}
\toprule

& \multicolumn{2}{c:}{\textbf{Eligibility}}
& \multicolumn{5}{c}{\textbf{Exclusion reasons}} \\

\cmidrule(lr){2-3}
\cmidrule(lr){4-8}

\textbf{Statistic}
& \textbf{Eligible} & \textbf{Ineligible}
& \textbf{N1} & \textbf{N2} & \textbf{N3} & \textbf{N4} & \textbf{N5} \\

\cmidrule(r{6pt}){1-3}
\cmidrule(l{6pt}){4-8}

Count
& 866 & 434
& 65 & 165 & 15 & 155 & 34 \\

Rate
& 66.6\% & 33.4\%
& 15.0\% & 38.0\% & 3.5\% & 35.7\% & 7.8\% \\

\bottomrule
\end{tabularx}

\end{table}

\textbf{Scope of the retained sample.}
Screening retains \(278\) of \(300\) 2WikiMultiHopQA questions, \(255\) of \(600\) MuSiQue questions, and \(333\) of \(400\) HotpotQA questions. The larger initial samples for MuSiQue and HotpotQA partly offset their higher exclusion rates, yielding more comparable retained sample sizes. Prevalence estimates in Section~\ref{sec:diagnosis} therefore characterize the evidence-eligible subset retained for the final diagnostic analysis rather than the full benchmark splits. Differences across datasets may reflect both task difficulty and which questions pass screening.

\subsubsection{Response-Level Annotation Rules}
\label{app:response_annotation_rules}

All \(2{,}598\) responses are independently labeled by two of six annotators (A1--A6) under frozen guidelines. For each question, the three model responses are assigned to disjoint annotator pairs, ensuring that each annotator sees at most one response to that question. Model identities and the other two responses are hidden. Task orders are independently shuffled using annotator-specific seeds derived from seed \(42\). A separate adjudicator (A7) resolves disagreements under the same guidelines.

\textbf{Pass A: substantive status and local soundness.}
Each nonempty line of the parsed trace \(r\) becomes a fixed step after any leading numeric prefix is removed. Annotators first classify each step as substantive or non-substantive according to its function. A step is substantive if removing it would eliminate a factual claim, inference, computation, evidence-coverage judgment, or entity resolution. Restatements of \(q\), tautologies, plans, headings, transitions, and empty scaffolding are non-substantive unless they also perform one of these functions. Both the submitted answer \(a\) and reference answer \(a^\star\) remain hidden throughout this pass.

Local soundness \(\ell_t\) is assessed only for substantive steps under the frozen guidelines. A step is sound (\(\ell_t=1\)) if every factual claim is warranted by \(\mathcal{D}_q\) and every inference follows from \(\mathcal{D}_q\) or preceding steps. Annotators may use general knowledge to understand the text, but not to supply facts unsupported by the evidence. Support may come from any passage in \(\mathcal{D}_q\), regardless of citation indices. Annotators assess each claim as a whole, including its subject, and judge hedged statements by what they assert in the context of the trace. Any unwarranted claim or inference makes the entire step unsound. Each unsound step receives one primary failure code from Table~\ref{tab:local_taxonomy}.

\begin{table}[h]
\centering
\small
\setlength{\tabcolsep}{5pt}
\renewcommand{\arraystretch}{1.12}
\caption{\textbf{Primary reasons for local unsoundness.}}
\label{tab:local_taxonomy}
\vspace{4pt}

\begin{tabularx}{\linewidth}{@{}
c
>{\raggedright\arraybackslash}p{0.32\linewidth}
>{\raggedright\arraybackslash}X
@{}}

\toprule
\textbf{Code} & \textbf{Failure type} & \textbf{Annotation criterion} \\
\midrule
\textbf{L1}
& Unsupported factual content
& States a factual claim that is not warranted by \(\mathcal{D}_q\). \\
\addlinespace[4pt]

\textbf{L2}
& Invalid inference
& Makes an inference, including an entity link, that does not follow from
\(\mathcal{D}_q\) or preceding sound steps. \\
\addlinespace[4pt]

\textbf{L3}
& Constraint or computation error
& Applies a constraint incorrectly or makes an error in calculation,
comparison, or ordering. \\
\addlinespace[4pt]

\textbf{L4}
& False evidence-coverage judgment
& Misjudges what \(\mathcal{D}_q\) establishes, such as claiming insufficient
evidence when the evidence is sufficient. \\
\addlinespace[4pt]

\textbf{L5}
& Other (rare)
& Contains another local soundness failure not covered by L1--L4. \\
\bottomrule
\end{tabularx}
\end{table}

\textbf{Pass B: alignment.}
Annotators assess whether \(r\) pursues the target requested by \(q\), preserving the specified entity, relation, constraints, and answer type. Relevant intermediate subproblems remain aligned even when the trace stops short of an answer or leaves facts uncombined. Missing premises or unfinished inferences alone do not imply misalignment. A trace is misaligned when its main line of reasoning pursues another entity or relation, drops a constraint, changes the answer type, or follows a distractor chain. Alignment is judged independently of local soundness: a trace may pursue the correct target despite containing unsupported claims or invalid inferences. An abstention is aligned if it addresses the requested target and concludes that the evidence does not determine it. The interface displays \(r\) as a single block without step-level controls; both \(a\) and \(a^\star\) remain hidden.

\textbf{Pass C: closure.}
The submitted answer \(a\) is revealed, while \(a^\star\) remains hidden. Annotators assess whether \(r\) supports \(a\) as a conclusion for the target pursued by the trace. Whether that target matches \(q\) is assessed separately by alignment, so an off-target trace can still satisfy closure. The final-answer line is excluded from \(r\). The trace's claims are taken as premises without reassessing their local soundness; facts appearing only in \(\mathcal{D}_q\) cannot fill gaps. Directly implied inferences are permitted, but a missing premise or unresolved alternative yields \(C(y)=0\). Mere non-contradiction is insufficient. An abstention is closed only if the trace supports the conclusion that its target cannot be resolved. A false claim of insufficient evidence may be locally unsound under L4 yet satisfy closure.

\textbf{Pass D: answer correctness and completeness.}
The reference answer \(a^\star\) is revealed, and annotators assess answer correctness and completeness. The submitted answer \(a\) is correct if it matches \(a^\star\), allowing aliases, spelling variants, fuller names, and semantic equivalents without conflicting content. Abstentions are incorrect because every retained question has a warranted answer.

Completeness \(K(y)\) assesses whether the trace's stated claims, taken as true, suffice to establish \(a^\star\). An unstated inference is permitted if it follows from these claims. A missing premise yields \(K(y)=0\), even if that premise appears in \(\mathcal{D}_q\). As with closure, local soundness is not reassessed, so a locally unsound trace may still satisfy \(K(y)=1\).

\textbf{Derived labels and consistency checks.} Local soundness is \(L=1\) if and only if every substantive step has \(\ell_t=1\), while global sufficiency is \(G=AC\). A response exhibits the LGG when \(L=1\) and \(G=0\). Globally insufficient responses are classified according to the first broken global link:
{\footnotesize
\[
\begin{aligned}
A=0
&\quad\Rightarrow\quad \text{\textsc{PathDeviation}},\\
A=1,\ C=0,\ K=0
&\quad\Rightarrow\quad \text{\textsc{IncompleteReasoning}},\\
A=1,\ C=0,\ K=1
&\quad\Rightarrow\quad \text{\textsc{AnswerDecoupling}}.
\end{aligned}
\]
}Any response labeled \(L=A=C=1\) but \(\mathrm{Acc}(y)=0\) is flagged for adjudication.

\subsubsection{Agreement and Adjudication}
\label{app:annotation_agreement_section}

\begin{table}[t]
\centering
\small
\setlength{\tabcolsep}{3.5pt}
\renewcommand{\arraystretch}{1.12}
\caption{\textbf{Inter-annotator agreement before adjudication.} \(n\) counts doubly annotated steps or responses; Maj.\ denotes the pooled prevalence of the most frequent label. Brackets show \(95\%\) bootstrap CIs for \(\alpha\). Specific agreement is reported for majority/minority labels and omitted for the multiclass local-failure reason.}
\label{tab:annotation_agreement}
\vspace{4pt}

\begin{tabular*}{\linewidth}{@{\extracolsep{\fill}}lccccc@{}}
\toprule
\textbf{Judgment}
& \(\boldsymbol{n}\)
& \textbf{Maj. (\%)}
& \textbf{Raw agr. (\%)}
& \(\boldsymbol{\alpha}\) \textbf{[95\% CI]}
& \textbf{Specific agr.} \\
\midrule
\multicolumn{6}{@{}c}{\textit{Direct annotations}} \\
\addlinespace[2pt]
Substantive status
& 12,733 & 95.1 & 99.9
& 0.989 [0.983, 0.994] & 0.999 / 0.990 \\

Step soundness \(\ell_t\)
& 12,101 & 90.0 & 95.1
& 0.726 [0.706, 0.744] & 0.973 / 0.754 \\

Local-failure reason
& 910 & 36.3 & 82.0
& 0.754 [0.721, 0.787] & --- \\

Alignment \(A\)
& 2,598 & 94.0 & 97.1
& 0.739 [0.677, 0.796] & 0.984 / 0.755 \\

Closure \(C\)
& 2,598 & 94.1 & 97.7
& 0.797 [0.739, 0.845] & 0.988 / 0.809 \\

Answer correctness
& 2,598 & 81.4 & 99.7
& 0.990 [0.982, 0.996] & 0.998 / 0.992 \\

Completeness \(K\)
& 2,598 & 88.3 & 95.0
& 0.759 [0.715, 0.800] & 0.972 / 0.787 \\

\addlinespace[2pt]
\cdashline{1-6}[3pt/2pt]
\addlinespace[2pt]

\multicolumn{6}{@{}c}{\textit{Derived labels}} \\
\addlinespace[2pt]
Local soundness \(L\)
& 2,598 & 74.5 & 89.9
& 0.735 [0.702, 0.767] & 0.932 / 0.802 \\

Global sufficiency \(G\)
& 2,598 & 88.5 & 95.0
& 0.754 [0.709, 0.795] & 0.972 / 0.782 \\

LGG indicator
& 2,598 & 93.5 & 97.0
& 0.756 [0.694, 0.809] & 0.984 / 0.772 \\
\bottomrule
\end{tabular*}
\end{table}

\textbf{Pre-adjudication agreement.}
We measure raw agreement and nominal Krippendorff's \(\alpha\)~\citep{krippendorff2004reliability} between the two primary annotations. We estimate \(95\%\) percentile confidence intervals from \(2{,}000\) question-level bootstrap resamples with seed \(42\), keeping each question's three responses, steps, and paired annotations together. Step-soundness agreement includes only steps both annotators consider substantive; local-failure-reason agreement further requires both to label the step unsound. We derive \(L\), \(G\), and the LGG indicator from each annotator's judgments.

Table~\ref{tab:annotation_agreement} summarizes agreement for each variable. Across directly annotated binary variables, raw agreement ranges from \(95.0\%\) to \(99.9\%\), and \(\alpha\) from \(0.726\) to \(0.990\). For the multiclass local-failure reason, raw agreement across \(910\) steps is \(82.0\%\), with \(\alpha=0.754\). Minority-label specific agreement across binary variables ranges from \(0.754\) to \(0.992\).

\textbf{Adjudication.}
A seventh annotator (A7) adjudicates every response with a primary-label disagreement or a consistency-check violation, following the same four-pass order and staged information disclosure as primary annotation. All derived labels are then recomputed from the adjudicated judgments. In total, \(823\) responses (\(31.7\%\)) undergo adjudication: \(819\) have primary-label disagreements, and \(4\) trigger only the consistency check. The remaining \(1{,}775\) responses have matching primary labels, pass the consistency check, and are merged without adjudication.

Trigger categories may overlap, and all rates use the \(2{,}598\) responses as the denominator. Step-level disagreements affect \(595\) responses (\(22.9\%\)) for soundness, \(142\) (\(5.5\%\)) for local-failure reason, and \(13\) (\(0.5\%\)) for substantive status. Response-level disagreements affect \(129\) responses (\(5.0\%\)) for completeness, \(76\) (\(2.9\%\)) for alignment, \(59\) (\(2.3\%\)) for closure, and \(8\) (\(0.3\%\)) for answer correctness. The consistency check flags \(6\) responses (\(0.2\%\)), including the \(4\) consistency-only cases. Step soundness offers more opportunities for disagreement because it is judged for every substantive step, and a single disputed step triggers adjudication of the response.

\subsection{Failure Composition}
\label{app:failure_composition}

Section~\ref{sec:diagnostic_findings} examines the LGG, where every substantive step is locally sound but the response is globally insufficient. Table~\ref{tab:dependency_failure_composition} situates these cases within the full joint distribution of local soundness \(L\), question-to-trace alignment \(A\), and trace-to-answer closure \(C\).

\begin{table}[h]
\centering
\small
\setlength{\tabcolsep}{5pt}
\renewcommand{\arraystretch}{1.08}
\caption{\textbf{Joint distribution of dependency violations.}
Rows enumerate the \(L\), \(A\), and \(C\) states computed from the final merged annotations, where \(1\) denotes a satisfied dependency and \(0\) a violation.
Shares use all \(2{,}598\) responses as the denominator and are rounded to one decimal place.}
\vspace{5pt}
\label{tab:dependency_failure_composition}
\begin{tabularx}{\linewidth}{@{}
>{\raggedright\arraybackslash}X
>{\centering\arraybackslash}m{0.07\linewidth}
>{\centering\arraybackslash}m{0.07\linewidth}
>{\centering\arraybackslash}m{0.07\linewidth}
>{\centering\arraybackslash}m{0.13\linewidth}
>{\centering\arraybackslash}m{0.15\linewidth}
@{}}
\toprule
\textbf{Observed violation}
& \(\boldsymbol{L}\) & \(\boldsymbol{A}\) & \(\boldsymbol{C}\)
& \(\boldsymbol{n}\) & \textbf{Share (\%)} \\
\midrule
None & 1 & 1 & 1 & \(1{,}914\) & 73.7 \\
Local unsoundness only & 0 & 1 & 1 & \(371\) & 14.3 \\
Alignment only & 1 & 0 & 1 & \(47\) & 1.8 \\
Closure only & 1 & 1 & 0 & \(94\) & 3.6 \\
Local unsoundness and alignment & 0 & 0 & 1 & \(132\) & 5.1 \\
Local unsoundness and closure & 0 & 1 & 0 & \(29\) & 1.1 \\
Alignment and closure & 1 & 0 & 0 & \(4\) & 0.2 \\
All three & 0 & 0 & 0 & \(7\) & 0.3 \\
\midrule
Total & & & & \(2{,}598\) & 100.0 \\
\bottomrule
\end{tabularx}
\end{table}

\textbf{Most failures affect a single dependency.}
Of the \(684\) responses with at least one violation, \(512\) (\(74.9\%\)) violate exactly one dependency and \(172\) (\(25.1\%\)) violate multiple dependencies. Across all \(2{,}598\) responses, local soundness fails in \(539\) (\(20.7\%\)), alignment in \(190\) (\(7.3\%\)), and closure in \(134\) (\(5.2\%\)); \(11\) responses fail both global links. Under the first-broken-link convention in Section~\ref{sec:formulation}, the \(190\) responses with \(A=0\) are classified as path deviation, while the \(123\) with \(A=1\) and \(C=0\) are classified as closure failures.

\textbf{Local soundness and global sufficiency are distinct.}
Of the \(539\) locally unsound responses, \(371\) satisfy both global dependencies, while \(168\) also fail globally. Conversely, the \(145\) LGG responses are locally sound but globally insufficient: \(51\) fail at alignment, and \(94\) remain aligned but fail at closure. The latter comprise \(25\) incomplete reasoning cases and \(69\) answer decoupling cases. Together, path deviation and answer decoupling account for \(82.8\%\) of the LGG.

\textbf{Connection to training and interventions.}
These labels describe whether the dependencies hold in observed responses. The \(S\), \(T\), and \(C\) objectives supervise the corresponding evidence-to-step, question-to-trace, and trace-to-answer relations, with each objective targeting a distinct dependent output. S-Switch, T-Switch, and C-Switch assess whether model preferences switch as intended when these relations are tested through counterfactual interventions (Section~\ref{sec:dependency_interventions}).

\textbf{Composition of local failures.}
A response with \(L=0\) contains at least one locally unsound substantive step. Table~\ref{tab:local_failure_distribution} reports the primary reason codes for the \(936\) unsound steps in the final merged annotations. Each step contributes one primary reason, so responses with multiple unsound steps contribute multiple records. Unsupported factual content and false evidence-coverage judgments together account for \(640\) steps (\(68.4\%\)), followed by invalid inferences (\(196\), \(20.9\%\)) and constraint or computation errors (\(100\), \(10.7\%\)). No step is assigned to the residual L5 category.

\begin{table}[h]
\centering
\small
\setlength{\tabcolsep}{7pt}
\renewcommand{\arraystretch}{1.12}
\vspace{-10pt}
\caption{\textbf{Composition of local-failure reasons.}
Counts and shares cover the 936 substantive steps labeled locally unsound in the final merged annotations.}
\label{tab:local_failure_distribution}
\vspace{4pt}

\begin{tabular*}{0.75\linewidth}{@{\extracolsep{\fill}}lcccccc@{}}
\toprule
\textbf{Statistic}
& \textbf{L1} & \textbf{L2} & \textbf{L3}
& \textbf{L4} & \textbf{L5} & \textbf{Total} \\
\midrule
Count
& 391 & 196 & 100 & 249 & 0 & 936 \\
Share (\%)
& 41.8 & 20.9 & 10.7 & 26.6 & 0.0 & 100.0 \\
\bottomrule
\end{tabular*}
\end{table}

\providecommand{\detci}[3]{%
  \shortstack{#1\\[-1pt]{\scriptsize\textcolor{black!55}{[#2,\,#3]}}}%
}

\subsection{Reference-Free Verification}
\label{app:detectors}

We examine whether verification can detect global failures even when every reasoning step is locally sound. We evaluate five reference-free detectors covering the three dependencies defined in Section~\ref{sec:formulation}: evidence support, question alignment, and answer closure. All detectors score the same \(2{,}598\) responses to \(866\) questions without reference answers. Evaluation groups are defined by the final merged annotations. AlignScore uses substantive-step annotations to select the steps it scores.

\subsubsection{Evaluation Setup}
\label{app:detector_setup}

\textbf{Trace groups.}
We partition the \(2{,}598\) responses into three disjoint groups based on their local soundness \(L\) and global sufficiency \(G\):

\begin{center}
\small
\renewcommand{\arraystretch}{1.15}
\begin{tabular*}{0.5\linewidth}{@{\extracolsep{\fill}}llr@{}}
\toprule
\textbf{Group} & \textbf{Condition} & \textbf{Count} \\
\midrule
Reliable \(\mathcal{R}_{\mathrm{rel}}\) & \(L=1,\ G=1\) & 1,914 \\
LGG \(\mathcal{R}_{\mathrm{LGG}}\) & \(L=1,\ G=0\) & 145 \\
Locally unsound \(\mathcal{R}_{\mathrm{unsound}}\) & \(L=0\) & 539 \\
\bottomrule
\end{tabular*}
\end{center}

Within the LGG, \(51\) responses exhibit path deviation, while \(94\) fail at answer closure: \(25\) show incomplete reasoning and \(69\) exhibit answer decoupling.

\textbf{Scores and metrics.} We compare each failure group with the reliable responses in \(\mathcal{R}_{\mathrm{rel}}\). Scores range from \(0\) to \(1\), and responses scoring strictly below a threshold are flagged as failures. \textbf{AUROC} is the probability that a failed response scores below a reliable one, with ties receiving half credit (\(0.5\): random ranking; \(1.0\): perfect separation). For \textbf{R@95}, we select the highest empirical threshold that incorrectly flags at most \(5\%\) of reliable responses, then report the percentage of failures detected at that threshold. Higher values indicate better performance for both metrics.

\textbf{Confidence intervals.}
We report \(95\%\) percentile confidence intervals from \(2{,}000\) question-level bootstrap resamples with seed \(42\). For each group mean, we sample questions with replacement from those represented in the group. Each time a question is drawn, all its responses within that group are included together. Section~\ref{app:detector_robustness} describes the bootstrap procedure for AUROC and R@95.

\textbf{Detector implementations.} Table~\ref{tab:reference_free_metric_impl} summarizes detector inputs and scoring rules. RAGAS~\citep{es-etal-2024-ragas} originally uses \texttt{gpt-3.5-turbo-16k}. VeriScore~\citep{song-etal-2024-veriscore} uses fine-tuned Mistral-7B-Instruct-v0.2 and Llama3-8B-Instruct models for claim extraction and verification, respectively, in its main experiments. Our RAGAS-style and VeriScore-style implementations use \texttt{gpt-5.6-luna}~\citep{openai2026gpt56luna} as a more recent evaluator backbone. Both assess the complete response, including the submitted answer, against the provided evidence without external retrieval; we abbreviate these adaptations as RAGAS and VeriScore below. AlignScore~\citep{zha-etal-2023-alignscore} uses the official \texttt{AlignScore-large} checkpoint, based on RoBERTa-large, to assess evidence support for annotated substantive steps. The alignment judge uses \texttt{gpt-5.6-luna} to assess whether the trace pursues the question's target. The closure detector uses a T5-XXL entailment model~\citep{honovich-etal-2022-true} to assess whether the trace establishes the submitted answer to the question. Both global detectors withhold the evidence to isolate their respective dependencies.

All detector backbones belong to different model families from the diagnostic generators. This choice aims to reduce potential bias toward responses from the evaluator's own model family.

\begin{table}[h]
\centering
\small
\setlength{\tabcolsep}{5pt}
\renewcommand{\arraystretch}{1.12}
\renewcommand{\tabularxcolumn}[1]{m{#1}}
\caption{\textbf{Detector inputs and scoring rules.}
Higher scores indicate stronger support for the evaluated dependency.
No detector receives the reference answer.}
\label{tab:reference_free_metric_impl}
\vspace{4pt}

\begin{tabularx}{\linewidth}{
@{}
>{\raggedright\arraybackslash}m{0.27\linewidth}
>{\raggedright\arraybackslash}m{0.13\linewidth}
>{\raggedright\arraybackslash}X
@{}
}
\toprule
\textbf{Detector} & \textbf{Dependency} & \textbf{Implementation} \\
\midrule
RAGAS~\citep{es-etal-2024-ragas}
& \(\mathcal{D}_q\to(r,a)\)
& Decompose the response into statements and report the fraction supported by the evidence. \\
\addlinespace[3pt]

VeriScore~\citep{song-etal-2024-veriscore}
& \(\mathcal{D}_q\to(r,a)\)
& Extract atomic factual claims and report the fraction supported by the evidence. \\
\addlinespace[3pt]

AlignScore~\citep{zha-etal-2023-alignscore}
& \(\mathcal{D}_q\to r\)
& Official v0.1.3 implementation with \texttt{AlignScore-large} and \texttt{nli\_sp}. Take the maximum passage score for each annotated substantive step, then average over steps. \\
\addlinespace[3pt]

Alignment judge
& \(q\to r\)
& Given only \(q\) and \(r\), assess whether the trace pursues the requested target and constraints. Use \(P(\textsc{Yes})\) from the first nonempty output token's log probabilities. \\
\addlinespace[3pt]

Closure detector
& \(r\to a\)
& Use \texttt{google/t5\_xxl\_true\_nli\_mixture}~\citep{honovich-etal-2022-true}, with \(r\) as the premise and ``the answer to \(q\) is \(a\)'' as the hypothesis. Report the binary entailment probability without providing the evidence. This question-conditioned score is a proxy for closure and may also reflect alignment. \\
\bottomrule
\end{tabularx}
\end{table}

\subsubsection{Dependency-Specific Detector Sensitivity}
\label{app:detector_results}

\begin{figure}[t]
\centering
\includegraphics[width=\linewidth]{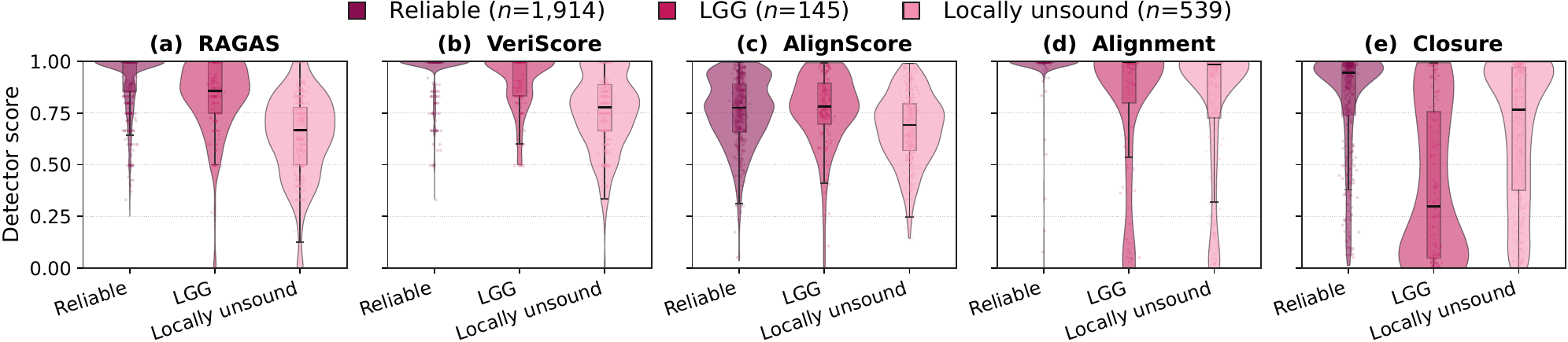}
\vspace{-15pt}
\caption{\textbf{Detector scores across response groups.}
Reliable responses are locally sound and globally sufficient; LGG responses are locally sound but globally insufficient; locally unsound responses contain at least one unsound substantive step. Higher scores indicate stronger support for the dependency checked by each detector. Violins show separately normalized distributions, boxes show interquartile ranges with median lines, and dots show up to 400 sampled responses per group.}
\label{fig:detector_distributions}
\end{figure}

\begin{figure}[t]
\centering
\includegraphics[width=\linewidth]{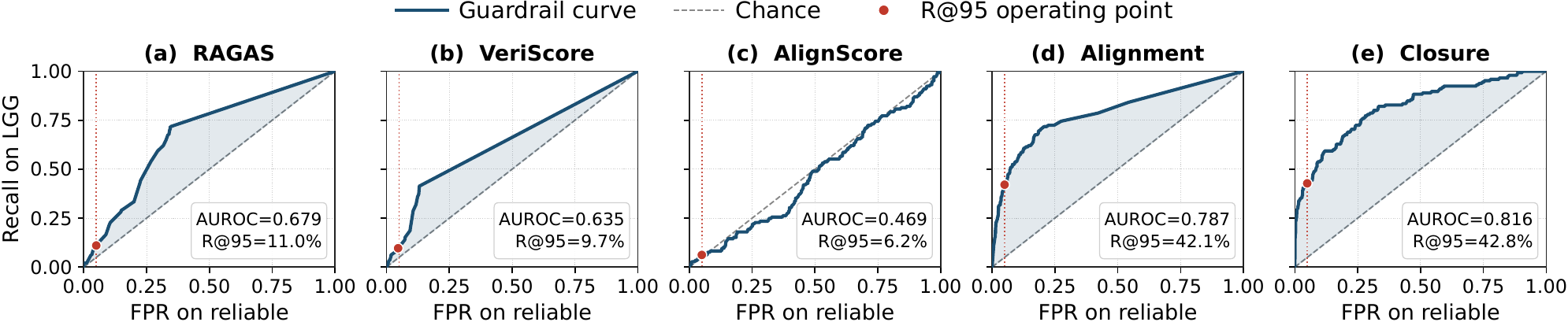}
\vspace{-15pt}
\caption{\textbf{LGG detection versus false alarms.}
The x-axis shows the false-positive rate (FPR): the fraction of reliable responses incorrectly flagged as failures. The y-axis shows recall: the fraction of LGG responses detected. Curves vary the detection threshold, with scores below the threshold flagged as failures. Red dots mark R@95: recall at the most permissive threshold with FPR at most \(5\%\). Dashed diagonals indicate chance performance; \((1,1)\) means every response is flagged.}
\label{fig:threshold_sweep}
\end{figure}

\textbf{Evidence-support detectors miss most LGG cases.} Table~\ref{tab:detector_scores} and Figure~\ref{fig:detector_distributions} show that RAGAS and VeriScore assign lower scores on average to locally unsound responses than to LGG responses. AlignScore performs near chance in distinguishing LGG from reliable responses (AUROC \(0.469\)). At R@95, the evaluated evidence-support detectors detect only \(6.2\%\)--\(11.0\%\) of LGG cases, while the alignment and closure detectors detect \(42.1\%\) and \(42.8\%\), respectively (Table~\ref{tab:detector_blindness}; Figure~\ref{fig:threshold_sweep}).

\textbf{Alignment and closure detect complementary failures.}
Within the LGG, alignment detects \(66.7\%\) of path deviations at R@95, compared with \(13.7\%\) for closure. For answer decoupling, closure detects \(56.5\%\), compared with \(17.4\%\) for alignment and \(8.7\%\)--\(10.1\%\) for the evidence-support detectors. Closure also detects \(64.0\%\) of incomplete reasoning cases. These results support explicit checks of both question-to-trace alignment and trace-to-answer closure.

\begin{table}[h]
\centering
\small
\setlength{\tabcolsep}{3pt}
\renewcommand{\arraystretch}{1.12}
\renewcommand{\tabularxcolumn}[1]{m{#1}}
\caption{\textbf{Detector scores by trace group.}
Each cell shows the mean above its 95\% question-level bootstrap confidence interval.
The indented rows partition the LGG.}
\label{tab:detector_scores}
\vspace{4pt}

\begin{tabularx}{\linewidth}{
@{}l r *{5}{>{\centering\arraybackslash}X}@{}
}
\toprule
& & \multicolumn{3}{c}{Evidence support}
& \(q\to r\) & \(r\to a\) \\
\cmidrule(lr){3-5}
\cmidrule(lr){6-6}
\cmidrule(lr){7-7}
\textbf{Trace group}
& \(\boldsymbol{n}\)
& \textbf{RAGAS}
& \textbf{VeriScore}
& \textbf{AlignScore}
& \textbf{Alignment}
& \textbf{Closure} \\
\midrule
Reliable
& 1,914
& \detci{0.918}{0.910}{0.925}
& \detci{0.966}{0.961}{0.972}
& \detci{0.756}{0.746}{0.767}
& \detci{0.984}{0.978}{0.989}
& \detci{0.812}{0.798}{0.826} \\
\midrule

LGG
& 145
& \detci{0.830}{0.800}{0.857}
& \detci{0.909}{0.886}{0.929}
& \detci{0.768}{0.737}{0.797}
& \detci{0.801}{0.742}{0.858}
& \detci{0.402}{0.340}{0.464} \\
\addlinespace[3pt]

\hspace{0.6em}Path deviation
& 51
& \detci{0.806}{0.747}{0.858}
& \detci{0.904}{0.864}{0.940}
& \detci{0.781}{0.737}{0.821}
& \detci{0.640}{0.534}{0.739}
& \detci{0.616}{0.520}{0.705} \\
\addlinespace[3pt]

\hspace{0.6em}Incomplete reasoning
& 25
& \detci{0.810}{0.739}{0.873}
& \detci{0.925}{0.875}{0.969}
& \detci{0.793}{0.739}{0.843}
& \detci{0.712}{0.554}{0.855}
& \detci{0.270}{0.137}{0.423} \\
\addlinespace[3pt]

\hspace{0.6em}Answer decoupling
& 69
& \detci{0.854}{0.813}{0.891}
& \detci{0.908}{0.872}{0.940}
& \detci{0.749}{0.697}{0.796}
& \detci{0.952}{0.902}{0.991}
& \detci{0.291}{0.205}{0.378} \\
\midrule

Locally unsound
& 539
& \detci{0.634}{0.615}{0.652}
& \detci{0.756}{0.739}{0.772}
& \detci{0.675}{0.658}{0.692}
& \detci{0.796}{0.767}{0.825}
& \detci{0.660}{0.627}{0.691} \\
\bottomrule
\end{tabularx}
\end{table}

\subsubsection{Operating-Point Robustness}
\label{app:detector_robustness}

R@95 depends on a threshold estimated from reliable responses. We assess metric uncertainty and whether calibrated thresholds maintain low false-positive rates and preserve sensitivity differences on held-out questions.

\textbf{Uncertainty and threshold transfer.} For AUROC and R@95, we resample the \(866\) questions with replacement, keeping each question's three responses together. The R@95 threshold is recalibrated per resample. This preserves dependence within questions and accounts for threshold uncertainty.

For threshold transfer, we form five fixed folds stratified by dataset, using seed \(42\). For each fold, we calibrate the R@95 threshold on reliable responses from the other four folds and apply it unchanged to held-out questions. Each question is evaluated once using a threshold calibrated on other questions. We pool held-out false positives and detections across folds. For confidence intervals, we repeat calibration and evaluation in each bootstrap resample with fixed fold assignments.

Table~\ref{tab:detector_robustness} reports cross-fitted false-positive rates of \(4.5\%\)--\(5.1\%\). Alignment and closure each retain LGG recall of \(42.8\%\), compared with \(6.2\%\)--\(10.3\%\) for the evidence-support detectors. The sensitivity pattern therefore persists on questions excluded from threshold calibration. However, each global detector still misses more than half of LGG cases at this operating point.

\begin{table}[!h]
\centering
\small
\setlength{\tabcolsep}{4pt}
\renewcommand{\arraystretch}{1.12}
\renewcommand{\tabularxcolumn}[1]{m{#1}}
\caption{\textbf{Uncertainty and threshold transfer for LGG detection.} Estimates appear above their \(95\%\) question-level bootstrap CIs. CF denotes five-fold cross-fitting; FPR is the false-positive rate on reliable responses. All values except AUROC are percentages.}
\label{tab:detector_robustness}
\vspace{4pt}

\begin{tabularx}{\linewidth}{
@{}l *{4}{>{\centering\arraybackslash}X}@{}
}
\toprule
\textbf{Detector}
& \textbf{AUROC}
& \textbf{R@95}
& \textbf{CF FPR}
& \textbf{CF recall} \\
\midrule
RAGAS
& \detci{0.679}{0.636}{0.722}
& \detci{11.0}{4.6}{15.9}
& \detci{4.5}{3.9}{5.1}
& \detci{10.3}{4.8}{15.5} \\
\addlinespace[3pt]

VeriScore
& \detci{0.635}{0.594}{0.675}
& \detci{9.7}{4.5}{14.7}
& \detci{4.6}{3.4}{4.9}
& \detci{9.7}{4.5}{14.8} \\
\addlinespace[3pt]

AlignScore
& \detci{0.469}{0.418}{0.524}
& \detci{6.2}{2.2}{10.3}
& \detci{5.1}{4.8}{5.5}
& \detci{6.2}{2.2}{10.4} \\
\addlinespace[3pt]

Alignment
& \detci{0.787}{0.739}{0.831}
& \detci{42.1}{32.4}{51.0}
& \detci{5.0}{4.7}{5.3}
& \detci{42.8}{32.4}{51.1} \\
\addlinespace[3pt]

Closure
& \detci{0.816}{0.774}{0.854}
& \detci{42.8}{33.8}{51.6}
& \detci{5.0}{4.8}{5.4}
& \detci{42.8}{33.8}{51.3} \\
\bottomrule
\end{tabularx}
\end{table}

\section{Training Data Construction: Additional Details}
\label{app:training_construction}

\textbf{Overview.}
This appendix describes the construction and validation of the shared data pool for \method{} and the fine-tuned baselines. Each family contains three generation anchors (Gold, Evidence-CF, and Question-CF) and their associated intervention pairs, as illustrated in Table~\ref{tab:comparison_example}. We retain \(2{,}852\) families; Table~\ref{tab:training_construction_flow} reports retention by dataset. A deterministic, dataset-stratified \(90/10\) split based on SHA-256 ranking yields \(2{,}566\) training families and \(286\) development families. Section~\ref{app:packing_and_audit} details packing, splitting, and quality auditing.

\[
\underbrace{
\text{Sample}
\rightarrow
\text{Screen}
\rightarrow
y^\star
\rightarrow
\text{Interventions}
\rightarrow
\text{Pack}
\rightarrow
\text{Split}
\rightarrow
\text{Audit}
}_{\text{\small training data construction}}
\]

\begin{table}[h]
\centering
\small
\setlength{\tabcolsep}{4pt}
\renewcommand{\arraystretch}{1.12}
\caption{\textbf{Example of training data construction.}
The original response and two counterfactual responses form three generation anchors. The evidence intervention changes a supporting fact, while the question intervention changes the requested target.}
\vspace{4pt}
\label{tab:comparison_example}
\begin{tabularx}{\linewidth}{
@{}
>{\raggedright\arraybackslash}m{0.15\linewidth}
>{\centering\arraybackslash}m{0.15\linewidth}
>{\raggedright\arraybackslash}X
>{\centering\arraybackslash}m{0.12\linewidth}
@{}
}
\toprule
\textbf{View} & \textbf{Input} & \textbf{Warranted trace} & \textbf{Answer} \\
\midrule
Original
& \((q,\mathcal D)\)
& \(s_1\): \(N\) was written by Ada. \(s_2\): Ada was born in Paris, France. \(s_3\): The requested country is France.
& France \\
Evidence edit
& \((q,\widetilde{\mathcal D}_e)\)
& \(s_1\): \(N\) was written by Ada. \(s_2\): Ada was born in Berlin, Germany. \(s_3\): The requested country is Germany.
& Germany \\
Question edit
& \((\widetilde q,\mathcal D)\)
& \(s_1\): \(N\) was written by Ada. \(s_2\): Ada was born in Paris, France. \(s_3\): The requested city is Paris.
& Paris \\
\bottomrule
\end{tabularx}
\end{table}

\textbf{Constructors and verifiers.} 
All construction, screening, and verification stages use GPT-5.6 Sol~\citep{openai2026gpt56} under separate frozen guidelines. Screens assess whether proposed inputs and targets are admissible without seeing generated traces. Each trace verifier returns the required semantic judgments as separate structured fields in a single call. The guidelines define each criterion independently, restrict the information that may justify it, and prohibit inferring one judgment from another. We conduct a final human audit of the constructed data, as detailed in Section~\ref{app:packing_and_audit}. The frozen guidelines are available in the released code at \path{construction/}. Sections~\ref{app:training_sampling}--\ref{app:packing_and_audit} detail the pipeline from source sampling to packing and human audit.

\begin{table}[t]
\centering
\small
\setlength{\tabcolsep}{4pt}
\renewcommand{\arraystretch}{1.12}
\caption{\textbf{Training data construction.}
Counts are unique source questions.
Processed denotes entry into construction; Screened denotes passing the initial screen.
The \(y^\star\), Evidence, and Question columns require verified original responses, evidence interventions, and question interventions, respectively.
Packed counts sources retained after semantic reconstruction and deterministic auditing.}
\label{tab:training_construction_flow}
\vspace{4pt}

\begin{tabular*}{\linewidth}{@{\extracolsep{\fill}}lccccccr@{}}
\toprule
\textbf{Dataset}
& \textbf{Sampled}
& \textbf{Processed}
& \textbf{Screened}
& \(\boldsymbol{y^\star}\)
& \textbf{Evidence}
& \textbf{Question}
& \textbf{Packed} \\
\midrule

2WikiMultiHopQA
& 5,000 & 2,000 & 1,841 & 1,841 & 1,622 & 1,622 & 1,276 \\

MuSiQue
& 15,000 & 15,000 & 2,827 & 2,827 & 1,685 & 946 & 559 \\

HotpotQA
& 10,000 & 6,000 & 2,608 & 2,608 & 1,862 & 1,189 & 1,017 \\

\midrule
Total
& 30,000
& 23,000
& 7,276
& 7,276
& 5,169
& 3,757
& 2,852 \\

\bottomrule
\end{tabular*}
\end{table}

\subsection{Sampling}
\label{app:training_sampling}

We sample without replacement from the official training splits, using seed \(42\) for subsampling, and deduplicate questions by source identifier. The sampled pool contains \(5{,}000\) questions from 2WikiMultiHopQA~\citep{ho-etal-2020-constructing}, \(15{,}000\) from MuSiQue~\citep{trivedi-etal-2022-musique}, and \(10{,}000\) from HotpotQA~\citep{yang-etal-2018-hotpotqa} (Table~\ref{tab:training_construction_flow}, \emph{Sampled} column). We sample and process more MuSiQue and HotpotQA questions to offset their higher screening exclusion rates and improve balance in the retained data, consistent with the diagnostic screening results (Section~\ref{app:eligibility_screening}). Of these, \(23{,}000\) unique questions enter construction (\emph{Processed} column): \(2{,}000\) from 2WikiMultiHopQA, \(15{,}000\) from MuSiQue, and \(6{,}000\) from HotpotQA. Eligibility is assessed only during source screening, which accounts for most exclusions in MuSiQue and HotpotQA (\emph{Processed} to \emph{Screened}).

\subsection{Source Screening}
\label{app:source_screening}

\textbf{Seed screen.} 
Source screening checks question validity and evidence sufficiency before trace or counterfactual construction, preventing later generation from masking defects in the original source. Let \(z=(q,\mathcal{D},a^\star)\) denote a processed source, where \(a^\star\) is the reference answer. The \textsc{screen-seed} stage receives only \(z\) and selects a support set \(E\subseteq\mathcal{D}\). Let \(W(q,\mathcal{D},a)\) indicate whether \(\mathcal{D}\) warrants \(a\) as an answer to \(q\), and \(U(q,\mathcal{D},a)\) whether it warrants \(a\) uniquely. We retain \(z\) only if
\[
|E|\geq 2
\land W(q,E,a^\star)
\land U(q,\mathcal{D},a^\star)
\land \neg W(q,\mathcal{D}\setminus E,a^\star).
\]
The selected support set must contain at least two passages and establish the reference answer. The full context must warrant that answer uniquely, and the remaining passages must not independently establish it. We also reject sources with malformed or underdetermined questions, broken entity links, unsupported reference answers, or impractically long answers.

\textbf{Retention.} 
Source screening retains \(7{,}276\) of the \(23{,}000\) processed questions (\(31.6\%\)): \(1{,}841/2{,}000\) for 2WikiMultiHopQA, \(2{,}827/15{,}000\) for MuSiQue, and \(2{,}608/6{,}000\) for HotpotQA (Table~\ref{tab:training_construction_flow}, \emph{Screened} column).

\subsection{Target Response Construction}
\label{app:target_response_construction}

\textbf{Target construction.} 
Given \((q,\mathcal{D},a^\star,E)\), the constructor generates a candidate response \(\widehat{y}=(\widehat{r},\widehat{a})\) and a completion indicator \(Q^\star\). The trace has no fixed length. Each step states one locally supported substantive claim, and the trace pursues the target specified by \(q\), making all required connections between dependent steps explicit. All evidence-dependent premises must be grounded in \(E\), although the trace need not use every passage in \(E\); no required premise may rely solely on \(\mathcal{D}\setminus E\). The submitted answer \(\widehat{a}\) must match \(a^\star\) under the fixed normalization.

\textbf{Target verification.}
The reference-blind \textsc{verify-target} verifier receives \((q,\mathcal{D},E,\widehat{r},\widehat{a})\), without access to \(a^\star\), and returns four separately defined judgments: local soundness \(L\), question--trace alignment \(A\), trace--answer closure \(C\), and evidence grounding \(R_E\). Each judgment uses only its designated inputs: \(L\) uses \((q,\mathcal{D},\widehat{r})\), \(A\) uses \((q,\widehat{r})\), and \(C\) uses \((\widehat{r},\widehat{a})\). For closure, neither the question nor the evidence may supply missing trace premises. \(R_E\) requires a complete step-level support map and rejects any necessary premise supported only outside \(E\). Semantic verification succeeds when all four criteria hold:
\[
V^\star =
L(q,\mathcal{D},\widehat{r})
\land A(q,\widehat{r})
\land C(\widehat{r},\widehat{a})
\land R_E(\widehat{r}).
\]
A deterministic gate \(H^\star\) checks normalized answer matching, support-identifier membership, schema validity, and structural well-formedness. We retain only candidates satisfying \(V^\star \land Q^\star \land H^\star\), denoting each accepted response by \(y^\star\).

\textbf{Retention.}
All \(7{,}276\) screened sources yield an accepted target response: \(1{,}841\) from 2WikiMultiHopQA, \(2{,}827\) from MuSiQue, and \(2{,}608\) from HotpotQA (Table~\ref{tab:training_construction_flow}, \(y^\star\) column).

\subsection{Evidence Intervention}
\label{app:evidence_intervention}

An evidence intervention changes one answer-determining claim while keeping the question fixed. The trace must first diverge where the claim is used and establish the new answer.

\textbf{Evidence construction.}
Given \((q,\mathcal{D},E,y^\star)\), the constructor replaces one passage \(e\in E\) with a counterfactual passage \(\widetilde e\), yielding
\[
\widetilde{\mathcal{D}}_e
=
(\mathcal{D}\setminus\{e\})\cup\{\widetilde e\},
\qquad
\widetilde y_e=(\widetilde r_e,\widetilde a_e),
\qquad
\widetilde a_e\not\simeq a^\star.
\]
The edit changes exactly one semantic claim while preserving the passage identifier, title, and unrelated content. Any coreferential repetitions of that claim are updated consistently. The replacement value must preserve the answer type and yield a uniquely warranted answer.

Let \(t_e\) denote the first trace position affected by the edit. The constructor preserves the maximal prefix, with \(\smash{\widetilde{s}_{j,e}=s^\star_j\ \text{for all }j<t_e}\), and rebuilds the trace from \(t_e\) using the edited context to establish the new answer. A completion indicator \(Q^e\) records whether construction succeeds.

\textbf{Evidence screen.}
To assess the intervention without relying on the generated traces, \textsc{screen-evidence} receives only \(\smash{(q,\mathcal{D},\widetilde{\mathcal{D}}_e,a^\star,\widetilde a_e)}\). It checks that the edit is localized and coherent and that all unedited passages remain unchanged. The edited context must uniquely warrant the new answer and no longer warrant the original answer; the original context must not warrant the new answer:
\[
U(q,\widetilde{\mathcal{D}}_e,\widetilde a_e)
\land
W(q,\widetilde{\mathcal{D}}_e,\widetilde a_e)
\land
\neg W(q,\widetilde{\mathcal{D}}_e,a^\star)
\land
\neg W(q,\mathcal{D},\widetilde a_e).
\]

\textbf{Evidence verification.}
In a single structured call, \textsc{verify-evidence} checks the edited response's local soundness, question--trace alignment, and trace--answer closure, together with step and closure switching. Let \(J_{\mathcal{D}'}(s)\) indicate whether step \(s\) is warranted under context \(\mathcal{D}'\). The shared prefix must remain valid under both contexts. At the first affected position \(t_e\), each step must be warranted under its own context but not the other:
\[
\forall j<t_e:\ J_{\mathcal{D}}(s^\star_j)=J_{\widetilde{\mathcal{D}}_e}(\widetilde{s}_{j,e})=1,\quad
J_{\mathcal{D}}(s^\star_{t_e})=J_{\widetilde{\mathcal{D}}_e}(\widetilde{s}_{t_e,e})=1,\quad
J_{\mathcal{D}}(\widetilde{s}_{t_e,e})=J_{\widetilde{\mathcal{D}}_e}(s^\star_{t_e})=0.
\]
The edited response must satisfy
\[
L(q,\widetilde{\mathcal{D}}_e,\widetilde r_e)
\land
A(q,\widetilde r_e)
\land
C(\widetilde r_e,\widetilde a_e),
\]
while closure must hold for each matched trace--answer pair and fail for both crossed pairs:
\[
C(r^\star,a^\star)
\land
C(\widetilde r_e,\widetilde a_e)
\land
\neg C(r^\star,\widetilde a_e)
\land
\neg C(\widetilde r_e,a^\star).
\]

Let \(V^e\) denote the conjunction of the screening and verification criteria. A deterministic gate \(H^e\) checks schema validity, single-passage replacement, preservation of unedited content, exact prefix identity, first divergence at \(t_e\), and normalized answer inequality. We retain only interventions satisfying \(\smash{V^e\land Q^e\land H^e}\). Matched evidence--response pairs serve as generation anchors; crossed trace--answer pairs are used only for margin scoring.

\textbf{Retention.}
Of the \(7{,}276\) accepted targets, \(5{,}169\) (\(71.0\%\)) yield a verified evidence intervention: \(1{,}622\) from 2WikiMultiHopQA, \(1{,}685\) from MuSiQue, and \(1{,}862\) from HotpotQA (Table~\ref{tab:training_construction_flow}, \emph{Evidence} column).

\subsection{Question Intervention}
\label{app:question_intervention}

A question intervention changes the requested target while keeping the evidence fixed. The edited question must induce a distinct reasoning path, with a locally sound trace that aligns with the new target and establishes the new answer.

\textbf{Question construction.}
For each source with an accepted evidence intervention, the constructor keeps \(\mathcal{D}\) fixed and changes one target-determining component of \(q\). It produces an edited question \(\widetilde q\) and response \(\widetilde y_q=(\widetilde r_q,\widetilde a_q)\), where \(\widetilde a_q\not\simeq a^\star\). The edited question must be natural and unambiguous, require multiple reasoning steps, and avoid revealing the answer while preserving unrelated constraints. No prefix constraint is imposed because the new target may change the reasoning path from the first step. A completion indicator \(Q^q\) records whether construction succeeds.

\textbf{Question screen.}
To assess the target change without relying on generated traces, \textsc{screen-question} receives only \(\smash{(q,\widetilde q,\mathcal{D},a^\star,\widetilde a_q)}\). It checks that the evidence remains unchanged, the requested target changes, and the edited question has a uniquely warranted answer. Neither answer may satisfy the other question:
\[
U(\widetilde q,\mathcal{D},\widetilde a_q)=1,\qquad
W(\widetilde q,\mathcal{D},\widetilde a_q)=1,\qquad
W(\widetilde q,\mathcal{D},a^\star)=0,\qquad
W(q,\mathcal{D},\widetilde a_q)=0.
\]
Degenerate, tautological, ambiguous, single-step, or answer-leaking questions are rejected.

\textbf{Question verification.}
In one structured call, \textsc{verify-question} separately assesses the edited response's local soundness, question--trace alignment, and trace--answer closure, together with four \(T\)-switching judgments. Each alignment judgment uses only its question--trace pair; merely mentioning a relevant entity or relation does not establish alignment. The edited response must satisfy
\[
L(\widetilde q,\mathcal{D},\widetilde r_q)=1,\qquad
A(\widetilde q,\widetilde r_q)=1,\qquad
C(\widetilde r_q,\widetilde a_q)=1.
\]
Target switching requires alignment for both matched pairs and misalignment for both crossed pairs:
\[
A(q,r^\star)=1,\qquad
A(\widetilde q,\widetilde r_q)=1,\qquad
A(q,\widetilde r_q)=0,\qquad
A(\widetilde q,r^\star)=0.
\]

Let \(V^q\) denote the conjunction of the screening and verification criteria. A deterministic gate \(H^q\) checks schema validity, identity of the shared evidence, normalized answer inequality, absence of literal answer leakage, support-identifier validity, and structural well-formedness. We retain only candidates satisfying \(\smash{V^q\land Q^q\land H^q}\). The matched question--response pair serves as a generation anchor, while crossed question--trace pairs are used only to score the two directional \(T\) margins. No closure-switching pair is constructed because the question and trace change together.

\textbf{Retention.}
Of the \(5{,}169\) sources with an accepted evidence intervention, \(3{,}757\) (\(72.7\%\)) also yield a verified question intervention: \(1{,}622\) from 2WikiMultiHopQA, \(946\) from MuSiQue, and \(1{,}189\) from HotpotQA (Table~\ref{tab:training_construction_flow}, \emph{Question} column).

\subsection{Packing, Splitting, and Audit}
\label{app:packing_and_audit}

\textbf{Acceptance and packing.}
Acceptance requires semantic approval \(V\), successful construction \(Q\), and deterministic checks \(H\). For an intervention candidate \(i\) of type \(x\in\{e,q\}\), we define \(V_i^x=\operatorname{Screen}_i^x\land\operatorname{Verify}_i^x\) and \(B_i^x=V_i^x\land Q_i^x\land H_i^x\). The target response follows the analogous rule \(B^\star=V^\star\land Q^\star\land H^\star\). In both cases, \(B=1\) denotes an accepted candidate.

We pack a source and its views into a family only if it passes the seed screen and has an accepted target response, at least one accepted evidence intervention, and exactly one accepted question intervention. Deterministic checks cover schema validity, identifier consistency, normalized answer relations, and structural well-formedness. Evidence interventions additionally require exactly one edited passage, byte-identical unedited passages, exact prefix preservation, and first divergence at \(t_e\). Question interventions require an unchanged context, a changed question, valid support identifiers, and no literal answer leakage. Semantic properties, including local soundness, alignment, closure, intervention isolation, and nondegeneracy, are assessed exclusively through \(V\).

Each family stores the complete original and edited contexts, the Gold, Evidence-CF, and Question-CF views, and their proof graphs, traces, answers, support identifiers, intervention metadata, semantic judgments, and construction provenance. Only matched views serve as generation targets. Crossed evidence--step, question--trace, and trace--answer pairs are formed within each family solely for the corresponding \(S\), \(T\), and \(C\) margins.

\textbf{Retention.}
Question verification yields \(3{,}757\) candidate families. Removing one MuSiQue family with a duplicate original question leaves \(3{,}756\). Reapplying the frozen construction and verification criteria retains \(2{,}852\) final families: \(1{,}276\) from 2WikiMultiHopQA, \(559\) from MuSiQue, and \(1{,}017\) from HotpotQA (Table~\ref{tab:training_construction_flow}, \emph{Packed} column).

\textbf{Splitting.}
We split at the family level so that each source and all its derived views remain in the same partition. Within each dataset, we sort families by the SHA-256 hash of a fixed salt, dataset name, and source identifier. The first \(10\%\), rounded to the nearest integer, form the development set; the remainder form the training set. This yields \(2{,}566\) training families and \(286\) development families, with no source or derived-view overlap.

\textbf{Human audit.}
After splitting, we draw a stratified random sample of \(300\) families using seed \(42\): \(100\) per dataset, comprising \(90\) training and \(10\) development families. Two annotators independently inspect each family's Gold response, one uniformly sampled Evidence-CF view, and its Question-CF view, without seeing automatic judgments, constructor rationales, or each other's decisions. The audit follows frozen guidelines shared by both annotators. They assess whether the Gold response is correct, supported, aligned, and closed; whether the evidence intervention is isolated and induces the required trace and answer changes; and whether the question intervention is valid and induces the required trace switch. A family passes only if all three inspected views pass.

Table~\ref{tab:human_audit} reports the adjudicated results. Dataset-level \(95\%\) confidence intervals use the Wilson method. Corpus-wide pass rates are weighted by each dataset's share of the final corpus, with \(95\%\) percentile intervals from \(2{,}000\) dataset-stratified bootstrap resamples using seed \(42\). We resample families within each dataset, keeping their three view labels together and corpus weights fixed. Because pass labels are near ceiling, we report raw agreement alongside Gwet's \(AC_1\). Before adjudication, annotators agree on \(292/300\) family decisions (\(97.3\%\), \(AC_1=0.972\)); a third annotator resolves the eight disagreements under the same frozen guidelines. After adjudication, \(290/300\) families pass (\(96.7\%\) unweighted; \(97.3\%\) corpus-weighted).

\begin{table}[t]
\centering
\small
\setlength{\tabcolsep}{4pt}
\renewcommand{\arraystretch}{1.12}
\caption{\textbf{Human audit results.}
Adjudicated pass rates (\%) with \(95\%\) confidence intervals.
A family passes only if all three audited views pass.}
\label{tab:human_audit}
\vspace{4pt}

\begin{tabular*}{\linewidth}{@{\extracolsep{\fill}}lccccc@{}}
\toprule
\textbf{Dataset}
& \textbf{Audited}
& \textbf{Gold}
& \textbf{Evidence-CF}
& \textbf{Question-CF}
& \textbf{Family} \\
\midrule

2WikiMultiHopQA
& 100
& 100.0\,[96.3,100.0]
& 99.0\,[94.6,99.8]
& 100.0\,[96.3,100.0]
& 99.0\,[94.6,99.8] \\

MuSiQue
& 100
& 98.0\,[93.0,99.4]
& 95.0\,[88.8,97.8]
& 99.0\,[94.6,99.8]
& 94.0\,[87.5,97.2] \\

HotpotQA
& 100
& 99.0\,[94.6,99.8]
& 97.0\,[91.5,99.0]
& 100.0\,[96.3,100.0]
& 97.0\,[91.5,99.0] \\

\midrule
Corpus-weighted
& 300
& 99.3\,[98.3,100.0]
& 97.5\,[95.6,99.0]
& 99.8\,[99.4,100.0]
& \textbf{97.3\,[95.3,98.9]} \\

\bottomrule
\end{tabular*}
\end{table}

\section{Edge Closure Learning: Additional Details}
\label{app:closure_objective_details}

This section explains how \textbf{\method{}} converts validated intervention data into three edge-specific training signals. An admissible response has a locally sound trace that aligns with the question and establishes the submitted answer. Evidence and question interventions are drawn from \(\mathcal I_e(z)\) and \(\mathcal I_q(z)\), respectively. All targets and interventions pass the semantic, completion, and deterministic checks described in Appendix~\ref{app:training_construction}, which also provides construction details and retention rates.

\subsection{Training Views and Validity}
\label{app:closure_views}

\textbf{Original view.}
A training source contains a question \(q\), evidence \(\mathcal D\), and an admissible target response \(y^\star=(r^\star,a^\star)\) whose submitted answer matches the reference. This view anchors generation under the original input.

\textbf{Evidence intervention.}
We replace a supporting passage \(\smash{e\in E_{\mathrm{dep}}(r^\star)}\), whose removal would leave a necessary premise of \(r^\star\) unsupported, to obtain the corresponding edited evidence context \(\smash{\widetilde{\mathcal D}_e}\) and an admissible response \(\smash{\widetilde y_e=(\widetilde r_e,\widetilde a_e)}\). The traces share a prefix and first diverge at \(t(e)\), the earliest step whose warrant changes under the edit. Validation requires each affected step to be warranted under its own evidence context but not the other, and each trace to establish its own answer but not the paired alternative. This intervention supplies the \(S\) edge from evidence to the first affected step and the \(C\) edge from trace to answer.

\textbf{Question intervention.}
With \(\mathcal D\) fixed, a localized target edit produces \(\widetilde q\) and an admissible response \(\widetilde y_q=(\widetilde r_q,\widetilde a_q)\). The edited trace must align with \(\widetilde q\) but not \(q\), while \(r^\star\) must align with \(q\) but not \(\widetilde q\). This intervention supplies the \(T\) edge, linking the requested target to the complete reasoning trace.

The resulting original and counterfactual pairs are
\[
S:\;(\mathcal D,s^\star_{t(e)})\leftrightarrow(\widetilde{\mathcal D}_e,\widetilde s_{t(e),e})
\qquad
T:\;(q,r^\star)\leftrightarrow(\widetilde q,\widetilde r_q)
\qquad
C:\;(r^\star,a^\star)\leftrightarrow(\widetilde r_e,\widetilde a_e).
\]
The notation omits inputs held fixed within each edge: \(q\) and the shared prefix for \(S\), \(\mathcal D\) for \(T\), and the original \((q,\mathcal D)\) for \(C\). The evidence intervention supplies both \(S\) and \(C\), while the question intervention supplies \(T\); no separate response is constructed for \(C\).

\subsection{Edge-Specific Scores}
\label{app:closure_scores}

Let \(\pi_\theta(z\mid u)\) denote the probability of output sequence \(z\) given input \(u\) under the model with parameters \(\theta\). We instantiate the generic score \(\ell_\theta(z\mid u)\) for three outputs: the first affected step, the complete trace, and the submitted answer:
\[
\begin{aligned}
F^S_{\theta,e}(s\mid\mathcal D')
&=
\frac{1}{|s|}
\log\pi_\theta
\bigl(s\mid q,\mathcal D',s^\star_{<t(e)}\bigr),\\
F^T_\theta(r\mid q')
&=
\frac{1}{|r|}
\log\pi_\theta
\bigl(r\mid q',\mathcal D\bigr),\\
F^C_\theta(a\mid r)
&=
\frac{1}{|a|}
\log\pi_\theta
\bigl(a\mid q,\mathcal D,r\bigr).
\end{aligned}
\]
Here, \(|\cdot|\) denotes token count. Each score averages the output's autoregressive token log probabilities, accounting for differences in candidate length. All scores are computed with teacher forcing, without sampling.

\textbf{Evidence to step.}
Both candidate steps are scored after the same validated prefix, so \(F^S\) isolates the first position where the evidence edit should change the trace. Restricting the score to this step separates the local dependency from later differences in the regenerated continuation.

\textbf{Question to trace.}
The score \(F^T\) covers the complete trace without the final answer, giving the reasoning its own preference signal. Under \(q\), the model should prefer \(r^\star\) to \(\smash{\widetilde r_q}\); under \(\widetilde q\), this preference should reverse.

\textbf{Trace to answer.}
In \(F^C\), \(q\) and \(\mathcal D\) remain fixed while the trace changes. The model should prefer \(a^\star\) after \(r^\star\) and \(\smash{\widetilde a_e}\) after \(\smash{\widetilde r_e}\). Holding the other inputs fixed ties the required preference reversal to the change in trace. The edited trace under the original context is used only for answer scoring; \(\smash{(q,\mathcal D,\widetilde r_e,\widetilde a_e)}\) is never a generation anchor.

For each edge \(x\), write \(F^x_\theta\) for its score and \(\smash{(u_x^{\mathrm{o}},z_x^{\mathrm{o}})\leftrightarrow(u_x^{\mathrm{c}},z_x^{\mathrm{c}})}\) for the original and counterfactual pair. The two directional margins are
\[
\Delta^{x,\mathrm{o}}_\theta
=
F^x_\theta(z_x^{\mathrm{o}}\mid u_x^{\mathrm{o}})
-
F^x_\theta(z_x^{\mathrm{c}}\mid u_x^{\mathrm{o}}),
\qquad
\Delta^{x,\mathrm{c}}_\theta
=
F^x_\theta(z_x^{\mathrm{c}}\mid u_x^{\mathrm{c}})
-
F^x_\theta(z_x^{\mathrm{o}}\mid u_x^{\mathrm{c}}).
\]
Both margins must meet the positive target \(\gamma_x\), so a fixed preference across the two inputs cannot satisfy both constraints. For each edge, \(\gamma_x>0\) sets the required separation, while \(\beta_x>0\) controls the smooth hinge transition.

\subsection{Complete Objective and Design Rationale}
\label{app:closure_design}

\textbf{Generation anchors.}
Each packed source \(z\) contains one accepted evidence intervention paired with the source's unique question intervention. The original response and the two counterfactual responses form three generation anchors, each paired with the input that warrants it.

Let \(S^{\mathrm{rsp}}_\theta(y\mid q',\mathcal D')\) denote the mean log likelihood over the trace and answer tokens of a complete response, excluding fixed formatting delimiters. The generation loss averages over the three anchors:
\[
\mathcal L_{\mathrm{gen}}
=
-\frac{1}{3}
\left[
S^{\mathrm{rsp}}_\theta(y^\star\mid q,\mathcal D)
+
S^{\mathrm{rsp}}_\theta(\widetilde y_e\mid q,\widetilde{\mathcal D}_e)
+
S^{\mathrm{rsp}}_\theta(\widetilde y_q\mid\widetilde q,\mathcal D)
\right].
\]

\textbf{Bidirectional supervision.}
The anchors directly encourage high likelihood for warranted responses. We complement them with bidirectional margins that supervise how output preferences should change under each intervention. Let \(h_\beta(t)=\beta\log(1+\exp(t/\beta))\) denote the smooth hinge, with \(\mathcal X=\{S,T,C\}\) and \(\mathcal V=\{\mathrm{o},\mathrm{c}\}\). The complete per-source objective is
\[
\mathcal L_{\method{}}
=
\mathcal L_{\mathrm{gen}}
+
\lambda\,
\mathbb E_{(x,v)\sim
\operatorname{Unif}(\mathcal X\times\mathcal V)}
\left[
h_{\beta_x}\!\left(
\gamma_x-\Delta^{x,v}_\theta
\right)
\right].
\]
The expectation averages uniformly over all six combinations of edge and direction, so \(\lambda\) controls the total margin weight relative to generation. The batch loss averages over sampled sources.

\textbf{Design rationale.}
Let \(\bar{\ell}_\theta\) denote mean token log-likelihood.
For a fixed tokenization, split \(y=(r,b)\), where \(b\) contains the answer delimiter, answer, and end-of-turn token.
Then
\[
S^{\mathrm{rsp}}_\theta(y\mid q,\mathcal D)
=
\frac{|r|}{|y|}\bar{\ell}_\theta(r\mid q,\mathcal D)
+
\frac{|b|}{|y|}\bar{\ell}_\theta(b\mid q,\mathcal D,r),
\qquad |y|=|r|+|b|.
\]
When \(|r|\gg|b|\), trace changes can dominate response preferences without changing answer preferences.
Separate \(T\) and \(C\) margins normalize their own scored spans, giving trace selection and answer closure dedicated supervision.
The \(S\) margin similarly isolates the first step affected by the evidence intervention.

\textbf{Training and inference.}
All terms are computed with teacher forcing and require no reference model. Additional views are used only during training. At inference, \method{} retains the standard mapping \((q,\mathcal D)\mapsto(r,a)\), with no auxiliary verifier, additional model calls, or decoding changes.

\textbf{Default hyperparameters.}
We use \(\lambda=1\), \(\gamma_x=0.5\), and \(\beta_x=0.1\) for all dependencies \(x\in\{S,T,C\}\), with the same defaults across datasets and backbones.
With token-normalized scores and an averaged margin loss, \(\lambda=1\) gives generation and margin supervision unit coefficients.
The target \(\gamma_x=0.5\) encourages a positive preference gap in both directions, while \(\beta_x=0.1\) smooths the hinge transition on a scale smaller than this gap.
For controlled comparisons, RM and all margin-based ablations share these values, with margin coefficients normalized to sum to \(\lambda\).
SFT and CF-SFT optimize generation alone, so the margin parameters do not apply.

\section{Experiments: Additional Details}
\label{app:experiments}

This appendix defines the evaluation protocol and validates the automatic judge used for behavioral outcomes. It then specifies the comparison objectives and training implementation before reporting full results, mechanism analyses, and robustness checks.

\subsection{Evaluator and Validation}
\label{app:evaluator}

\providecommand{\evalci}[3]{%
  \shortstack{#1\\[-1pt]{\scriptsize\textcolor{black!65}{[#2, #3]}}}%
}

\textbf{Evaluator.}
Human annotation across all methods, backbones, and training seeds is time-consuming. LLM-based judges offer a scalable alternative, but their reliability requires validation~\citep{gu2026survey,tan2025judgebench,chu-etal-2025-tracsum}. We use GPT-5.6 Sol~\citep{openai2026gpt56} (\texttt{gpt-5.6-sol}) with \texttt{high} reasoning effort for all experimental results and evaluate \texttt{low} and \texttt{medium} as sensitivity conditions. We compare all three configurations with the final human adjudications of the \(2{,}598\) diagnostic responses. Tables~\ref{tab:evaluator_validation} and~\ref{tab:evaluator_effort} report validation results for \texttt{high} and comparisons across reasoning efforts, respectively.

\textbf{Judgment definitions.}
All judgments follow the frozen diagnostic guidelines (Section~\ref{app:annotation}). Answer correctness compares the submitted answer with the reference, allowing semantic equivalence. Local soundness \(L\) requires every substantive step to be warranted; alignment \(A\) requires the trace to pursue the requested target; and closure \(C\) requires it to establish the submitted answer. Reference completeness \(K\) asks whether the trace's stated premises suffice to establish the reference answer, even if the final inference remains implicit. The guidelines specify the information that may justify each judgment:
\begingroup
\small
\[
\mathcal V_L=(q,\mathcal D,r),
\quad
\mathcal V_A=(q,r),
\quad
\mathcal V_C=(r,a),
\quad
\mathcal V_K=(q,r,a^\star),
\quad
\mathcal V_{\mathrm{Acc}}=(q,a,a^\star).
\]
\endgroup
For \(C\) and \(K\), annotators and the evaluator treat the trace's claims as premises without reassessing their evidence support. Facts appearing only in the evidence cannot fill gaps in the trace. Answer correctness is assessed separately and does not determine alignment or closure.

\textbf{Validation set and metrics.}
We evaluate all three configurations on the same \(2{,}598\) diagnostic responses to \(866\) questions across three datasets and three backbones. Records are matched by question identifier and backbone, with identical inputs and responses across configurations. We compare predictions against the final human adjudications. All configurations yield complete binary labels without exclusions or imputation, and stored \(L\) and \(G\) labels match their recomputed values. The \texttt{low}, \texttt{medium}, and \texttt{high} configurations label \(1\), \(2\), and \(1\) responses, respectively, as reliable but incorrect; these cases remain included.

For each criterion, we report raw agreement, failure F1 (\(F_{1,0}\)), balanced accuracy, Cohen's \(\kappa\), and the signed failure-rate difference:
\[
\Delta p_0(x)
=
\Pr_{\mathrm{LLM}}(\widehat x=0)
-
\Pr_{\mathrm{human}}(x=0),
\qquad
x\in\{\mathrm{Acc},L,A,C,K,G\}.
\]
Failure F1 treats label \(0\) as positive; balanced accuracy averages recall across passing and failing labels. Positive \(\Delta p_0\) indicates a higher automated failure rate. TRR and LGG differences use the same automated-minus-human convention, with all rate differences expressed in percentage points.

\textbf{Validation uncertainty.}
We estimate \(95\%\) percentile confidence intervals using \(2{,}000\) dataset-stratified question-level bootstrap resamples with seed \(42\). Each sampled question retains its three responses and their paired human and evaluator labels. Shared draws are used across criteria, configurations, and paired comparisons. Pooled validation estimates weight responses equally, while dataset-specific estimates use the corresponding stratum.

\textbf{Agreement with human adjudication.}
Table~\ref{tab:evaluator_validation} reports results for \texttt{high}, the configuration used throughout the experiments. Raw agreement ranges from \(97.31\%\) to \(99.73\%\) across the five directly evaluated criteria. Failure F1 reaches \(0.993\) for correctness, \(0.937\) for closure, and \(0.972\) for completeness. For local soundness and alignment, failure F1 is \(0.938\) and \(0.971\), respectively, with automated failure rates \(2.16\) and \(0.12\) percentage points above human estimates. Derived global sufficiency achieves \(99.08\%\) agreement and \(0.962\) failure F1. Overall, raw agreement is high, with the largest discrepancies concentrated in local soundness. All experimental methods are evaluated with the same evaluator, configuration, and frozen guidelines, providing a consistent basis for comparison. We further assess agreement with human judgments through a separate audit of the main experimental results (Appendix~\ref{app:experimental_human_audit}).

\begin{table}[h]
\centering
\footnotesize
\setlength{\tabcolsep}{2.5pt}
\renewcommand{\arraystretch}{1.15}
\caption{\textbf{Evaluator agreement with human adjudication.}
Results use \texttt{high} on all \(2{,}598\) diagnostic responses.
\(n_0\) counts human failure labels.
Agreement and balanced accuracy are percentages; \(F_{1,0}\) and \(\kappa\) are unitless.
\(\Delta p_0\) is the automated minus human failure rate in percentage points.
Brackets show \(95\%\) confidence intervals.}
\label{tab:evaluator_validation}
\vspace{4pt}

\begin{tabular*}{\linewidth}{@{\extracolsep{\fill}}lcccccc@{}}
\toprule
\textbf{Criterion}
& \(n_0\)
& \textbf{Agreement}
& \(F_{1,0}\)
& \textbf{BAcc.}
& \(\kappa\)
& \(\Delta p_0\) \\
\midrule

Correctness
& 484
& \evalci{99.73}{99.50}{99.92}
& \evalci{0.993}{0.987}{0.998}
& \evalci{99.83}{99.70}{99.95}
& \evalci{0.991}{0.984}{0.997}
& \evalci{+0.27}{+0.08}{+0.50} \\

Local soundness \(L\)
& 539
& \evalci{97.31}{96.57}{97.96}
& \evalci{0.938}{0.922}{0.953}
& \evalci{97.82}{97.18}{98.41}
& \evalci{0.921}{0.900}{0.940}
& \evalci{+2.16}{+1.46}{+2.89} \\

Alignment \(A\)
& 190
& \evalci{99.58}{99.31}{99.81}
& \evalci{0.971}{0.953}{0.986}
& \evalci{98.80}{97.72}{99.65}
& \evalci{0.969}{0.949}{0.985}
& \evalci{+0.12}{-0.12}{+0.38} \\

Closure \(C\)
& 134
& \evalci{99.35}{99.00}{99.65}
& \evalci{0.937}{0.904}{0.965}
& \evalci{97.19}{95.27}{98.85}
& \evalci{0.934}{0.899}{0.963}
& \evalci{+0.12}{-0.23}{+0.46} \\

Completeness \(K\)
& 285
& \evalci{99.38}{99.00}{99.69}
& \evalci{0.972}{0.955}{0.986}
& \evalci{98.42}{97.21}{99.36}
& \evalci{0.968}{0.949}{0.984}
& \evalci{0.00}{-0.35}{+0.35} \\

\midrule
Global sufficiency \(G\)
& 313
& \evalci{99.08}{98.65}{99.42}
& \evalci{0.962}{0.944}{0.977}
& \evalci{98.37}{97.31}{99.24}
& \evalci{0.957}{0.937}{0.974}
& \evalci{+0.31}{-0.08}{+0.69} \\

\bottomrule
\end{tabular*}
\end{table}

\textbf{Sensitivity to reasoning effort.}
Table~\ref{tab:evaluator_effort} compares \texttt{low}, \texttt{medium}, and \texttt{high} on the same responses. The \texttt{high} configuration achieves the highest mean agreement (\(99.07\%\)) and mean failure F1 (\(0.962\)) across \(\mathrm{Acc},L,A,C,K\), excluding the derived label \(G\). Relative to \texttt{low}, mean agreement increases by \(1.19\) points (\(95\%\) CI: \([0.90,\,1.50]\)), and mean failure F1 increases by \(0.066\) (\([0.050,\,0.082]\)). Across configurations, TRR is \(1.54\)--\(2.23\) points below the human estimate, while LGG rate differences range from \(-0.31\) to \(+0.12\) points.

\begin{table}[h]
\centering
\small
\setlength{\tabcolsep}{3pt}
\renewcommand{\arraystretch}{1.15}
\caption{\textbf{Sensitivity to evaluator reasoning effort.}
Mean agreement (\%), mean failure F1, and maximum absolute failure-rate difference summarize \(\mathrm{Acc},L,A,C,K\).
Rate differences are automated minus human in percentage points.
Brackets show \(95\%\) paired bootstrap confidence intervals.}
\label{tab:evaluator_effort}
\vspace{4pt}

\begin{tabular*}{0.75\linewidth}{@{\extracolsep{\fill}}lccccc@{}}
\toprule
\textbf{Effort}
& \textbf{Mean agr.}
& \textbf{Mean \(F_{1,0}\)}
& \(\max|\Delta p_0|\)
& \(\Delta\mathrm{TRR}\)
& \(\Delta\mathrm{LGG}\) \\
\midrule

\texttt{low}
& 97.88
& 0.897
& 1.54
& \evalci{-1.54}{-2.35}{-0.81}
& \evalci{0.00}{-0.58}{+0.58} \\

\texttt{medium}
& 97.96
& 0.902
& 2.12
& \evalci{-2.23}{-3.08}{-1.46}
& \evalci{+0.12}{-0.46}{+0.69} \\

\texttt{high}
& 99.07
& 0.962
& 2.16
& \evalci{-1.85}{-2.54}{-1.23}
& \evalci{-0.31}{-0.73}{+0.15} \\

\bottomrule
\end{tabular*}
\end{table}

\textbf{Agreement on derived metrics.}
Under \texttt{high}, TRR is \(71.82\%\), compared with \(73.67\%\) under human adjudication, a difference of \(-1.85\) points (\(95\%\) CI: \([-2.54,\,-1.23]\)). LGG rates are closer: \(5.27\%\) versus \(5.58\%\), a difference of \(-0.31\) points (\([-0.73,\,0.15]\)), reflecting similar downward shifts in LSR and TRR. At the response level, \texttt{high} identifies \(126\) of the \(145\) human-labeled LGG cases and assigns \(11\) additional cases, yielding precision of \(91.97\%\), recall of \(86.90\%\), and F1 of \(0.894\). Corresponding LGG F1 values are \(0.848\) for \texttt{low} and \(0.853\) for \texttt{medium}.

\textbf{Dataset-specific differences.}
Table~\ref{tab:evaluator_dataset_bias} breaks down the metric differences by dataset. TRR differences are \(-0.12\) points on 2WikiMultiHopQA, \(-1.70\) on HotpotQA, and \(-3.92\) on MuSiQue. LGG differences are \(+0.12\), \(+0.20\), and \(-1.44\) points, respectively. These results characterize agreement on the diagnostic responses; the separate human audit of the main experiment assesses annotations of responses generated by the evaluated methods.

\begin{table}[h]
\centering
\small
\setlength{\tabcolsep}{5pt}
\renewcommand{\arraystretch}{1.15}
\caption{\textbf{Dataset-specific metric differences for \texttt{high}.}
\(n\) counts responses.
Differences are automated minus human in percentage points, with \(95\%\) paired bootstrap confidence intervals.
The pooled row weights responses equally.}
\label{tab:evaluator_dataset_bias}
\vspace{4pt}

\begin{tabular*}{0.75\linewidth}{@{\extracolsep{\fill}}lccc@{}}
\toprule
\textbf{Dataset}
& \(n\)
& \(\Delta\mathrm{TRR}\)
& \(\Delta\mathrm{LGG}\) \\
\midrule

2WikiMultiHopQA
& 834
& \evalci{-0.12}{-0.48}{0.00}
& \evalci{+0.12}{0.00}{+0.36} \\

HotpotQA
& 999
& \evalci{-1.70}{-2.80}{-0.70}
& \evalci{+0.20}{-0.40}{+0.90} \\

MuSiQue
& 765
& \evalci{-3.92}{-5.75}{-2.35}
& \evalci{-1.44}{-2.75}{-0.26} \\

\midrule
All (pooled)
& 2,598
& \evalci{-1.85}{-2.54}{-1.23}
& \evalci{-0.31}{-0.73}{+0.15} \\

\bottomrule
\end{tabular*}
\end{table}

\subsection{Experimental Evaluation Protocol}
\label{app:experimental_protocol}

\textbf{Evaluation set and generation.}
All methods are evaluated on the same \(866\) retained diagnostic questions: \(278\) from 2WikiMultiHopQA, \(255\) from MuSiQue, and \(333\) from HotpotQA. Each adapted checkpoint generates one response per question using the diagnostic study's structured CoT prompt. Backbone results use the fixed diagnostic generations. Generation uses greedy decoding with at most \(1{,}024\) new tokens and an \(8{,}192\)-token sequence limit; explicit thinking mode is disabled where supported. All responses are assessed using the evaluator and \texttt{high} configuration described in Appendix~\ref{app:evaluator}.

\textbf{Metrics.}
We recompute \(\widehat L\) from the substantive-step labels and derive \(\widehat G=\widehat A\land\widehat C\). Across \(n\) responses, trace reliability and LGG rates are
\[
\widehat{\mathrm{TRR}}
=
\frac{1}{n}\sum_{i=1}^{n}\widehat L_i\widehat G_i,
\qquad
\widehat{\mathrm{LGG}}
=
\frac{1}{n}\sum_{i=1}^{n}\widehat L_i(1-\widehat G_i).
\]
Answer accuracy is the mean correctness label; LSR and GSR are the mean \(\widehat L\) and \(\widehat G\) labels, respectively. Alignment and closure rates are computed analogously. We separately report a supplementary deterministic score based on normalized, bidirectional whole-token-span matching, which differs from both semantic evaluation and strict exact match (Appendix~\ref{app:full_results}). All rates are reported as percentages.

\textbf{Aggregation.}
For adapted methods, we first compute each metric within each dataset--backbone--seed combination, then average over the three training seeds, \(\{42,43,44\}\), within each dataset--backbone setting. Aggregate main results assign equal weight to the nine settings. Backbone results are aggregated equally across the same nine settings using their fixed generations. Thus, each dataset and backbone contributes equally to the aggregate main results, whereas the pooled evaluator-validation estimates in Appendix~\ref{app:evaluator} weight individual responses equally.

\textbf{Uncertainty estimation for aggregate main results.}
We compute \(95\%\) paired hierarchical percentile confidence intervals from \(5{,}000\) bootstrap resamples with seed \(2027\). We resample backbones and, for each sampled backbone, resample questions within each fixed dataset and training seeds. Draws are shared across methods, and each training-seed draw is shared across datasets. Backbone results retain their fixed generations without training-seed resampling. Each replicate applies the same equal-weight aggregation as the point estimates, and the \(2.5\)th and \(97.5\)th percentiles form the confidence interval. The separate human audit, objective ablations, and dependency-intervention comparisons use the procedures specified in their respective appendix sections.

\subsection{Comparison Objectives}
\label{app:objectives}

\textbf{Common setup.}
All adapted methods, including SFT, Counterfactual SFT (CF-SFT), Full-Response Margin (RM), \method{}, and its ablations, start independently from the same pretrained backbone. They share the packed source partitions, source sampler, and context serialization. Methods using counterfactual supervision also share the accepted interventions and generation anchors. For each sampled source, one evidence intervention is selected uniformly and paired with the source's unique question intervention. All likelihoods are computed with teacher forcing, and minibatch averaging is implicit. For a complete response \(y=(r,a)\) conditioned on input \(u\), we define the token-normalized score
\[
S_\theta^{\mathrm{rsp}}(y\mid u)
=
\frac{1}{|y|}\log\pi_\theta(y\mid u).
\]
We write \(u:\ z^+\succ z^-\) to indicate a preference for \(z^+\) over \(z^-\) under input \(u\). Superscripts \(\mathrm{o}\) and \(\mathrm{c}\) denote original and counterfactual conditioning views, respectively. Both margin-based objectives use the smooth hinge
\[
h_\beta(z)
=
\beta\log\!\left(1+\exp(z/\beta)\right),
\qquad \beta>0.
\]

\textbf{SFT: original-response supervision.}
Given the original question \(q\) and evidence \(\mathcal D\), standard SFT trains on the validated response \(y^\star\) as its sole generation anchor:
\[
\mathcal L_{\mathrm{SFT}}
=
-S_\theta^{\mathrm{rsp}}(y^\star\mid q,\mathcal D).
\]
CF-SFT, RM, \method{}, and all controlled ablations retain this anchor. SFT thus isolates original-response supervision before adding counterfactual anchors or preference margins.

\textbf{Counterfactual SFT: matched generation anchors.}
CF-SFT trains on the original response under \((q,\mathcal D)\), the evidence-intervention response under \((q,\widetilde{\mathcal D}_e)\), and the question-intervention response under \((\widetilde q,\mathcal D)\). The objective averages their generation losses:
\[
\mathcal L_{\mathrm{CF\text{-}SFT}} \equiv \mathcal L_{\mathrm{gen}} = -\frac{1}{3}\Bigl[S_\theta^{\mathrm{rsp}}(y^\star\mid q,\mathcal D) + S_\theta^{\mathrm{rsp}}(\widetilde y_e\mid q,\widetilde{\mathcal D}_e) + S_\theta^{\mathrm{rsp}}(\widetilde y_q\mid\widetilde q,\mathcal D)\Bigr].
\]
Each response is trained only under an input that warrants it. RM, \method{}, and all controlled ablations retain these three generation anchors. CF-SFT thus isolates counterfactual generation supervision before adding preference margins.

\textbf{Full-Response Margin: preferences over complete responses.}
RM retains the three generation anchors in \(\mathcal L_{\mathrm{gen}}\) and adds bidirectional margins over complete responses \(y=(r,a)\). Each input should favor its matched response over the paired alternative, yielding four comparisons:

\begin{center}
\renewcommand{\arraystretch}{1.12}

\begin{tabular*}{0.75\linewidth}{@{\extracolsep{\fill}}cll@{}}
\toprule
View & Conditioning & Preference \\
\midrule

\(\mathsf R_e^{\mathrm{o}}\)
& \((q,\mathcal D)\)
& \(y^\star\succ\widetilde y_e\) \\

\(\mathsf R_e^{\mathrm{c}}\)
& \((q,\widetilde{\mathcal D}_e)\)
& \(\widetilde y_e\succ y^\star\) \\[3pt]

\(\mathsf R_q^{\mathrm{o}}\)
& \((q,\mathcal D)\)
& \(y^\star\succ\widetilde y_q\) \\

\(\mathsf R_q^{\mathrm{c}}\)
& \((\widetilde q,\mathcal D)\)
& \(\widetilde y_q\succ y^\star\) \\

\bottomrule
\end{tabular*}
\end{center}

For each row, let \(u_{i,v}\) denote the conditioning input and \((y^+_{i,v},y^-_{i,v})\) the preferred and dispreferred responses, with \(i\in\{e,q\}\) and \(v\in\mathcal V=\{\mathrm{o},\mathrm{c}\}\). The preference margin and its penalty are
\[
\Delta_\theta^{\mathrm{rsp},i,v}
=
S_\theta^{\mathrm{rsp}}(y^+_{i,v}\mid u_{i,v})
-
S_\theta^{\mathrm{rsp}}(y^-_{i,v}\mid u_{i,v}),
\qquad
\Psi_{i,v}
=
h_{\beta_{\mathrm{rsp}}}\!\left(
\gamma_{\mathrm{rsp}}-\Delta_\theta^{\mathrm{rsp},i,v}
\right).
\]
The full objective is
\[
\mathcal L_{\mathrm{RM}}
=
\mathcal L_{\mathrm{gen}}
+
\frac{\lambda}{6}
\sum_{v\in\mathcal V}
\left(2\Psi_{e,v}+\Psi_{q,v}\right).
\]
The four terms receive weights in the ratio \(2:2:1:1\), allocating two thirds of the total margin weight to evidence interventions and one third to question interventions. This matches the allocation to \(S,C\) and \(T\), respectively, in \method{}, while scoring complete responses. RM uses the same target margin and hinge smoothing as \method{}, with \(\gamma_{\mathrm{rsp}}=0.5\) and \(\beta_{\mathrm{rsp}}=0.1\), and requires no reference model. With generation anchors and margin allocation held fixed, this baseline tests the added value of dependency-specific supervision beyond preferences over complete responses.

\textbf{\method{}: preferences on individual dependencies.}
\method{} retains the same generation anchors and adds bidirectional margins on three dependent outputs: the first affected step for \(S\), the complete reasoning trace for \(T\), and the answer for \(C\). Let
\[
p_e=s^\star_{<t(e)}=\widetilde s_{<t(e),e}
\]
denote the prefix shared by the original and evidence-edited traces. The six comparisons are:

\begin{center}
\renewcommand{\arraystretch}{1.12}

\begin{tabular*}{0.75\linewidth}{@{\extracolsep{\fill}}cll@{}}
\toprule
View & Conditioning & Preference \\
\midrule

\(\mathsf S^{\mathrm{o}}\)
& \((q,\mathcal D,p_e)\)
& \(s^\star_{t(e)}\succ\widetilde s_{t(e),e}\) \\

\(\mathsf S^{\mathrm{c}}\)
& \((q,\widetilde{\mathcal D}_e,p_e)\)
& \(\widetilde s_{t(e),e}\succ s^\star_{t(e)}\) \\[3pt]

\(\mathsf T^{\mathrm{o}}\)
& \((q,\mathcal D)\)
& \(r^\star\succ\widetilde r_q\) \\

\(\mathsf T^{\mathrm{c}}\)
& \((\widetilde q,\mathcal D)\)
& \(\widetilde r_q\succ r^\star\) \\[3pt]

\(\mathsf C^{\mathrm{o}}\)
& \((q,\mathcal D,r^\star)\)
& \(a^\star\succ\widetilde a_e\) \\

\(\mathsf C^{\mathrm{c}}\)
& \((q,\mathcal D,\widetilde r_e)\)
& \(\widetilde a_e\succ a^\star\) \\

\bottomrule
\end{tabular*}
\end{center}

The \(S\) views change the evidence while keeping the question and shared prefix fixed. The \(T\) views change the question while keeping the evidence fixed. The \(C\) views change the trace while keeping both the question and evidence fixed. In \(\mathsf C^{\mathrm{c}}\), the edited trace \(\widetilde r_e\) serves only as conditioning for answer scoring; it is never a generation target under the original evidence.

Let \(\mathcal X=\{S,T,C\}\). For each row, \(\Delta_\theta^{x,v}\) is the difference between the preferred and dispreferred edge scores, with penalty
\[
\Phi_{x,v}
=
h_{\beta_x}\!\left(\gamma_x-\Delta_\theta^{x,v}\right).
\]
The complete objective is
\[
\mathcal L_{\method{}}
=
\mathcal L_{\mathrm{gen}}
+
\frac{\lambda}{6}
\sum_{x\in\mathcal X}
\sum_{v\in\mathcal V}
\Phi_{x,v}.
\]
Each dependency receives one third of the total margin weight, split equally between its two directions. Appendix~\ref{app:closure_scores} provides the full score definitions and conditioning views.

\textbf{Controlled ablations.}
All ablations retain the same three generation anchors, including both counterfactual responses, and the same training protocol as \method{}. They vary which edges and directions receive margin supervision. Each retained edge contributes one margin per retained direction, as summarized below.

\begin{center}
\small
\setlength{\tabcolsep}{5pt}
\renewcommand{\arraystretch}{1.12}

\begin{tabular*}{0.75\linewidth}{@{\extracolsep{\fill}}lccc@{}}
\toprule
Variant & Edges & Directions & Weight per term \\
\midrule

Full
& \(\{S,T,C\}\)
& \(\{\mathrm{o},\mathrm{c}\}\)
& \(\lambda/6\) \\

\(-S\)
& \(\{T,C\}\)
& \(\{\mathrm{o},\mathrm{c}\}\)
& \(\lambda/4\) \\

\(-T\)
& \(\{S,C\}\)
& \(\{\mathrm{o},\mathrm{c}\}\)
& \(\lambda/4\) \\

\(-C\)
& \(\{S,T\}\)
& \(\{\mathrm{o},\mathrm{c}\}\)
& \(\lambda/4\) \\

One-sided
& \(\{S,T,C\}\)
& \(\{\mathrm{o}\}\)
& \(\lambda/3\) \\

\bottomrule
\end{tabular*}
\end{center}

\emph{Edge removal (\(-S\), \(-T\), \(-C\)).}
Removing edge \(x_0\) drops both directional margins, \(\Phi_{x_0,\mathrm{o}}\) and \(\Phi_{x_0,\mathrm{c}}\), leaving four comparisons:
\[
\mathcal L_{-x_0}
=
\mathcal L_{\mathrm{gen}}
+
\frac{\lambda}{4}
\sum_{x\in\mathcal X\setminus\{x_0\}}
\sum_{v\in\mathcal V}\Phi_{x,v}.
\]
The \(-S\), \(-T\), and \(-C\) variants assess the contributions of evidence-to-step, question-to-trace, and trace-to-answer margins, respectively. The intervention data and generation anchors associated with the removed edge remain in training.

\emph{Direction removal (one-sided).}
This variant retains all three edges but only the original-condition margins, \(\mathsf S^{\mathrm{o}}\), \(\mathsf T^{\mathrm{o}}\), and \(\mathsf C^{\mathrm{o}}\):
\[
\mathcal L_{\mathrm{orig}}
=
\mathcal L_{\mathrm{gen}}
+
\frac{\lambda}{3}
\sum_{x\in\mathcal X}\Phi_{x,\mathrm{o}}.
\]
It assesses the contribution of enforcing preference reversal under counterfactual inputs, beyond original-condition margins and counterfactual generation supervision.

In every variant, the margin coefficients sum to \(\lambda\). Relative to the full objective, each retained term is scaled by \(3/2\) after edge removal and by \(2\) after direction removal. This preserves the total coefficient weight across different sets of active comparisons.

\subsection{Training Implementation and Compute}
\label{app:training_implementation}

\textbf{Training protocol.}
We pool sources across the three datasets and use a fixed split of \(2{,}566\) training and \(286\) development sources. For each backbone and seed, all adapted methods start independently from the same pretrained checkpoint and share the source sampler, context serialization, and optimization settings. CF-SFT, RM, \method{}, and its ablations also use the same three generation anchors defined in Appendix~\ref{app:objectives}.

\textbf{Number of trained models.}
We train \(48\) independent LoRA adapters, one per method--backbone--seed combination. The main comparison includes \(36\) adapters covering four objectives (SFT, CF-SFT, RM, and \method{}), three backbones, and three seeds, \(\{42,43,44\}\). Four controlled ablations on Qwen3-8B add \(12\) adapters using the same seeds. This yields \(24\) adapters for Qwen3-8B and \(12\) each for Qwen3-32B and Ministral-3-8B. Each adapter is trained on the pooled datasets, with epoch checkpoints serving as selection candidates from the same run.

\textbf{Optimization.}
We train with data parallelism across four GPUs using the settings in Table~\ref{tab:training_hyperparameters}. Batch sizes count source packs, with the number of likelihood evaluations per pack determined by the objective. A full training run comprises \(3{,}210\) microsteps per worker, yielding \(1{,}605\) optimizer updates with gradient accumulation.

\begin{table}[t]
\centering
\small
\setlength{\tabcolsep}{5pt}
\renewcommand{\arraystretch}{1.12}
\caption{\textbf{Training hyperparameters.} Settings are shared across
objectives within each backbone. Margin parameters apply to RM,
\method{}, and its ablations.}
\label{tab:training_hyperparameters}
\begin{tabularx}{\linewidth}{@{}
>{\raggedright\arraybackslash}p{0.32\linewidth}
>{\raggedright\arraybackslash}X@{}}
\toprule
\textbf{Setting} & \textbf{Value} \\
\midrule
LoRA targets & All linear layers in the language model \\
LoRA rank / scale / dropout & \(16\) / \(32\) / \(0\) \\
Precision / gradient checkpointing & BF16 / enabled \\
\midrule
Batch size per GPU & \(1\) source pack per microstep \\
Gradient accumulation & \(2\) microsteps per optimizer update \\
Effective global batch size & \(8\) source packs (\(4\) GPUs \(\times 1\times 2\)) \\
Scheduled training budget & \(5\) epochs; \(1{,}605\) optimizer updates \\
Training seeds & \(\{42,43,44\}\) \\
\midrule
Optimizer & AdamW \\
Peak learning rate
& \(5\times10^{-5}\) for Qwen3-8B and Qwen3-32B;
  \(10^{-4}\) for Ministral-3-8B \\
Learning-rate schedule & Linear decay with \(50\) warmup optimizer updates \\
Weight decay / gradient clipping & \(0\) / maximum gradient norm \(1.0\) \\
Maximum sequence length & \(8{,}192\) tokens, including input and scored output \\
Development evaluation & Every epoch (\(321\) optimizer updates) \\
Margin weight & \(\lambda=1\) \\
Margin target / temperature & \(\gamma=0.5\) / \(\beta=0.1\), shared across all margins \\
\bottomrule
\end{tabularx}
\end{table}

\textbf{Checkpoint selection.}
At the end of each epoch, we evaluate the model using greedy decoding with a maximum of \(512\) new tokens. For each run, we select the checkpoint with the highest weighted development answer accuracy:
\[
A_{\mathrm{sel}}
=
\tfrac{1}{2}A_{\mathrm{o}}
+\tfrac{1}{4}A_{\mathrm{e}}
+\tfrac{1}{4}A_{\mathrm{q}},
\]
where \(A_{\mathrm{o}}\), \(A_{\mathrm{e}}\), and \(A_{\mathrm{q}}\) denote answer accuracy under the original, evidence-edited, and question-edited inputs, respectively. Answers are scored using deterministic normalization and matching against the target answer and available aliases. Ties favor the earlier checkpoint. We apply the same selection rule across all objectives, backbones, and seeds.

\textbf{Hardware and compute.}
Main experiments with Qwen3-8B and Ministral-3-8B use four NVIDIA RTX A6000 GPUs with \(256\)\,GB of host memory; Qwen3-32B uses four NVIDIA H100 GPUs. For the Qwen3-8B ablations, seed \(42\) uses four RTX A6000 GPUs, while seeds \(43\) and \(44\) use four H100 GPUs because these GPUs were available.

Table~\ref{tab:training_cost} reports wall-clock time per complete training run, averaged over completed seeds. Timing starts at W\&B initialization after model loading and ends at the last logged event. It includes training, development evaluation, and checkpoint I/O within this interval, but excludes data construction, separate test evaluation, and unlogged startup or shutdown overhead. GPU hours are computed as recorded wall-clock hours multiplied by four. Across all \(48\) recorded runs, logged usage totals \(3{,}544.0\) GPU hours across both GPU types. All methods share the scheduled source exposure and optimizer-update budget, while the number of likelihood evaluations depends on the objective. All adapted methods use the same single-pass generation procedure at inference.

\begin{table}[t]
\centering
\small
\setlength{\tabcolsep}{5pt}
\renewcommand{\arraystretch}{1.12}
\caption{\textbf{Recorded time per training run (hours).}
Values are means \(\pm\) sample standard deviations across completed seeds.
All runs use four GPUs.
Main experiments use RTX A6000 GPUs for the 8B backbones and H100 GPUs for Qwen3-32B.
Ablation statistics pool RTX A6000 timings for seed \(42\) and H100 timings for seeds \(43\) and \(44\).}
\label{tab:training_cost}
\vspace{4pt}

\begin{tabular*}{\linewidth}{@{\extracolsep{\fill}}lccc@{}}
\toprule
\textbf{Method}
& \textbf{Qwen3-8B}
& \textbf{Ministral-3-8B}
& \textbf{Qwen3-32B} \\
\midrule

SFT
& \(5.20\pm0.13\)
& \(4.89\pm0.13\)
& \(4.54\pm0.02\) \\

CF-SFT
& \(10.64\pm0.18\)
& \(10.09\pm0.13\)
& \(8.59\pm0.04\) \\

RM
& \(22.01\pm0.32\)
& \(20.93\pm0.14\)
& \(17.12\pm0.06\) \\

\method{}
& \(43.73\pm0.05\)
& \(42.05\pm0.26\)
& \(33.57\pm0.04\) \\

\midrule
\method{} w/o \(S\)
& \(19.75\pm11.76\)
& ---
& --- \\

\method{} w/o \(T\)
& \(22.09\pm9.17\)
& ---
& --- \\

\method{} w/o \(C\)
& \(19.63\pm11.81\)
& ---
& --- \\

\method{} one-sided
& \(14.00\pm11.60\)
& ---
& --- \\

\bottomrule
\end{tabular*}
\end{table}

\subsection{Extended Main Results}
\label{app:full_results}

This section provides evaluation details and analyses for the main results in Section~\ref{sec:main_results}.

\subsubsection{Evaluation Setup}

All methods are evaluated on the same \(866\) diagnostic questions: \(278\) from 2WikiMultiHopQA, \(255\) from MuSiQue, and \(333\) from HotpotQA. The main comparison includes \(31{,}176\) responses from \(36\) adapted checkpoints, plus \(2{,}598\) fixed diagnostic backbone responses. Adapted methods use three training seeds, \(\{42,43,44\}\). For each metric, we compute rates within each dataset--backbone--seed setting, average across seeds, and then equally average the nine dataset--backbone settings. The backbone uses one fixed response per question and model, with the same averaging across settings. This gives each dataset equal weight despite differing question counts. Evaluator validation is reported separately in Appendix~\ref{app:evaluator}.

\subsubsection{Human Audit of Experimental Results}
\label{app:experimental_human_audit}

\begin{wraptable}{r}{0.5\textwidth}
\centering
\small
\setlength{\tabcolsep}{2pt}
\renewcommand{\arraystretch}{1.12}
\caption{\textbf{Evaluator agreement on the human-audit sample.}
Acc.\ denotes answer correctness.
\(n_0\) counts human-labeled failures among \(360\) responses.
Failure F1 treats label \(0\) as positive; \(G=A\land C\) is derived.}
\label{tab:appendix_human_audit_agreement}
\vspace{4pt}

\begin{tabular*}{\linewidth}{@{\extracolsep{\fill}}lcccc@{}}
\toprule
\textbf{Label}
& \(\boldsymbol{n_0}\)
& \textbf{Agreement (\%)}
& \(\boldsymbol{F_{1,0}}\)
& \(\boldsymbol{\kappa}\) \\
\midrule
\(\mathrm{Acc.}\) & 31 & 99.72 & 0.984 & 0.983 \\
\(L\)            & 35 & 95.83 & 0.737 & 0.715 \\
\(A\)            & 14 & 99.17 & 0.903 & 0.899 \\
\(C\)            & 32 & 99.44 & 0.969 & 0.966 \\
\(K\)            & 63 & 98.33 & 0.951 & 0.941 \\
\addlinespace[2pt]
\cdashline{1-5}[3pt/2pt]
\addlinespace[2pt]
\(G\)            & 44 & 98.61 & 0.945 & 0.937 \\
\bottomrule
\end{tabular*}
\end{wraptable}

\textbf{Audit design.}
We audit SFT, CF-SFT, RM, and \method{} on the same \(90\) questions, with \(10\) questions per dataset--backbone setting and \(360\) responses in total. Each question is assigned to one backbone, and all audited checkpoints use training seed \(42\). Two reviewers assess every response: they agree on \(335\) responses, and a third reviewer adjudicates the remaining \(25\). We use the final human labels throughout. Automated and human judgments evaluate identical questions, evidence, traces, and submitted answers. Equal sample sizes across settings make pooled rates equivalent to equal-setting means.

\textbf{Agreement with human adjudication.}
Table~\ref{tab:appendix_human_audit_agreement} reports agreement between automated and human labels. Across the five primary judgments, raw agreement ranges from \(95.83\%\) to \(99.72\%\), with answer correctness agreeing on \(359/360\) responses. Agreement for derived global sufficiency is \(98.61\%\), with failure F1 of \(0.945\). Disagreement patterns differ by criterion: the evaluator identifies \(31\) of \(32\) human-labeled closure failures, misses one, and flags one additional response; for local soundness, it misses \(14\) of \(35\) human failures and flags one additional response.

\textbf{Metric differences and uncertainty.}
Table~\ref{tab:appendix_human_audit_results} compares automated and human scores on the same responses. We define their difference as \(\Delta M=M_{\mathrm{auto}}-M_{\mathrm{human}}\) and report paired differences in Table~\ref{tab:appendix_human_audit_differences}. All confidence intervals in this audit are \(95\%\) percentile intervals from \(10{,}000\) paired bootstrap resamples using seed \(42\). Each resample draws \(10\) questions with replacement within each fixed dataset--backbone setting, preserving all four methods and both label sources. This procedure captures question-sampling uncertainty within the audited seed-\(42\) runs.

\begin{table}[htbp]
\centering
\small
\setlength{\tabcolsep}{4pt}
\renewcommand{\arraystretch}{1.12}
\caption{\textbf{Automated and human scores on the audit sample.}
Each method contributes \(90\) responses.
All scores are percentages: Acc.\(\uparrow\), LSR\(\uparrow\), GSR\(\uparrow\), TRR\(\uparrow\), and LGG\(\downarrow\).
LGG additionally reports the number of responses with \(L=1\) and \(G=0\) in parentheses.
Bold marks the best human score in each column.}
\label{tab:appendix_human_audit_results}
\vspace{4pt}

\begin{tabular*}{\linewidth}{@{\extracolsep{\fill}}llrrrrr@{}}
\toprule
\textbf{Method}
& \textbf{Labels}
& \textbf{Acc.}
& \textbf{LSR}
& \textbf{GSR}
& \textbf{TRR}
& \textbf{LGG} \\
\midrule
\multirow{2}{*}{SFT}
& Automated
& 91.11 & 95.56 & 85.56 & 83.33 & 12.22 (11) \\
& Human
& 91.11 & \textbf{94.44} & 85.56 & 82.22 & 12.22 (11) \\

\addlinespace[2pt]
\cdashline{1-7}[3pt/2pt]
\addlinespace[2pt]
\multirow{2}{*}{CF-SFT}
& Automated
& 88.89 & 92.22 & 86.67 & 82.22 & 10.00 (9) \\
& Human
& 90.00 & 85.56 & 87.78 & 77.78 & 7.78 (7) \\

\addlinespace[2pt]
\cdashline{1-7}[3pt/2pt]
\addlinespace[2pt]
\multirow{2}{*}{RM}
& Automated
& 92.22 & 93.33 & 85.56 & 82.22 & 11.11 (10) \\
& Human
& \textbf{92.22} & 91.11 & 86.67 & 81.11 & 10.00 (9) \\

\addlinespace[2pt]
\cdashline{1-7}[3pt/2pt]
\addlinespace[2pt]
\multirow{2}{*}{\method{}}
& Automated
& 92.22 & 94.44 & 90.00 & 86.67 & 7.78 (7) \\
& Human
& \textbf{92.22} & 90.00 & \textbf{91.11}
& \textbf{84.44} & \textbf{5.56 (5)} \\
\bottomrule
\end{tabular*}
\end{table}

Across methods, automated and human TRR are \(83.61\%\) and \(81.39\%\), respectively, giving a difference of \(+2.22\) percentage points (\(95\%\) CI: \([0.83,\,3.61]\)). Automated LGG is \(10.28\%\) (\(37\) responses), versus \(8.89\%\) (\(32\) responses) under human labels, a difference of \(+1.39\) points (\([0.00,\,2.78]\)). For \method{}, the corresponding differences are \(+2.22\) points for both TRR and LGG. The pooled TRR interval lies above zero, whereas the pooled LGG interval includes zero.

\begin{table}[htbp]
\centering
\small
\setlength{\tabcolsep}{4pt}
\renewcommand{\arraystretch}{1.12}
\caption{\textbf{Automated minus human metric differences.}
Values are percentage-point differences with \(95\%\) paired bootstrap confidence intervals.
The pooled row averages all four methods.}
\label{tab:appendix_human_audit_differences}
\vspace{4pt}

\begin{tabular*}{\linewidth}{@{\extracolsep{\fill}}lcc@{}}
\toprule
\textbf{Method}
& \(\boldsymbol{\Delta\mathrm{TRR}}\)
& \(\boldsymbol{\Delta\mathrm{LGG}}\) \\
\midrule
SFT
& \(+1.11\;[0.00,\,+3.33]\)
& \(0.00\;[0.00,\,0.00]\) \\

CF-SFT
& \(+4.44\;[+1.11,\,+8.89]\)
& \(+2.22\;[0.00,\,+5.56]\) \\

RM
& \(+1.11\;[-2.22,\,+4.44]\)
& \(+1.11\;[-2.22,\,+4.44]\) \\

\method{}
& \(+2.22\;[0.00,\,+5.56]\)
& \(+2.22\;[0.00,\,+5.56]\) \\

\addlinespace[2pt]
\cdashline{1-3}[3pt/2pt]
\addlinespace[2pt]
All methods
& \(+2.22\;[+0.83,\,+3.61]\)
& \(+1.39\;[0.00,\,+2.78]\) \\
\bottomrule
\end{tabular*}
\end{table}

\textbf{Method comparison under human labels.}
Under human labels, \method{} achieves a TRR of \(84.44\%\) and an LGG of \(5.56\%\) (\(5\) responses), the highest and lowest point estimates, respectively. Its paired TRR gain is \(6.67\) points over CF-SFT (\(95\%\) CI: \([1.11,\,12.22]\)) and \(3.33\) points over RM (\([-2.22,\,10.00]\)). The audit therefore supports a TRR improvement over CF-SFT, while the comparison with RM remains uncertain. The lower LGG point estimate relative to both baselines also persists under human review. These findings concern the audited responses from seed \(42\); variation across training seeds is assessed by the main experiments.

\subsubsection{Deterministic Matching Accuracy}
\label{app:matching_acc}

\textbf{Scoring rule.}
We apply the deterministic matcher used during training to the stored final answers on the diagnostic evaluation set. This evaluates the same predictions under an alternative scoring rule. Let \(N(a)\) denote the token sequence obtained by removing an optional leading \texttt{Final answer:} label, lowercasing, replacing characters outside ASCII letters, digits, and whitespace with spaces, removing the articles \texttt{a}, \texttt{an}, and \texttt{the}, and collapsing whitespace. Write \(u\preceq v\) when \(u\) is a contiguous whole-token span of \(v\). The per-response score is
\[
\mathrm{Match}(a,a^\star)
=
\mathbf{1}\!\left[
\begin{aligned}
&N(a)\ne\varnothing \land N(a^\star)\ne\varnothing\\
&{}\land\bigl(
N(a)\preceq N(a^\star)
\lor
N(a^\star)\preceq N(a)
\bigr)
\end{aligned}
\right].
\]
We denote its mean under the aggregation above by \(\mathrm{Acc}_{\mathrm{match}}\). Only the stored prediction and reference answer are compared, without aliases or manual overrides. Although recorded as \texttt{acc\_em} in the implementation, this bidirectional span-matching score differs from strict exact match. Development checkpoint selection uses the same normalizer and pairwise matcher but additionally accepts available aliases and evaluates original and counterfactual inputs with a different generation length limit.

\textbf{Sensitivity to the scoring rule.}
Table~\ref{tab:appendix_answer_accuracy} compares semantic and deterministic accuracy. \method{} retains the highest aggregate point estimate under deterministic matching, reaching \(88.44\%\), compared with \(87.16\%\) for RM and \(84.96\%\) for CF-SFT. Its gains are \(1.27\) and \(3.48\) percentage points, respectively, while its semantic-accuracy gain over RM is \(2.28\) points. Across individual settings, \method{} exceeds RM in five of nine settings and CF-SFT in eight, with one tie (Table~\ref{tab:appendix_training_accuracy_settings}). The aggregate accuracy advantage persists under deterministic matching, supporting its robustness to the scoring rule, while the magnitude and setting-level gains vary.

Deterministic matching can reject semantic equivalents or accept incomplete answers. In the diagnostic responses, \texttt{East Germany} versus \texttt{GDR} fails matching but passes semantic correctness, whereas \texttt{Latin} versus \texttt{Medieval Latin} passes matching but fails semantic correctness. The disagreement columns report predictions accepted by only one scoring rule, using the existing semantic labels.

\begin{table}[h]
\centering
\footnotesize
\setlength{\tabcolsep}{3.5pt}
\renewcommand{\arraystretch}{1.12}
\caption{\textbf{Answer accuracy under two scoring rules.}
Values are percentages, averaged over seeds and then equally across the nine settings.
Match only and Semantic only report the proportions accepted exclusively by the corresponding rule, using the same weighting.}
\label{tab:appendix_answer_accuracy}
\vspace{4pt}

\begin{tabular*}{\linewidth}{@{\extracolsep{\fill}}lcccc@{}}
\toprule
\textbf{Method}
& \(\mathrm{Acc}_{\mathrm{sem}}\)
& \(\mathrm{Acc}_{\mathrm{match}}\)
& \textbf{Match only}
& \textbf{Semantic only} \\
\midrule

Backbone
& 79.75 & 74.73 & 0.45 & 5.47 \\

SFT
& 87.76 & 84.21 & 0.51 & 4.06 \\

CF-SFT
& 88.66 & 84.96 & 0.36 & 4.06 \\

RM
& 90.51 & 87.16 & 0.47 & 3.81 \\

\method{}
& 92.79 & 88.44 & 0.35 & 4.70 \\

\bottomrule
\end{tabular*}
\end{table}

\begin{table}[h]
\centering
\footnotesize
\setlength{\tabcolsep}{3.5pt}
\renewcommand{\arraystretch}{1.12}
\caption{\textbf{Deterministic answer accuracy by dataset and backbone (\%).}
Adapted methods report means \(\pm\) sample standard deviations over seeds \(42\), \(43\), and \(44\).
Backbone results use the fixed diagnostic predictions.}
\label{tab:appendix_training_accuracy_settings}
\vspace{4pt}

\begin{tabular*}{\linewidth}{@{\extracolsep{\fill}}clccc@{}}
\toprule
\textbf{Model}
& \textbf{Method}
& \textbf{2WikiMultiHopQA}
& \textbf{MuSiQue}
& \textbf{HotpotQA} \\
\midrule

\multirow{5}{*}{\textit{Qwen3-8B}}
& Backbone
& 92.45 & 50.20 & 84.98 \\

& SFT
& \(98.44\pm0.55\)
& \(65.88\pm0.68\)
& \(81.98\pm1.08\) \\

& CF-SFT
& \(98.92\pm0.95\)
& \(69.02\pm0.39\)
& \(82.58\pm0.79\) \\

& RM
& \(98.56\pm0.00\)
& \(70.98\pm3.59\)
& \(87.49\pm1.21\) \\

& \method{}
& \(99.52\pm0.21\)
& \(76.60\pm3.43\)
& \(87.29\pm0.63\) \\

\addlinespace[3pt]
\cdashline{1-5}[3pt/2pt]
\addlinespace[3pt]

\multirow{5}{*}{\textit{Qwen3-32B}}
& Backbone
& 96.40 & 57.65 & 86.49 \\

& SFT
& \(98.68\pm0.21\)
& \(68.63\pm1.36\)
& \(87.09\pm1.04\) \\

& CF-SFT
& \(98.80\pm0.21\)
& \(71.63\pm1.48\)
& \(86.79\pm0.60\) \\

& RM
& \(99.64\pm0.36\)
& \(75.42\pm1.20\)
& \(88.99\pm0.92\) \\

& \method{}
& \(99.04\pm0.21\)
& \(79.61\pm3.86\)
& \(89.29\pm1.76\) \\

\addlinespace[3pt]
\cdashline{1-5}[3pt/2pt]
\addlinespace[3pt]

\multirow{5}{*}{\textit{Ministral-3-8B}}
& Backbone
& 84.17 & 41.57 & 78.68 \\

& SFT
& \(99.52\pm0.21\)
& \(72.29\pm0.99\)
& \(85.39\pm2.00\) \\

& CF-SFT
& \(99.28\pm0.62\)
& \(72.03\pm2.61\)
& \(85.59\pm1.59\) \\

& RM
& \(99.64\pm0.36\)
& \(74.77\pm3.54\)
& \(88.99\pm0.17\) \\

& \method{}
& \(99.28\pm0.36\)
& \(77.91\pm1.48\)
& \(87.39\pm0.79\) \\

\bottomrule
\end{tabular*}
\end{table}


\subsubsection{Per-Seed Results}
\label{app:per_seed_results}

Tables~\ref{tab:appendix_seeds_qwen3-8b}--\ref{tab:appendix_seeds_ministral-3-8b} report both answer accuracy measures and the main reasoning metrics for each training seed. Here, \(\mathrm{Acc}_{\mathrm{match}}\) uses deterministic matching, while \(\mathrm{Acc}_{\mathrm{sem}}\) uses semantic correctness labels. All metrics are percentages; higher values are better except for LGG. The standard deviations in Table~\ref{tab:appendix_training_accuracy_settings} summarize variation across training seeds, not uncertainty from question sampling.

Backbone results use the fixed diagnostic generations. All responses are scored by the same automated evaluator described in Appendix~\ref{app:evaluator}. Generation uses greedy decoding with at most \(1{,}024\) new tokens and an \(8{,}192\)-token sequence limit; explicit thinking mode is disabled where supported. Appendix~\ref{app:training_implementation} documents checkpoint selection.


\begin{table}[p]
\centering
\fontsize{10}{11.5}\selectfont
\setlength{\tabcolsep}{3pt}
\renewcommand{\arraystretch}{1.08}
\caption{\textbf{Per-seed results for Qwen3-8B (\%).}
Dataset headings give the number of questions.
Fixed denotes the untuned backbone's fixed diagnostic generations.}
\label{tab:appendix_seeds_qwen3-8b}
\vspace{4pt}

\begin{tabular*}{\linewidth}{@{\extracolsep{\fill}}lccccccc@{}}
\toprule
\textbf{Method} & \textbf{Seed}
& \(\mathrm{Acc}_{\mathrm{match}}\) & \(\mathrm{Acc}_{\mathrm{sem}}\)
& \textbf{LSR} & \textbf{GSR} & \textbf{TRR} & \textbf{LGG} \\
\midrule

\multicolumn{8}{c}{\textit{2WikiMultiHopQA} (\(n=278\))} \\
\addlinespace[2pt]

Backbone & Fixed
& 92.45 & 92.45 & 95.68 & 94.24 & 90.65 & 5.04 \\
\addlinespace[2pt]

\multirow{3}{*}{SFT}
& 42 & 98.56 & 98.56 & 98.56 & 99.28 & 98.20 & 0.36 \\
& 43 & 98.92 & 98.92 & 98.20 & 99.28 & 98.20 & 0.00 \\
& 44 & 97.84 & 97.84 & 97.48 & 98.92 & 97.48 & 0.00 \\
\addlinespace[2pt]

\multirow{3}{*}{CF-SFT}
& 42 & 98.56 & 98.56 & 98.56 & 99.64 & 98.56 & 0.00 \\
& 43 & 98.20 & 98.56 & 98.20 & 98.56 & 97.84 & 0.36 \\
& 44 & 100.00 & 100.00 & 100.00 & 100.00 & 100.00 & 0.00 \\
\addlinespace[2pt]

\multirow{3}{*}{RM}
& 42 & 98.56 & 98.92 & 98.92 & 100.00 & 98.92 & 0.00 \\
& 43 & 98.56 & 99.28 & 99.28 & 100.00 & 99.28 & 0.00 \\
& 44 & 98.56 & 98.92 & 98.92 & 99.28 & 98.20 & 0.72 \\
\addlinespace[2pt]

\multirow{3}{*}{\method{}}
& 42 & 99.64 & 99.64 & 99.28 & 99.64 & 99.28 & 0.00 \\
& 43 & 99.28 & 99.64 & 99.28 & 99.64 & 99.28 & 0.00 \\
& 44 & 99.64 & 100.00 & 100.00 & 100.00 & 100.00 & 0.00 \\

\addlinespace[3pt]
\cdashline{1-8}[3pt/2pt]
\addlinespace[3pt]

\multicolumn{8}{c}{\textit{MuSiQue} (\(n=255\))} \\
\addlinespace[2pt]

Backbone & Fixed
& 50.20 & 55.29 & 43.14 & 71.76 & 32.16 & 10.98 \\
\addlinespace[2pt]

\multirow{3}{*}{SFT}
& 42 & 65.49 & 69.41 & 87.84 & 65.88 & 61.96 & 25.88 \\
& 43 & 65.49 & 69.80 & 81.57 & 58.04 & 50.59 & 30.98 \\
& 44 & 66.67 & 70.98 & 74.51 & 63.53 & 53.33 & 21.18 \\
\addlinespace[2pt]

\multirow{3}{*}{CF-SFT}
& 42 & 68.63 & 73.73 & 87.84 & 68.24 & 65.88 & 21.96 \\
& 43 & 69.41 & 73.73 & 80.39 & 60.78 & 54.90 & 25.49 \\
& 44 & 69.02 & 73.33 & 78.82 & 59.61 & 53.73 & 25.10 \\
\addlinespace[2pt]

\multirow{3}{*}{RM}
& 42 & 67.06 & 71.37 & 77.25 & 56.47 & 48.63 & 28.63 \\
& 43 & 74.12 & 75.69 & 78.82 & 57.25 & 50.59 & 28.24 \\
& 44 & 71.76 & 74.90 & 81.57 & 57.65 & 49.41 & 32.16 \\
\addlinespace[2pt]

\multirow{3}{*}{\method{}}
& 42 & 75.69 & 80.78 & 88.24 & 76.86 & 70.59 & 17.65 \\
& 43 & 73.73 & 80.39 & 80.78 & 69.41 & 60.00 & 20.78 \\
& 44 & 80.39 & 82.35 & 83.53 & 72.55 & 67.06 & 16.47 \\

\addlinespace[3pt]
\cdashline{1-8}[3pt/2pt]
\addlinespace[3pt]

\multicolumn{8}{c}{\textit{HotpotQA} (\(n=333\))} \\
\addlinespace[2pt]

Backbone & Fixed
& 84.98 & 91.89 & 88.89 & 95.20 & 84.98 & 3.90 \\
\addlinespace[2pt]

\multirow{3}{*}{SFT}
& 42 & 81.68 & 88.29 & 96.40 & 90.69 & 88.29 & 8.11 \\
& 43 & 81.08 & 89.19 & 93.39 & 89.49 & 84.98 & 8.41 \\
& 44 & 83.18 & 90.39 & 93.09 & 90.99 & 86.49 & 6.61 \\
\addlinespace[2pt]

\multirow{3}{*}{CF-SFT}
& 42 & 82.88 & 90.09 & 95.50 & 93.39 & 89.79 & 5.71 \\
& 43 & 81.68 & 88.89 & 92.79 & 91.59 & 86.19 & 6.61 \\
& 44 & 83.18 & 90.09 & 93.99 & 91.59 & 87.39 & 6.61 \\
\addlinespace[2pt]

\multirow{3}{*}{RM}
& 42 & 86.79 & 92.49 & 91.89 & 91.89 & 86.79 & 5.11 \\
& 43 & 88.89 & 95.20 & 94.29 & 93.39 & 89.79 & 4.50 \\
& 44 & 86.79 & 93.09 & 92.79 & 93.99 & 88.29 & 4.50 \\
\addlinespace[2pt]

\multirow{3}{*}{\method{}}
& 42 & 86.79 & 95.20 & 97.00 & 97.30 & 94.59 & 2.40 \\
& 43 & 87.09 & 94.59 & 94.89 & 96.40 & 91.59 & 3.30 \\
& 44 & 87.99 & 96.10 & 97.00 & 96.40 & 94.59 & 2.40 \\

\bottomrule
\end{tabular*}
\end{table}


\begin{table}[p]
\centering
\fontsize{10}{11.5}\selectfont
\setlength{\tabcolsep}{3pt}
\renewcommand{\arraystretch}{1.08}
\caption{\textbf{Per-seed results for Qwen3-32B (\%).}
Dataset headings give the number of questions.
Fixed denotes the untuned backbone's fixed diagnostic generations.}
\label{tab:appendix_seeds_qwen3-32b}
\vspace{4pt}

\begin{tabular*}{\linewidth}{@{\extracolsep{\fill}}lccccccc@{}}
\toprule
\textbf{Method} & \textbf{Seed}
& \(\mathrm{Acc}_{\mathrm{match}}\) & \(\mathrm{Acc}_{\mathrm{sem}}\)
& \textbf{LSR} & \textbf{GSR} & \textbf{TRR} & \textbf{LGG} \\
\midrule

\multicolumn{8}{c}{\textit{2WikiMultiHopQA} (\(n=278\))} \\
\addlinespace[2pt]

Backbone & Fixed
& 96.40 & 96.40 & 96.40 & 99.64 & 96.04 & 0.36 \\
\addlinespace[2pt]

\multirow{3}{*}{SFT}
& 42 & 98.92 & 98.92 & 98.56 & 100.00 & 98.56 & 0.00 \\
& 43 & 98.56 & 98.56 & 98.20 & 99.64 & 98.20 & 0.00 \\
& 44 & 98.56 & 98.92 & 98.92 & 100.00 & 98.92 & 0.00 \\
\addlinespace[2pt]

\multirow{3}{*}{CF-SFT}
& 42 & 98.56 & 98.56 & 98.56 & 100.00 & 98.56 & 0.00 \\
& 43 & 98.92 & 98.92 & 98.56 & 100.00 & 98.56 & 0.00 \\
& 44 & 98.92 & 98.92 & 98.92 & 100.00 & 98.92 & 0.00 \\
\addlinespace[2pt]

\multirow{3}{*}{RM}
& 42 & 100.00 & 100.00 & 99.64 & 100.00 & 99.64 & 0.00 \\
& 43 & 99.28 & 99.28 & 99.28 & 100.00 & 99.28 & 0.00 \\
& 44 & 99.64 & 99.64 & 99.64 & 100.00 & 99.64 & 0.00 \\
\addlinespace[2pt]

\multirow{3}{*}{\method{}}
& 42 & 98.92 & 98.92 & 98.92 & 100.00 & 98.92 & 0.00 \\
& 43 & 98.92 & 98.92 & 98.92 & 100.00 & 98.92 & 0.00 \\
& 44 & 99.28 & 99.28 & 99.28 & 100.00 & 99.28 & 0.00 \\

\addlinespace[3pt]
\cdashline{1-8}[3pt/2pt]
\addlinespace[3pt]

\multicolumn{8}{c}{\textit{MuSiQue} (\(n=255\))} \\
\addlinespace[2pt]

Backbone & Fixed
& 57.65 & 65.88 & 58.43 & 79.61 & 52.55 & 5.88 \\
\addlinespace[2pt]

\multirow{3}{*}{SFT}
& 42 & 69.41 & 72.94 & 84.71 & 63.14 & 56.86 & 27.84 \\
& 43 & 69.41 & 72.55 & 80.00 & 64.31 & 56.47 & 23.53 \\
& 44 & 67.06 & 71.37 & 81.96 & 63.92 & 56.86 & 25.10 \\
\addlinespace[2pt]

\multirow{3}{*}{CF-SFT}
& 42 & 70.59 & 74.51 & 87.45 & 66.27 & 60.00 & 27.45 \\
& 43 & 70.98 & 73.73 & 85.49 & 63.92 & 57.25 & 28.24 \\
& 44 & 73.33 & 76.08 & 84.31 & 71.37 & 62.75 & 21.57 \\
\addlinespace[2pt]

\multirow{3}{*}{RM}
& 42 & 76.47 & 78.82 & 83.92 & 59.22 & 51.76 & 32.16 \\
& 43 & 75.69 & 79.61 & 82.35 & 63.92 & 53.73 & 28.63 \\
& 44 & 74.12 & 76.86 & 82.75 & 61.18 & 53.73 & 29.02 \\
\addlinespace[2pt]

\multirow{3}{*}{\method{}}
& 42 & 75.29 & 80.39 & 91.37 & 78.43 & 76.08 & 15.29 \\
& 43 & 80.78 & 86.27 & 90.59 & 83.92 & 80.39 & 10.20 \\
& 44 & 82.75 & 86.67 & 92.16 & 84.31 & 81.18 & 10.98 \\

\addlinespace[3pt]
\cdashline{1-8}[3pt/2pt]
\addlinespace[3pt]

\multicolumn{8}{c}{\textit{HotpotQA} (\(n=333\))} \\
\addlinespace[2pt]

Backbone & Fixed
& 86.49 & 94.89 & 94.59 & 96.70 & 91.89 & 2.70 \\
\addlinespace[2pt]

\multirow{3}{*}{SFT}
& 42 & 86.49 & 94.29 & 99.10 & 92.79 & 92.49 & 6.61 \\
& 43 & 88.29 & 95.50 & 98.50 & 94.29 & 93.39 & 5.11 \\
& 44 & 86.49 & 93.99 & 97.60 & 94.29 & 92.19 & 5.41 \\
\addlinespace[2pt]

\multirow{3}{*}{CF-SFT}
& 42 & 86.19 & 93.99 & 98.20 & 93.09 & 91.59 & 6.61 \\
& 43 & 87.39 & 94.59 & 99.10 & 93.39 & 92.49 & 6.61 \\
& 44 & 86.79 & 94.89 & 96.40 & 94.89 & 92.19 & 4.20 \\
\addlinespace[2pt]

\multirow{3}{*}{RM}
& 42 & 89.79 & 96.70 & 97.90 & 96.70 & 94.89 & 3.00 \\
& 43 & 87.99 & 96.70 & 97.30 & 95.20 & 93.39 & 3.90 \\
& 44 & 89.19 & 97.00 & 97.30 & 97.00 & 95.20 & 2.10 \\
\addlinespace[2pt]

\multirow{3}{*}{\method{}}
& 42 & 87.99 & 97.90 & 99.10 & 98.50 & 97.90 & 1.20 \\
& 43 & 88.59 & 98.20 & 99.70 & 98.50 & 98.20 & 1.50 \\
& 44 & 91.29 & 98.80 & 99.40 & 99.10 & 98.80 & 0.60 \\

\bottomrule
\end{tabular*}
\end{table}


\begin{table}[p]
\centering
\fontsize{10}{11.5}\selectfont
\setlength{\tabcolsep}{3pt}
\renewcommand{\arraystretch}{1.08}
\caption{\textbf{Per-seed results for Ministral-3-8B (\%).}
Dataset headings give the number of questions.
Fixed denotes the untuned backbone's fixed diagnostic generations.}
\label{tab:appendix_seeds_ministral-3-8b}
\vspace{4pt}

\begin{tabular*}{\linewidth}{@{\extracolsep{\fill}}lccccccc@{}}
\toprule
\textbf{Method} & \textbf{Seed}
& \(\mathrm{Acc}_{\mathrm{match}}\) & \(\mathrm{Acc}_{\mathrm{sem}}\)
& \textbf{LSR} & \textbf{GSR} & \textbf{TRR} & \textbf{LGG} \\
\midrule

\multicolumn{8}{c}{\textit{2WikiMultiHopQA} (\(n=278\))} \\
\addlinespace[2pt]

Backbone & Fixed
& 84.17 & 84.17 & 91.01 & 87.77 & 80.94 & 10.07 \\
\addlinespace[2pt]

\multirow{3}{*}{SFT}
& 42 & 99.28 & 99.28 & 99.28 & 100.00 & 99.28 & 0.00 \\
& 43 & 99.64 & 99.64 & 99.64 & 100.00 & 99.64 & 0.00 \\
& 44 & 99.64 & 99.64 & 99.64 & 100.00 & 99.64 & 0.00 \\
\addlinespace[2pt]

\multirow{3}{*}{CF-SFT}
& 42 & 99.64 & 100.00 & 99.64 & 99.64 & 99.64 & 0.00 \\
& 43 & 99.64 & 100.00 & 100.00 & 100.00 & 100.00 & 0.00 \\
& 44 & 98.56 & 98.92 & 98.56 & 100.00 & 98.56 & 0.00 \\
\addlinespace[2pt]

\multirow{3}{*}{RM}
& 42 & 99.64 & 100.00 & 100.00 & 100.00 & 100.00 & 0.00 \\
& 43 & 99.28 & 99.64 & 100.00 & 99.64 & 99.64 & 0.36 \\
& 44 & 100.00 & 100.00 & 100.00 & 100.00 & 100.00 & 0.00 \\
\addlinespace[2pt]

\multirow{3}{*}{\method{}}
& 42 & 98.92 & 99.64 & 100.00 & 99.64 & 99.64 & 0.36 \\
& 43 & 99.64 & 100.00 & 100.00 & 100.00 & 100.00 & 0.00 \\
& 44 & 99.28 & 99.64 & 99.64 & 100.00 & 99.64 & 0.00 \\

\addlinespace[3pt]
\cdashline{1-8}[3pt/2pt]
\addlinespace[3pt]

\multicolumn{8}{c}{\textit{MuSiQue} (\(n=255\))} \\
\addlinespace[2pt]

Backbone & Fixed
& 41.57 & 49.41 & 30.20 & 65.49 & 24.71 & 5.49 \\
\addlinespace[2pt]

\multirow{3}{*}{SFT}
& 42 & 73.33 & 76.47 & 82.75 & 67.84 & 60.39 & 22.35 \\
& 43 & 71.37 & 74.12 & 89.02 & 68.63 & 64.31 & 24.71 \\
& 44 & 72.16 & 74.90 & 86.67 & 60.39 & 55.69 & 30.98 \\
\addlinespace[2pt]

\multirow{3}{*}{CF-SFT}
& 42 & 73.73 & 76.86 & 85.49 & 66.67 & 61.57 & 23.92 \\
& 43 & 69.02 & 74.51 & 91.37 & 69.41 & 67.45 & 23.92 \\
& 44 & 73.33 & 76.08 & 88.24 & 63.53 & 58.82 & 29.41 \\
\addlinespace[2pt]

\multirow{3}{*}{RM}
& 42 & 74.51 & 78.82 & 85.88 & 69.02 & 60.78 & 25.10 \\
& 43 & 71.37 & 74.90 & 87.06 & 58.82 & 52.55 & 34.51 \\
& 44 & 78.43 & 80.00 & 92.55 & 57.25 & 53.73 & 38.82 \\
\addlinespace[2pt]

\multirow{3}{*}{\method{}}
& 42 & 76.86 & 82.35 & 89.41 & 78.43 & 72.55 & 16.86 \\
& 43 & 79.61 & 82.75 & 90.20 & 78.82 & 74.90 & 15.29 \\
& 44 & 77.25 & 80.78 & 90.98 & 72.55 & 67.45 & 23.53 \\

\addlinespace[3pt]
\cdashline{1-8}[3pt/2pt]
\addlinespace[3pt]

\multicolumn{8}{c}{\textit{HotpotQA} (\(n=333\))} \\
\addlinespace[2pt]

Backbone & Fixed
& 78.68 & 87.39 & 80.78 & 90.69 & 76.28 & 4.50 \\
\addlinespace[2pt]

\multirow{3}{*}{SFT}
& 42 & 85.89 & 92.79 & 94.29 & 93.99 & 89.19 & 5.11 \\
& 43 & 83.18 & 89.79 & 93.39 & 93.09 & 87.09 & 6.31 \\
& 44 & 87.09 & 92.49 & 96.10 & 94.59 & 91.29 & 4.80 \\
\addlinespace[2pt]

\multirow{3}{*}{CF-SFT}
& 42 & 86.79 & 93.69 & 96.10 & 95.20 & 91.89 & 4.20 \\
& 43 & 86.19 & 92.49 & 95.80 & 93.69 & 90.09 & 5.71 \\
& 44 & 83.78 & 90.09 & 96.10 & 91.29 & 88.29 & 7.81 \\
\addlinespace[2pt]

\multirow{3}{*}{RM}
& 42 & 89.19 & 94.59 & 95.50 & 95.80 & 92.19 & 3.30 \\
& 43 & 88.89 & 96.40 & 95.80 & 95.50 & 92.49 & 3.30 \\
& 44 & 88.89 & 94.89 & 97.00 & 93.09 & 90.99 & 6.01 \\
\addlinespace[2pt]

\multirow{3}{*}{\method{}}
& 42 & 87.09 & 95.20 & 96.10 & 96.40 & 93.09 & 3.00 \\
& 43 & 86.79 & 94.89 & 95.80 & 96.70 & 93.09 & 2.70 \\
& 44 & 88.29 & 96.10 & 97.90 & 97.30 & 95.20 & 2.70 \\

\bottomrule
\end{tabular*}
\end{table}

\providecommand{\ablci}[3]{%
\shortstack{\(#1\)\\[-1pt]{\scriptsize\([#2,\,#3]\)}}%
}

\subsection{Objective Ablations}
\label{app:ablations}

\subsubsection{Ablation Setup}

We conduct all ablations on Qwen3-8B using training seeds \(\{42,43,44\}\). Removing \(S\), \(T\), or \(C\) drops both directional margins for that dependency. The one-sided variant retains all three dependencies but only their original-condition margins. All variants share the generation anchors, training data, backbone initialization, scheduled optimization budget, and checkpoint-selection rule.

To preserve the total margin coefficient \(\lambda\), retained terms are scaled by \(3/2\) after dependency removal and by \(2\) in the one-sided variant. The resulting objectives contain four and three margin terms, respectively, compared with six in the full objective.

\subsubsection{Detailed Results}

Following the main ablation analysis, we pool all \(866\) questions within each seed and then average over the three seeds. This weights questions equally, whereas the main comparison in Appendix~\ref{app:full_results} weights datasets equally. For example, the full objective achieves pooled TRR of \(87.34\%\), compared with \(86.33\%\) under equal dataset weights. Tables~\ref{tab:appendix_ablation_seeds} and~\ref{tab:appendix_ablation_datasets} report per-seed and dataset-specific results, respectively. Under deterministic answer matching, the full objective achieves \(88.07\%\) pooled accuracy, exceeding all four ablations; the closest variant is w/o \(C\) at \(86.84\%\).

\begin{table}[t]
\centering
\fontsize{8.5}{9}\selectfont
\setlength{\tabcolsep}{3pt}
\renewcommand{\arraystretch}{1.12}
\caption{\textbf{Per-seed objective ablations on Qwen3-8B.}
All values are percentages over the \(866\) questions pooled within each seed.
\(\mathrm{Acc}_{\mathrm{match}}\) uses deterministic matching, while \(\mathrm{Acc}_{\mathrm{sem}}\) uses semantic correctness labels.}
\label{tab:appendix_ablation_seeds}
\vspace{4pt}

\begin{tabular*}{\linewidth}{@{\extracolsep{\fill}}lccccccccc@{}}
\toprule
\textbf{Variant} & \textbf{Seed}
& \(\mathrm{Acc}_{\mathrm{match}}\)
& \(\mathrm{Acc}_{\mathrm{sem}}\)
& \textbf{LSR} & \(A\) & \(C\)
& \textbf{GSR} & \textbf{TRR} & \textbf{LGG} \\
\midrule

\multirow{3}{*}{w/o \(S\)}
& 42 & 84.64 & 87.99 & 90.30 & 92.73 & 95.27 & 87.99 & 83.49 & 6.81 \\
& 43 & 84.06 & 88.11 & 88.45 & 93.19 & 94.80 & 87.99 & 81.87 & 6.58 \\
& 44 & 86.03 & 89.84 & 91.45 & 94.57 & 94.46 & 89.03 & 83.83 & 7.62 \\

\addlinespace[3pt]
\cdashline{1-10}[3pt/2pt]
\addlinespace[3pt]

\multirow{3}{*}{w/o \(T\)}
& 42 & 82.33 & 87.07 & 91.22 & 91.34 & 92.84 & 84.64 & 81.18 & 10.05 \\
& 43 & 81.76 & 85.57 & 91.69 & 92.03 & 92.38 & 84.99 & 80.60 & 11.09 \\
& 44 & 83.26 & 87.30 & 91.92 & 93.07 & 92.15 & 85.57 & 81.87 & 10.05 \\

\addlinespace[3pt]
\cdashline{1-10}[3pt/2pt]
\addlinespace[3pt]

\multirow{3}{*}{w/o \(C\)}
& 42 & 87.30 & 90.42 & 90.88 & 94.69 & 93.88 & 89.03 & 82.56 & 8.31 \\
& 43 & 87.76 & 90.30 & 89.38 & 94.46 & 93.88 & 88.57 & 82.45 & 6.93 \\
& 44 & 85.45 & 89.15 & 91.92 & 94.00 & 93.88 & 88.45 & 83.83 & 8.08 \\

\addlinespace[3pt]
\cdashline{1-10}[3pt/2pt]
\addlinespace[3pt]

\multirow{3}{*}{One-sided}
& 42 & 85.91 & 90.42 & 88.11 & 93.65 & 92.84 & 86.72 & 80.48 & 7.62 \\
& 43 & 84.06 & 88.57 & 88.11 & 93.65 & 92.84 & 86.72 & 79.79 & 8.31 \\
& 44 & 85.68 & 89.49 & 91.45 & 94.46 & 90.18 & 85.57 & 81.06 & 10.39 \\

\addlinespace[3pt]
\cdashline{1-10}[3pt/2pt]
\addlinespace[3pt]

\multirow{3}{*}{\method{}}
& 42 & 87.64 & 92.38 & 95.15 & 95.96 & 95.61 & 92.03 & 89.03 & 6.12 \\
& 43 & 87.07 & 92.03 & 92.15 & 94.92 & 94.46 & 89.49 & 84.76 & 7.39 \\
& 44 & 89.49 & 93.30 & 94.00 & 95.03 & 95.03 & 90.53 & 88.22 & 5.77 \\

\bottomrule
\end{tabular*}
\end{table}

\begin{table}[t]
\centering
\fontsize{8.5}{10}\selectfont
\setlength{\tabcolsep}{3pt}
\renewcommand{\arraystretch}{1.12}
\caption{\textbf{Dataset-specific ablations on Qwen3-8B.}
Values are three-seed mean percentages.
Each dataset reports deterministic matching accuracy, TRR, and LGG.}
\label{tab:appendix_ablation_datasets}
\vspace{4pt}

\begin{tabular*}{\linewidth}{@{\extracolsep{\fill}}lccccccccc@{}}
\toprule
\multirow{2}{*}{\textbf{Variant}}
& \multicolumn{3}{c}{\textbf{2WikiMultiHopQA}}
& \multicolumn{3}{c}{\textbf{MuSiQue}}
& \multicolumn{3}{c}{\textbf{HotpotQA}} \\
\cmidrule(lr){2-4}
\cmidrule(lr){5-7}
\cmidrule(lr){8-10}
& \(\mathrm{Acc}_{\mathrm{match}}\) & \textbf{TRR} & \textbf{LGG}
& \(\mathrm{Acc}_{\mathrm{match}}\) & \textbf{TRR} & \textbf{LGG}
& \(\mathrm{Acc}_{\mathrm{match}}\) & \textbf{TRR} & \textbf{LGG} \\
\midrule

w/o \(S\)
& 99.04 & 98.56 & 0.12
& 69.28 & 57.39 & 18.56
& 85.09 & 89.79 & 3.90 \\

w/o \(T\)
& 98.80 & 98.92 & 0.00
& 64.84 & 54.12 & 26.80
& 82.28 & 87.19 & 6.51 \\

w/o \(C\)
& 98.56 & 98.80 & 0.00
& 73.99 & 56.08 & 21.83
& 86.89 & 90.29 & 3.50 \\

One-sided
& 98.56 & 98.68 & 0.00
& 71.11 & 51.11 & 24.05
& 84.88 & 87.69 & 4.40 \\

\addlinespace[3pt]
\cdashline{1-10}[3pt/2pt]
\addlinespace[3pt]

\method{}
& 99.52 & 99.52 & 0.00
& 76.60 & 65.88 & 18.30
& 87.29 & 93.59 & 2.70 \\

\bottomrule
\end{tabular*}
\end{table}

\subsubsection{Failure Decomposition}

We partition LGG responses using the same first-broken-link convention as the diagnostic study. Path deviation (PD) corresponds to \(L=1,A=0\); incomplete reasoning (IR) to \(L=1,A=1,C=0,K=0\); and answer decoupling (AD) to \(L=1,A=1,C=0,K=1\). These categories are disjoint and together comprise \(L=1,G=0\). Table~\ref{tab:appendix_ablation_failures} reports their rates over all evaluated responses, so they sum to the LGG rate before rounding.

Removing \(T\) increases all three components relative to the full objective: PD rises from \(2.27\%\) to \(4.08\%\), IR from \(3.66\%\) to \(4.70\%\), and AD from \(0.50\%\) to \(1.62\%\). For the one-sided variant, the largest increase occurs in IR, from \(3.66\%\) to \(5.62\%\). These descriptive patterns show that changes associated with an objective can span multiple failure types.

\begin{table}[t]
\centering
\small
\setlength{\tabcolsep}{4pt}
\renewcommand{\arraystretch}{1.12}
\caption{\textbf{Deterministic accuracy and LGG decomposition.}
Values are percentages pooled over \(866\) questions within each seed, then averaged over three seeds.
PD, IR, and AD use all responses as the denominator and sum to LGG before rounding.}
\label{tab:appendix_ablation_failures}
\vspace{4pt}

\begin{tabular*}{\linewidth}{@{\extracolsep{\fill}}lccccc@{}}
\toprule
\textbf{Variant}
& \(\mathrm{Acc}_{\mathrm{match}}\)
& \textbf{PD}
& \textbf{IR}
& \textbf{AD}
& \textbf{LGG} \\
\midrule

w/o \(S\)
& 84.91 & 2.69 & 3.66 & 0.65 & 7.01 \\

w/o \(T\)
& 82.45 & 4.08 & 4.70 & 1.62 & 10.39 \\

w/o \(C\)
& 86.84 & 2.77 & 4.20 & 0.81 & 7.78 \\

One-sided
& 85.22 & 2.42 & 5.62 & 0.73 & 8.78 \\

\addlinespace[3pt]
\cdashline{1-6}[3pt/2pt]
\addlinespace[3pt]

\method{}
& 88.07 & 2.27 & 3.66 & 0.50 & 6.43 \\

\bottomrule
\end{tabular*}
\end{table}

\subsubsection{Paired Comparisons}

Table~\ref{tab:appendix_ablation_gains} reports paired differences with \(95\%\) percentile confidence intervals from \(10{,}000\) bootstrap resamples using seed \(2031\). Each resample draws questions from the pooled \(866\)-question set and resamples training seeds, preserving pairing across variants. The same seed draw applies across all datasets.

The full objective improves TRR over every ablation, with all four confidence intervals above zero. LGG reduction intervals also exclude zero for w/o \(T\) and one-sided supervision, while those for w/o \(S\) and w/o \(C\) include zero. The closure gain over w/o \(C\) is \(1.15\) points, with an interval of \([-0.12,\,2.35]\). Together, these comparisons support the full objective's contribution to joint trace reliability, while improvements in individual failure measures are less uniformly resolved.

\begin{table}[t]
\centering
\small
\setlength{\tabcolsep}{4pt}
\renewcommand{\arraystretch}{1.15}
\caption{\textbf{Paired gains from the full objective.}
Cells show gains in percentage points above their \(95\%\) bootstrap confidence intervals.
Metric and TRR gains are full minus ablated; LGG reduction is ablated minus full.
Positive values favor \method{}.}
\label{tab:appendix_ablation_gains}
\vspace{4pt}

\begin{tabular*}{\linewidth}{@{\extracolsep{\fill}}lcccc@{}}
\toprule
\textbf{Comparator}
& \textbf{Metric}
& \textbf{Metric gain}
& \textbf{TRR gain}
& \textbf{LGG reduction} \\
\midrule

w/o \(S\)
& LSR
& \ablci{+3.70}{+1.96}{+5.54}
& \ablci{+4.27}{+2.23}{+6.35}
& \ablci{+0.58}{-1.23}{+2.35} \\
\addlinespace[3pt]

w/o \(T\)
& \(A\)
& \ablci{+3.16}{+1.39}{+5.08}
& \ablci{+6.12}{+3.70}{+8.62}
& \ablci{+3.96}{+2.42}{+5.50} \\
\addlinespace[3pt]

w/o \(C\)
& \(C\)
& \ablci{+1.15}{-0.12}{+2.35}
& \ablci{+4.39}{+1.92}{+6.93}
& \ablci{+1.35}{-0.69}{+3.16} \\
\addlinespace[3pt]

One-sided
& \(\mathrm{Acc}_{\mathrm{sem}}\)
& \ablci{+3.08}{+1.50}{+4.70}
& \ablci{+6.89}{+4.43}{+9.31}
& \ablci{+2.35}{+0.23}{+4.73} \\

\bottomrule
\end{tabular*}
\end{table}

\subsection{Dependency Interventions}
\label{app:interventions}

\subsubsection{Evaluation Protocol}

\textbf{Development set.}
We evaluate \(286\) development sources: \(128\) from 2WikiMultiHopQA, \(56\) from MuSiQue, and \(102\) from HotpotQA. These sources are excluded from gradient updates but used for checkpoint selection based on weighted development answer accuracy. Intervention scores do not enter selection, so we report this analysis as development-set diagnostics. Three fixed backbones and \(36\) adapted checkpoints yield \(39\) scoring runs on the same sources.

\textbf{Controlled inputs.}
Each source provides a validated original response, evidence and question counterfactuals, and the first affected step under the evidence edit. When multiple evidence edits are available, we select the first under a fixed ordering by passage identifier and serialized edit content. The selected edit is shared across all checkpoints.

For \(S\), we change the evidence while holding the question and shared trace prefix fixed, comparing the original and counterfactual affected steps. For \(T\), we change the question while holding the evidence fixed, comparing the original and question-counterfactual traces without their final answers. For \(C\), we change the supplied trace while holding the original question and evidence fixed, comparing the original and evidence-counterfactual answers. Thus, \(C\)-switching measures whether answer preferences follow the supplied trace with external evidence held fixed.

\textbf{Scoring.}
For a candidate completion \(y\) under serialized input \(x\), we compute its token-normalized log likelihood with teacher forcing:
{\footnotesize
\[
f_\theta(y\mid x)
=
\frac{1}{|y|}
\sum_{t=1}^{|y|}
\log\pi_\theta(y_t\mid x,y_{<t}).
\]
}For each dependency \(d\in\{S,T,C\}\), let \(x_o,x_c\) denote the original and counterfactual inputs, and \(y_o,y_c\) their corresponding preferred candidates. We define
{\footnotesize
\[
\Delta_o^d = f_\theta(y_o\mid x_o)-f_\theta(y_c\mid x_o),
\quad
\Delta_c^d = f_\theta(y_c\mid x_c)-f_\theta(y_o\mid x_c),
\quad
\mathrm{Switch}_d = \mathbf{1}\!\left[\Delta_o^d>0\land\Delta_c^d>0\right].
\]
}Successful switching requires positive margins in both directions; ties count as failures.
Evaluation uses a zero threshold, distinct from the training target margin \(\gamma=0.5\).
Twelve candidate scores determine six directional margins without response generation or LLM judging.
For adapted methods, rates are averaged over three seeds per dataset--backbone setting, then equally across nine settings.

\subsubsection{Directional Results}

Table~\ref{tab:appendix_intervention_directions} reports success in each direction and the joint switch rate. For \(C\), RM prefers the original answer under the original trace on \(100\%\) of pairs but prefers the counterfactual answer under the edited trace on only \(30.41\%\). Its \(69.59\)-point switch-rate gap relative to \method{} therefore arises entirely in the counterfactual direction. For \(T\), \method{} achieves \(86.02\%\) original-direction success, \(98.39\%\) counterfactual-direction success, and \(84.41\%\) joint switching. Its remaining \(T\)-switch failures are concentrated in the original-conditioning direction.

\begin{table}[!h]
\centering
\fontsize{8.5}{9}\selectfont
\setlength{\tabcolsep}{3pt}
\renewcommand{\arraystretch}{1.12}
\caption{\textbf{Directional preference success and joint switching (\%).}
O and CF denote positive original and counterfactual margins; SW requires both on the same pair.
Rates are equally averaged across nine settings after seed averaging for adapted methods.}
\label{tab:appendix_intervention_directions}
\vspace{4pt}

\begin{tabular*}{\linewidth}{@{\extracolsep{\fill}}lccccccccc@{}}
\toprule
\multirow{2}{*}{\textbf{Method}}
& \multicolumn{3}{c}{\(S\)}
& \multicolumn{3}{c}{\(T\)}
& \multicolumn{3}{c}{\(C\)} \\
\cmidrule(lr){2-4}
\cmidrule(lr){5-7}
\cmidrule(lr){8-10}
& \textbf{O} & \textbf{CF} & \textbf{SW}
& \textbf{O} & \textbf{CF} & \textbf{SW}
& \textbf{O} & \textbf{CF} & \textbf{SW} \\
\midrule

Backbone
& 99.60 & 98.44 & 98.24
& 66.90 & 82.64 & 49.54
& 94.90 & 79.73 & 74.64 \\

SFT
& 100.00 & 99.68 & 99.68
& 85.00 & 94.36 & 79.36
& 97.51 & 89.13 & 86.63 \\

CF-SFT
& 100.00 & 99.94 & 99.94
& 83.53 & 96.34 & 79.87
& 97.70 & 92.24 & 89.94 \\

RM
& 100.00 & 99.91 & 99.91
& 83.68 & 95.75 & 79.42
& 100.00 & 30.41 & 30.41 \\

\addlinespace[3pt]
\cdashline{1-10}[3pt/2pt]
\addlinespace[3pt]

\method{}
& 100.00 & 99.91 & 99.91
& 86.02 & 98.39 & 84.41
& 100.00 & 100.00 & 100.00 \\

\bottomrule
\end{tabular*}
\end{table}

\subsubsection{Per-Setting Results}

Table~\ref{tab:appendix_intervention_settings} reports switch rates for each dataset--backbone setting. \method{} achieves \(100\%\) \(C\)-switching across all \(2{,}574\) source--checkpoint evaluations, comprising the same \(286\) sources evaluated under nine backbone--seed checkpoints. Mean \(T\)-switch rates are \(59.90\%\) on 2WikiMultiHopQA, \(94.64\%\) on MuSiQue, and \(98.69\%\) on HotpotQA. Question-guided trace switching therefore remains less consistent on 2WikiMultiHopQA, while \(S\)-switching is already near ceiling across the fine-tuned methods.

\begin{table}[h]
\centering
\fontsize{8.5}{9}\selectfont
\setlength{\tabcolsep}{3pt}
\renewcommand{\arraystretch}{1.12}
\caption{\textbf{Dependency switch rates by dataset and backbone (\%).}
Adapted methods report three-seed means; Backbone uses one fixed run.
Datasets contain \(128\), \(56\), and \(102\) sources, respectively.}
\label{tab:appendix_intervention_settings}
\vspace{4pt}

\begin{tabular*}{\linewidth}{@{\extracolsep{\fill}}lccccccccc@{}}
\toprule
\multirow{2}{*}{\textbf{Method}}
& \multicolumn{3}{c}{\textbf{2WikiMultiHopQA}}
& \multicolumn{3}{c}{\textbf{MuSiQue}}
& \multicolumn{3}{c}{\textbf{HotpotQA}} \\
\cmidrule(lr){2-4}
\cmidrule(lr){5-7}
\cmidrule(lr){8-10}
& \(S\) & \(T\) & \(C\)
& \(S\) & \(T\) & \(C\)
& \(S\) & \(T\) & \(C\) \\
\midrule

\multicolumn{10}{c}{\textit{Qwen3-8B}} \\
\addlinespace[2pt]

Backbone
& 99.22 & 37.50 & 84.38
& 96.43 & 46.43 & 94.64
& 99.02 & 65.69 & 75.49 \\

SFT
& 99.74 & 62.50 & 94.01
& 99.40 & 80.95 & 92.26
& 100.00 & 93.79 & 86.93 \\

CF-SFT
& 100.00 & 55.73 & 94.79
& 100.00 & 86.31 & 96.43
& 100.00 & 94.44 & 89.22 \\

RM
& 100.00 & 55.99 & 98.96
& 100.00 & 85.12 & 0.60
& 100.00 & 97.06 & 6.54 \\

\method{}
& 100.00 & 56.77 & 100.00
& 100.00 & 95.24 & 100.00
& 100.00 & 99.02 & 100.00 \\

\addlinespace[3pt]
\cdashline{1-10}[3pt/2pt]
\addlinespace[3pt]

\multicolumn{10}{c}{\textit{Qwen3-32B}} \\
\addlinespace[2pt]

Backbone
& 100.00 & 39.84 & 59.38
& 98.21 & 42.86 & 69.64
& 100.00 & 67.65 & 55.88 \\

SFT
& 100.00 & 65.10 & 87.50
& 100.00 & 82.74 & 75.00
& 100.00 & 94.44 & 69.93 \\

CF-SFT
& 100.00 & 59.38 & 91.15
& 100.00 & 88.69 & 76.79
& 100.00 & 94.77 & 76.80 \\

RM
& 99.48 & 57.55 & 68.23
& 100.00 & 86.31 & 11.90
& 100.00 & 93.14 & 10.13 \\

\method{}
& 100.00 & 58.33 & 100.00
& 100.00 & 94.64 & 100.00
& 100.00 & 98.04 & 100.00 \\

\addlinespace[3pt]
\cdashline{1-10}[3pt/2pt]
\addlinespace[3pt]

\multicolumn{10}{c}{\textit{Ministral-3-8B}} \\
\addlinespace[2pt]

Backbone
& 97.66 & 41.41 & 85.94
& 94.64 & 44.64 & 76.79
& 99.02 & 59.80 & 69.61 \\

SFT
& 99.22 & 58.85 & 99.22
& 99.40 & 82.74 & 89.88
& 99.35 & 93.14 & 84.97 \\

CF-SFT
& 99.48 & 59.64 & 99.22
& 100.00 & 85.12 & 92.26
& 100.00 & 94.77 & 92.81 \\

RM
& 99.74 & 60.94 & 62.76
& 100.00 & 83.93 & 8.33
& 100.00 & 94.77 & 6.21 \\

\method{}
& 99.22 & 64.58 & 100.00
& 100.00 & 94.05 & 100.00
& 100.00 & 99.02 & 100.00 \\

\bottomrule
\end{tabular*}
\end{table}

\subsubsection{Paired Comparisons}

\begin{wraptable}{r}{0.5\textwidth}
\centering
\small
\setlength{\tabcolsep}{3pt}
\renewcommand{\arraystretch}{1.12}
\caption{\textbf{Paired switch gains over CF-SFT.}
Gains are computed as \method{} minus CF-SFT and reported in percentage points, with \(95\%\) paired bootstrap confidence intervals.}
\label{tab:appendix_intervention_cis}
\vspace{4pt}

\begin{tabular*}{\linewidth}{@{\extracolsep{\fill}}lrr@{}}
\toprule
\textbf{Dependency}
& \textbf{Gain}
& \textbf{95\% CI} \\
\midrule
\(S\) & \(-0.03\)  & \([-0.17,\,+0.00]\) \\
\(T\) & \(+4.54\)  & \([+1.93,\,+7.33]\) \\
\(C\) & \(+10.06\) & \([+7.51,\,+12.81]\) \\
\bottomrule
\end{tabular*}
\end{wraptable}

Table~\ref{tab:appendix_intervention_cis} compares \method{} with CF-SFT using \(95\%\) percentile confidence intervals from \(10{,}000\) bootstrap resamples with seed \(20260915\).
Questions are resampled within datasets, with shared draws across methods and backbones.
Training seeds are resampled within each backbone, with shared draws across datasets and methods.
Dataset--backbone settings remain fixed.

The full objective improves \(T\)-switching by \(4.54\) percentage points and \(C\)-switching by \(10.06\) points, with both intervals above zero.
The \(S\)-switch difference is \(-0.03\) points, with an interval that includes zero.
Complete per-setting and per-seed directional rates are provided in the supplementary CSV.
These scores measure preference reversal between fixed candidates; perfect \(C\)-switching on these development sources does not imply perfect closure in freely generated responses.

\end{document}